\documentclass[11pt]{article}
\usepackage[most]{tcolorbox}
\usepackage{amssymb} 
\usepackage{booktabs}
\usepackage[dvipsnames]{xcolor}
\usepackage[table]{xcolor} 
\usepackage{caption}
\usepackage{pifont}  
\usepackage{kotex}

\usepackage[final]{acl}

\usepackage{times}
\usepackage{amsfonts}
\usepackage{latexsym}
\usepackage[T1]{fontenc}

\usepackage{stfloats}
\usepackage[utf8]{inputenc}
\usepackage{array} 
\usepackage{booktabs}
\usepackage[most]{tcolorbox}
\usepackage{cleveref}
\usepackage[table]{xcolor}
\usepackage{amsmath}
\usepackage{multirow}
\usepackage{placeins}

\usepackage{booktabs,tabularx,ragged2e,array}
\usepackage{subcaption}
\usepackage{graphicx}
\usepackage{pifont}
\usepackage{makecell}
\usepackage{enumitem}

\usepackage{algorithm}
\usepackage{algpseudocode}
\usepackage[table]{xcolor}
\usepackage{tabularx}
\definecolor{SRCAnchor}{RGB}{70,120,200}
\definecolor{TGTAnchor}{RGB}{220,145,60}
\definecolor{GapGray}{RGB}{235,235,235}

\newcommand{\srcanchor}[2]{\cellcolor{SRCAnchor!#1}\textcolor{black}{#2}}
\newcommand{\tgtanchor}[2]{\cellcolor{TGTAnchor!#1}\textcolor{black}{#2}}
\newcommand{\gapcell}[1]{\cellcolor{GapGray}{#1}}
\usepackage[most]{tcolorbox}

\usepackage{adjustbox}
\usepackage{placeins}

\tcbset{
  promptbox/.style={
    enhanced,
    breakable,
    colback=gray!3,
    colframe=black!35,
    boxrule=0.35pt,
    arc=1.2pt,
    left=4pt,
    right=4pt,
    top=4pt,
    bottom=4pt,
    fonttitle=\bfseries\small,
    coltitle=black,
    colbacktitle=gray!15,
    attach boxed title to top left={xshift=2mm,yshift=-1mm},
    boxed title style={
      boxrule=0pt,
      arc=1pt,
      left=3pt,
      right=3pt,
      top=1pt,
      bottom=1pt
    }
  }
}

\definecolor{SRCAnchor}{RGB}{70,120,200}
\definecolor{TGTAnchor}{RGB}{220,145,60}
\definecolor{GapGray}{RGB}{235,235,235}
\definecolor{SectionGray}{RGB}{246,246,246}
\definecolor{MeanGray}{RGB}{250,250,250}
\definecolor{GroupSep}{RGB}{215,215,215}
\definecolor{DropRed}{RGB}{170,55,55}
\newcommand{\fdrop}[2]{#1\,{\scriptsize\textcolor{DropRed}{(#2)}}}
\newcommand{\logodir}{figures/logos}
\newcommand{\modelwithlogo}[2]{%
  \raisebox{-0.16em}{\includegraphics[height=0.86em]{\logodir/#1-logo.pdf}}\hspace{0.35em}#2%
}

\DeclareUnicodeCharacter{266B}{}
\newcommand{\role}[1]{\textbf{\textsf{#1:}}}
\algrenewcommand\algorithmicindent{0.8em}

\definecolor{CanvasGreen}{HTML}{2E7D32}
\definecolor{CanvasRed}{HTML}{B3261E}
\definecolor{GroupSep}{HTML}{B0B0B0}

\newcommand{\canvasgain}[1]{\textcolor{CanvasGreen}{#1}}
\newcommand{\canvasloss}[1]{\textcolor{CanvasRed}{#1}}
\newcommand{\canvasbest}[1]{\textbf{#1}}
\newcommand{\canvasgainbest}[1]{\textbf{\textcolor{CanvasGreen}{#1}}}

\title{Code-Switching Reveals Language Anchoring in Multilingual LLMs}

\author{
{\bfseries
Jeonghyun Park$^{1}$,
Seunghyun Yoon$^{2}$,
Yonghyun Jun$^{1}$,
Hwanhee Lee$^{1\dagger}$
}
\\
$^{1}$Chung-Ang University, Seoul, Korea, $^{2}$Adobe Research, USA
\\
\texttt{\{tom0365, zgold5670, hwanheelee\}@cau.ac.kr, syoon@adobe.com}
\\
\url{https://jeonghyunpark2002.github.io/CANVAS_project_page}
}

\begin{document}
\maketitle

\footnotetext{\textsuperscript{$\dagger$}Corresponding author.}

  

\begin{abstract}
Multilingual Large Language Models (MLLMs) are increasingly expected to handle Code-Switched (CS) inputs, yet mixing languages frequently degrades performance relative to source- or target-language monolingual counterparts.
To understand this degradation, we use grammar-forced CS as a controlled diagnostic setting for locating CS representations relative to their source and target counterparts.
We introduce \textbf{Anchor Bias}, a geometric measure that quantifies \textit{language anchoring}—whether a CS hidden state aligns closer to its source or target language counterpart. 
Across diverse MLLMs, Anchor Bias reveals a consistent grammar-frame effect: source-framed CS stays source-anchored, whereas target-framed CS shifts target-ward and shows larger Question Answering (QA) degradation.
Motivated by this representational pattern, we propose \textbf{CANVAS} (\textbf{C}ontextual \textbf{A}nchor-based \textbf{N}eural \textbf{V}ector \textbf{A}lignment \textbf{S}teering), an inference-time intervention that extracts a source-side canvas from the input and softly steers target-language hidden states toward the source anchor during prefill.
CANVAS consistently recovers QA F1 across MLLMs and CS conditions, showing that internal anchoring signals provide an actionable target for mitigating CS inference failures.
The Code is available at \url{https://github.com/jeonghyunpark2002/CANVAS.git}
\end{abstract}


\section{Introduction}
\label{sec:intro}

Multilingual Large Language Models (MLLMs) are increasingly expected to handle Code-Switched (CS) inputs, where users mix multiple languages within an interaction. However, standard evaluations~\cite{hu2020xtrememassivelymultilingualmultitask,  liang-etal-2020-xglue} largely assume monolingual contexts, overlooking this pervasive practice. Crucially, CS is not random lexical noise but a structured linguistic phenomenon governed by grammatical organization \cite{dogruoz-etal-2021-survey,winata-etal-2023-decades}. Consequently, an MLLM competent on monolingual inputs can still face unique processing failures when an information need is expressed through a structured mixed-language form, frequently degrading performance compared to monolingual counterparts.

\begin{figure}[!t]
  \centering
  \includegraphics[width=0.92\linewidth]{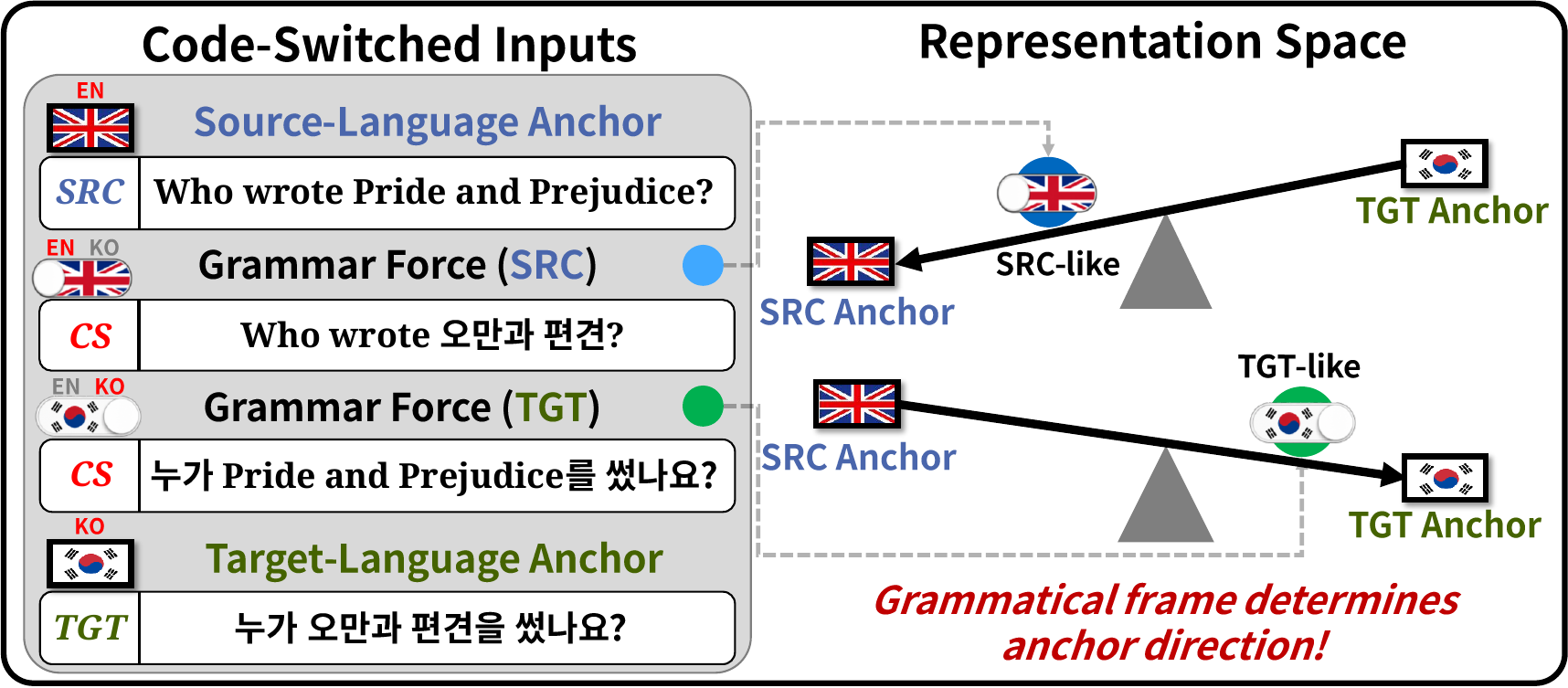}
  \caption{\textbf{Locating code-switched inputs in the multilingual representation space.} Source- and target-framed CS variants preserve different grammatical structures for the same information need, plotted alongside their matched monolingual anchors.}
  \label{fig:schematic}
\end{figure}

Prior work evaluates CS across robustness, translation, generation, and multilingual generalization \cite{laureano-de-leon-etal-2024-code,huzaifah2024evaluating,abdaljalil2025evaluating,mohamed2025lost,heredia2025conditioning,winata2026can}. 
While these studies show that mixed-language inputs expose fragilities hidden by monolingual benchmarks, and existing MLLM analyses suggest latent language preferences or English-centered processing pathways \cite{wendler-etal-2024-llamas,zhao2024largelanguagemodelshandle,sharma-etal-2025-faux}, neither explains how a CS instance is geometrically positioned relative to its matched monolingual counterparts.

Understanding the representational pattern behind this performance drop requires mapping where the mixed-language hidden state lies relative to its monolingual references: source-aligned, target-shifted, or intermediate. Since identical QA failures can stem from different internal representational causes, we probe the geometric layout of the model's representation space rather than relying only on answer-level analysis.

To this end, we study code-switched QA in a controlled setting designed for representation-level diagnosis.
For each CS question, we use semantically matched source- and target-language counterparts as monolingual references.
Following CodeMixQA \citep{winata2026can}, we leverage grammar-forced CS variants to isolate the \emph{grammatical frame}. By holding the question fixed while isolating which language organizes the utterance, we test whether the CS state stays between monolingual references or moves toward the frame language's representation.

\Cref{fig:schematic} illustrates this framework with an English--Korean example. 
A source-framed question (e.g., \textit{''Who wrote 오만과 편견?''}) embeds target lexical items into a source syntactic backbone, while a target-framed question (e.g., \textit{'누가 Pride and Prejudice를 썼나요?''}) does the reverse.
Together with the matched source-only and target-only questions, which serve as the monolingual anchors, this paired construction isolates the frame variable and reveals which monolingual subsystem the MLLM internally follows when interpreting the same cross-lingual intent.

To quantify this representation-level phenomenon, we introduce \textbf{Anchor Bias}, a geometric measure assessing whether a CS hidden state aligns closer to its source- or target-language counterpart. 
Our analysis across diverse MLLMs reveals a consistent, frame-dependent pattern: source-framed CS remains close to the source anchor, whereas target-framed CS exhibits a sharp representational shift target-ward, coinciding with larger performance drops.
This suggests that CS inputs tend to degrade more when their representations move away from the source-side trajectory that better preserves answer quality.

Motivated by these geometric findings, we propose \textbf{CANVAS} (\textbf{C}ontextual \textbf{A}nchor-based \textbf{N}eural \textbf{V}ector \textbf{A}lignment \textbf{S}teering) to test whether a source-ward representation correction can mitigate this shift via an inference-time intervention.
CANVAS extracts a source-side canvas from the in-context input and softly steers target-language hidden states toward that canvas during prefill. 

Across MLLMs and CS conditions, CANVAS consistently improves QA F1 without parameter updates. Its largest gains appear under target-framed CS, suggesting that source-canvas alignment can mitigate the target-ward anchoring pattern identified by our analysis, a pattern that reflects realistic bilingual usage such as native-language sentence frames with embedded English technical terms \cite{bali2014borrowing,winata-etal-2023-decades,yong-etal-2023-prompting}.

\section{Experimental Setup}
\label{sec:probe}

\subsection{Experimental Framework}
\label{subsec:experimental_framework}
\paragraph{Task setup.}
We analyze how MLLMs internally process mixed-language intent using code-switched QA as a controlled setting. For a given information need, we evaluate a matched quadrant of question variants ($\mathcal{Q}(x)$) across a source language, a target language, and two distinct CS forms, as illustrated in \Cref{fig:schematic}. Following common English--target-language evaluation conventions \cite{laureano-de-leon-etal-2024-code,winata-etal-2023-decades,winata2026can}, we treat English as the source language and the paired non-English language as the target; Appendix~\ref{sec:es_pivot} tests source-language generalization. 
Each instance keeps the factual query fixed and varies only the linguistic frame, enabling direct representational and behavioral comparison.
\paragraph{Dataset and Variant Construction.}
To align our evaluation, we utilize the fact-seeking queries from SimpleQA Verified \cite{haas2026simpleqaverifiedreliablefactuality} as the source language questions ($q^{\mathrm{SRC}}$) and map them to the corresponding grammar-forced CS pairs from CodeMixQA~\cite{winata2026can}. Each pair provides two structural alignments: GF-SRC (English structure with target lexical inserts, $q^{\mathrm{SRC}}$) and GF-TGT (target structure with English lexical inserts, $q^{\mathrm{TGT}}$). 
For the target-language counterpart ($q^{\mathrm{TGT}}$), we translate the source question via Qwen3-235B-A22B \cite{yang2025qwen3}.
This pipeline yields a complete matched quadruplet $\mathcal{Q}(x)=\{q^{\mathrm{SRC}}, q^{\mathrm{TGT}}, q^{\mathrm{GF\text{-}SRC}}, q^{\mathrm{GF\text{-}TGT}}\}$ evaluated against a unified reference answer. We pair $q^{\mathrm{SRC}}$ with nine target languages (Bengali, Spanish, French, Hindi, Korean, Marathi, Toba Batak, Urdu, Chinese), evaluating 955 unique source queries across 2{,}880 matched comparisons (see Appendix~\ref{app:dataset_statistics} for detailed statistics).

\paragraph{Target-Language Anchors.}
Because the GF-TGT condition adheres to the target-language syntactic frame, word-level translation largely preserves its surface backbone. Consequently, $q^{\mathrm{TGT}}$ serves as a natural target-language anchor for representation-level analysis. Appendix~\ref{app:translation_caveat} provides further discussion on alignment.

\paragraph{Terminology.}
\textsc{SRC}/\textsc{TGT} refer \emph{only} to the fixed language pair of an experiment; GF-SRC/GF-TGT indicate which of the two supplies the matrix-language grammar of a CS sentence. For representations we write $s_{\mathrm{SRC}}$/$s_{\mathrm{TGT}}$ for the normalized similarity of a CS hidden state to its \textsc{SRC}/\textsc{TGT} counterpart (\Cref{sec:anchor_bias}), so that $\mathrm{AB}=s_{\mathrm{SRC}}-s_{\mathrm{TGT}}$: $\mathrm{AB}>0$ is SRC-ward and $\mathrm{AB}<0$ is TGT-ward.

\subsection{Implementation Details}
\label{subsec:implementation_details}
\paragraph{Models.}
To examine whether language anchoring appears across architectures and model families, we evaluate various MLLMs. 
We use Aya-Expanse-8B \cite{dang2024aya}, Qwen3.5-4B, Qwen3.5-9B, Qwen3.5-27B \cite{yang2025qwen3}, Llama-3.1-70B-Instruct, Llama-3.3-70B-Instruct \cite{meta2024llama3}, Mixtral-8x7B-Instruct \cite{jiang2024mixtral}, Phi-3.5-MoE-Instruct \cite{abdin2024phi}, Qwen3-30B-A3B \cite{yang2025qwen3}, and Qwen3.6-35B-A3B \cite{yang2025qwen3}. 

\paragraph{Inference Setups.}
We use greedy decoding throughout our experiments. Translations are generated with Qwen3-235B-A22B-2507~\cite{yang2025qwen3} via OpenRouter; all other models in the pool are run locally. Details are reported in Appendix~\ref{app:implementation} (with prompt templates in Appendix~\ref{app:prompts}).

\subsection{Evaluation Metrics}
\label{subsec:evaluation_metrics}

\paragraph{Evaluation Metric.}
The model receives one question variant at a time and generates an answer. We score each generated answer against the SimpleQA reference answer using normalized token-overlap F1. We report F1 per condition and summarize CS degradation as the gap between F1 on $q^{\mathrm{SRC}}$ and F1 on $q^{\mathrm{CS}}$.

\paragraph{Internal Hidden Signals.}
Output F1 establishes whether CS changes answer quality, but it does not reveal how the model internally interprets the mixed-language question. We therefore extract hidden states from the question region for matched source, target, and CS variants and compare the CS representation against both monolingual counterparts. This representation comparison asks where CS lies relative to the two reference language states and forms the basis of the anchor-bias analysis in \Cref{sec:anchor_bias}.


\section{Language Anchoring of Code-Switched Inputs}
\label{sec:anchor_bias}
We investigate where a code-switched hidden state lies in the multilingual representation space relative to its matched monolingual counterparts. Geometrically, if a CS input is processed as a balanced bilingual construction, its representation should occupy an intermediate space. Conversely, if the mixed input adheres to a dominant language processing pathway, its hidden state will skew toward either the source or target anchor. To formally quantify this position, we introduce \textit{anchor bias}, a geometric metric that reveals a consistent, frame-dependent language anchoring pattern across diverse MLLM architectures and transformer layers.

\subsection{Formulating Anchor Bias}
\paragraph{Metric Derivation.}
For a matched question set, let $q_s=q^{\mathrm{SRC}}$, $q_t=q^{\mathrm{TGT}}$, and $q_c=q^{\mathrm{CS}}$. For a model $M$ and question variant $q$, let $H_M(q)\in\mathbb{R}^{L\times T\times d}$ denote the hidden states over layers, tokens, and dimensions.
We mean-pool the question-content token span $C(q)$ (excluding shared instruction and template markers) to obtain the layer-wise context vector $\mathbf{r}_{\ell}(q) \in \mathbb{R}^d$:
\begin{equation}
\mathbf{r}_{\ell}(q)
=
\frac{1}{|C(q)|}
\sum_{t\in C(q)}
H_M(q)[\ell,t,:].
\end{equation}
We define the layer-wise similarity as $s_{\ell}(q_i,q_j)=\cos(\mathbf{r}_{\ell}(q_i),\mathbf{r}_{\ell}(q_j))$. The raw layer-wise anchor bias of a CS question is then defined as $\mathrm{AB}_{\ell}(q_c)=s_{\ell}(q_s,q_c)-s_{\ell}(q_t,q_c)$, where positive values indicate closer proximity to the source anchor and negative values signal a target-ward skew.

\begin{table*}[t]
\centering
\small
\setlength{\tabcolsep}{2.4pt}
\renewcommand{\arraystretch}{1.06}
\begin{adjustbox}{max width=\textwidth}
\begin{tabular}{lccc!{\hspace{4pt}\color{GroupSep}\vrule width 0.25pt\hspace{4pt}}cccc!{\hspace{4pt}\color{GroupSep}\vrule width 0.25pt\hspace{4pt}}ccc}
\toprule
\multirow{2}{*}{\raisebox{-0.55ex}{\textbf{Model}}} &
\multicolumn{3}{c}{$\mathrm{AB}^{\mathrm{upper}}_{\mathrm{nrm}}$} &
\multicolumn{4}{c}{\textbf{QA behavior (F1)}} &
\multicolumn{3}{c}{\textbf{SRC-anchored layers (\%)}} \\
\cmidrule(lr){2-4} \cmidrule(lr){5-8} \cmidrule(lr){9-11}
& \textsc{GF-Src} & \textsc{GF-Tgt} & $\Delta_{\mathrm{AB}}$
& \textsc{SRC} & \textsc{GF-Src} & \textsc{GF-Tgt} & \textsc{TGT}
& \textsc{GF-Src} & \textsc{GF-Tgt} & $\Delta_{\mathrm{layer}}$ \\
\midrule

\modelwithlogo{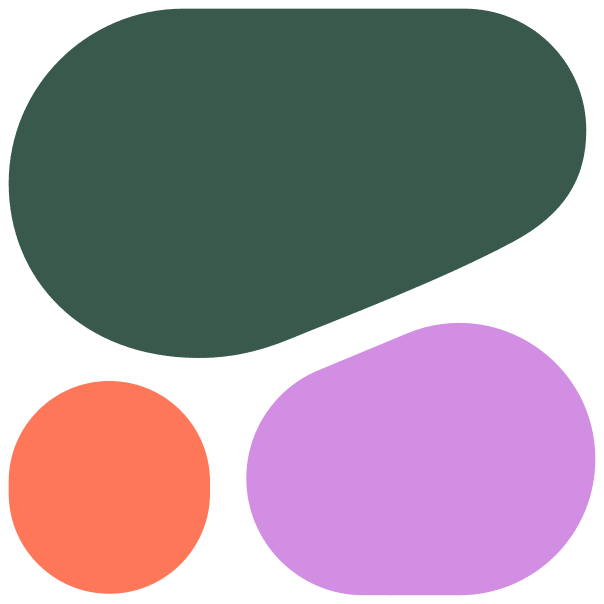}{Aya-Expanse-8B}
& \srcanchor{22}{+0.300} & \tgtanchor{16}{-0.220} & \gapcell{+0.520}
& 9.3 & \fdrop{7.0}{-2.3} & \fdrop{4.4}{-4.9} & 2.4
& 85.6 & 42.8 & +42.8 \\

\modelwithlogo{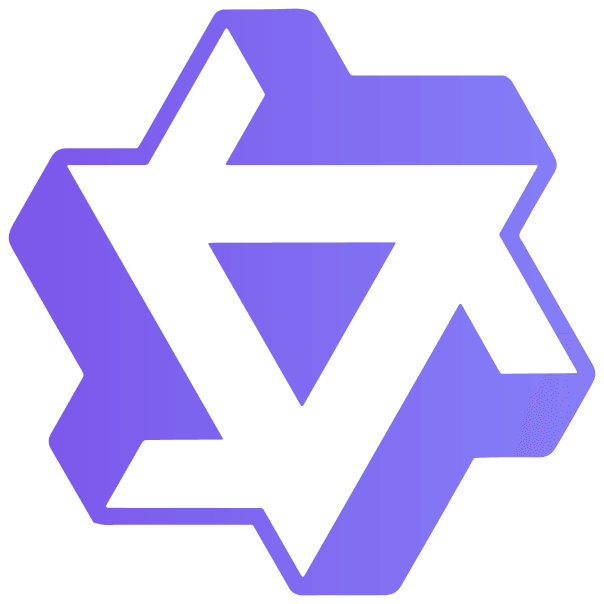}{Qwen3.5-4B}
& \srcanchor{12}{+0.169} & \tgtanchor{19}{-0.262} & \gapcell{+0.431}
& 10.0 & \fdrop{8.2}{-1.8} & \fdrop{7.2}{-2.8} & 4.8
& 70.7 & 29.2 & +41.5 \\

\modelwithlogo{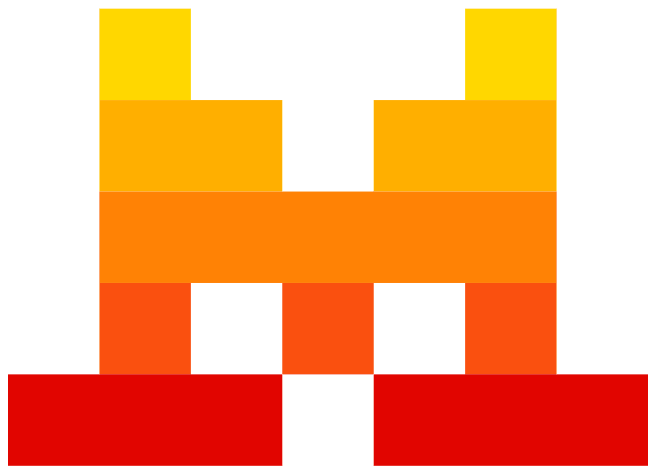}{Mixtral-8x7B}
& \srcanchor{20}{+0.273} & \tgtanchor{34}{-0.452} & \gapcell{+0.725}
& 10.8 & \fdrop{6.7}{-4.1} & \fdrop{5.8}{-5.0} & 4.5
& 74.6 & 33.2 & +41.4 \\

\modelwithlogo{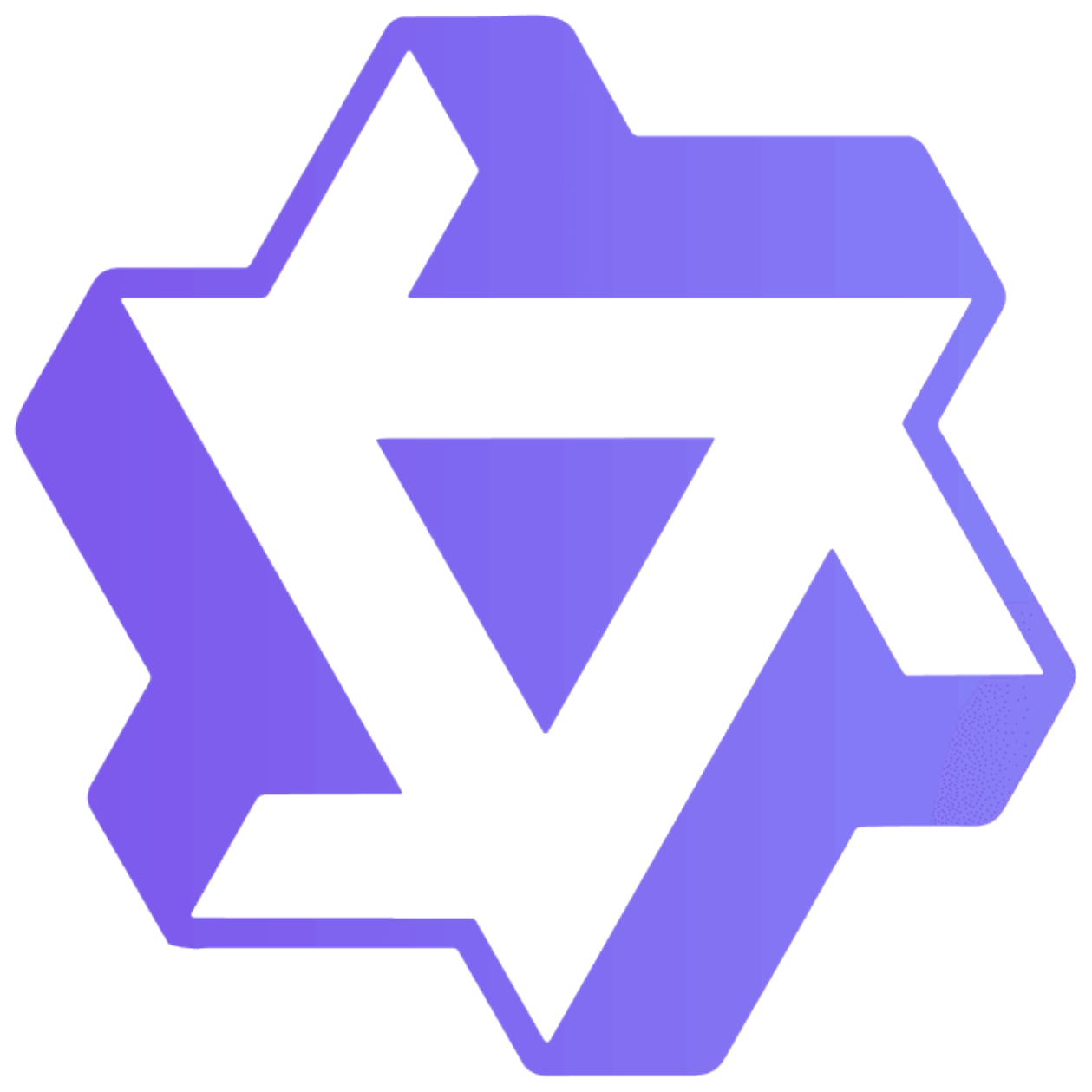}{Qwen3-30B-A3B}
& \srcanchor{30}{+0.393} & \tgtanchor{6}{-0.081} & \gapcell{+0.474}
& 11.2 & \fdrop{8.8}{-2.4} & \fdrop{6.5}{-4.7} & 7.7
& 78.5 & 42.1 & +36.4 \\

\modelwithlogo{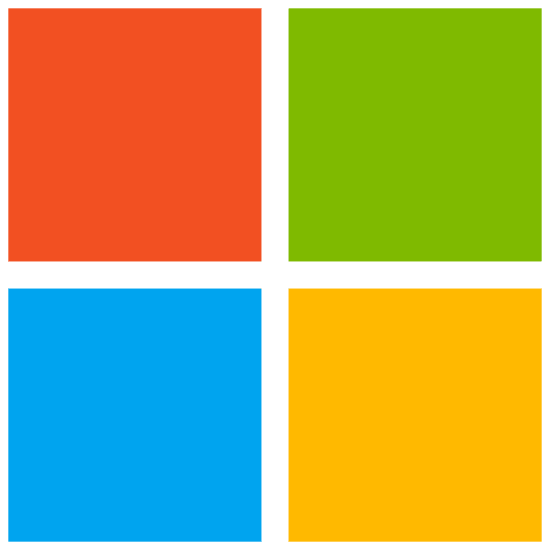}{Phi-3.5-MoE}
& \srcanchor{18}{+0.248} & \tgtanchor{22}{-0.300} & \gapcell{+0.548}
& 13.1 & \fdrop{9.6}{-3.5} & \fdrop{6.5}{-6.6} & 3.0
& 64.8 & 31.7 & +33.1 \\

\modelwithlogo{qwen}{Qwen3.5-9B}
& \srcanchor{14}{+0.185} & \tgtanchor{19}{-0.256} & \gapcell{+0.441}
& 13.1 & \fdrop{11.9}{-1.2} & \fdrop{8.1}{-5.0} & 8.7
& 72.9 & 32.5 & +40.4 \\

\modelwithlogo{qwen}{Qwen3.5-27B}
& \srcanchor{11}{+0.154} & \tgtanchor{15}{-0.197} & \gapcell{+0.351}
& 16.6 & \fdrop{13.3}{-3.3} & \fdrop{10.4}{-6.2} & 11.8
& 73.5 & 34.4 & +39.1 \\

\modelwithlogo{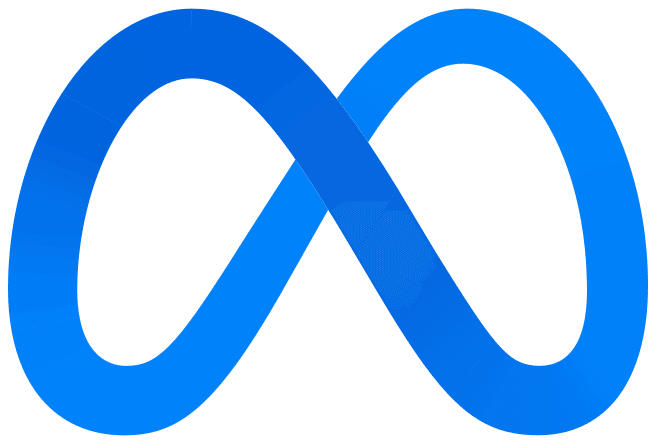}{Llama3.3-70B}
& \srcanchor{20}{+0.267} & \tgtanchor{14}{-0.191} & \gapcell{+0.458}
& 18.9 & \fdrop{13.7}{-5.2} & \fdrop{11.6}{-7.3} & 6.5
& 81.4 & 39.5 & +41.9 \\

\modelwithlogo{qwen}{Qwen3.6-35B-A3B}
& \srcanchor{16}{+0.215} & \tgtanchor{76}{-1.011} & \gapcell{+1.226}
& 21.4 & \fdrop{17.0}{-4.4} & \fdrop{13.5}{-10.3} & 11.1
& 76.2 & 38.1 & +38.1 \\

\modelwithlogo{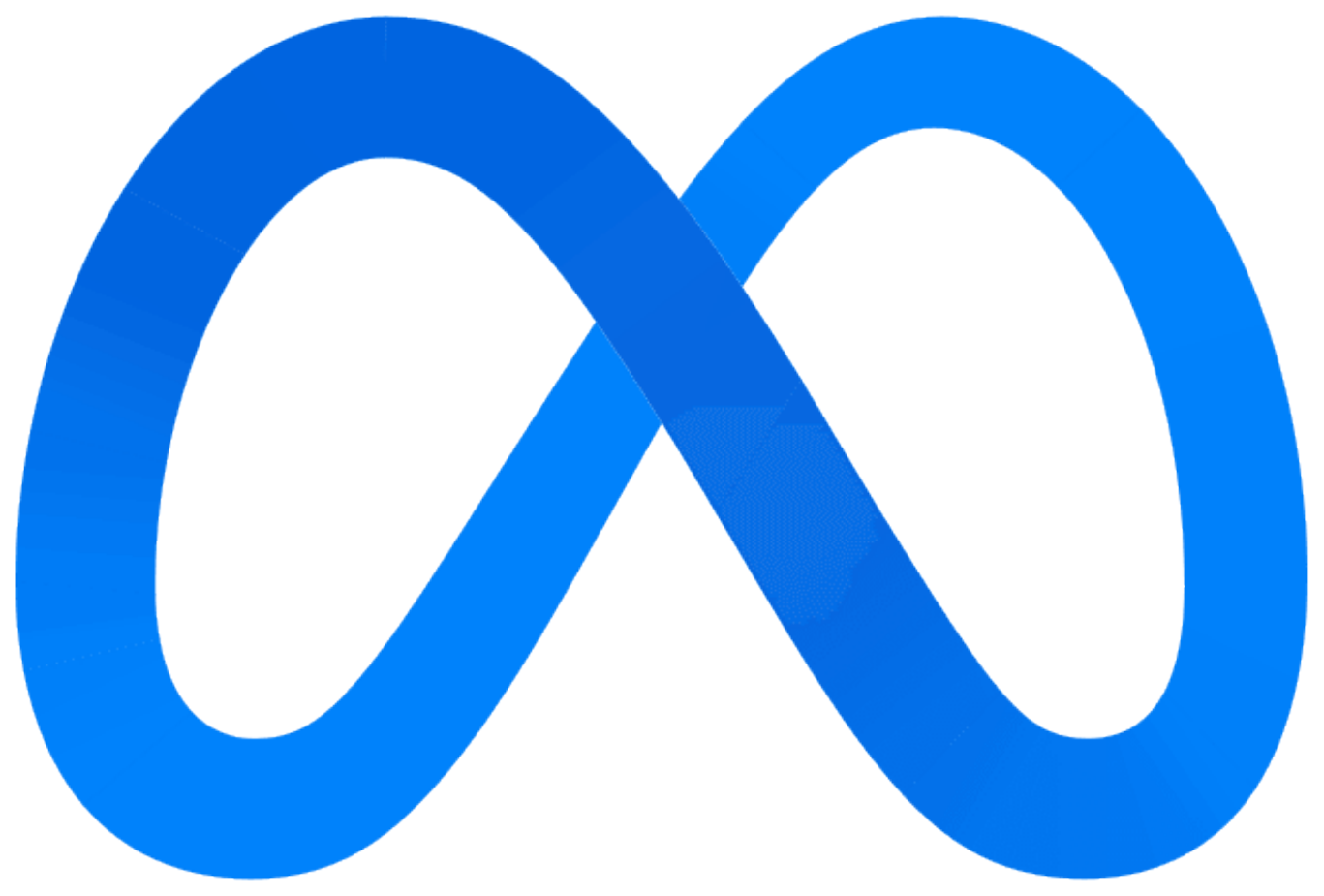}{Llama3.1-70B}
& \srcanchor{19}{+0.257} & \tgtanchor{17}{-0.225} & \gapcell{+0.482}
& 23.1 & \fdrop{16.9}{-6.2} & \fdrop{13.5}{-9.6} & 6.2
& 80.6 & 38.0 & +42.6 \\

\midrule
\rowcolor{MeanGray}
\textbf{\textit{Mean}}
& \srcanchor{18}{+0.246} & \tgtanchor{24}{-0.320} & \gapcell{+0.566}
& 14.8 & \fdrop{11.3}{-3.5} & \fdrop{8.7}{-6.2} & 6.7
& 75.9 & 36.2 & +39.7 \\
\bottomrule
\end{tabular}
\end{adjustbox}
\caption{\textbf{Effect of grammatical frame across models.}
Upper-layer anchor bias, QA F1 on source, code-switched, and target-language inputs, and the percentage of source-anchored layers. Red values indicate F1 drops relative to \textsc{SRC}; \textsc{TGT} is reported as a monolingual reference.}
\label{tab:main_contrast}
\end{table*}

\paragraph{Anisotropy Normalization.}
Raw cosine similarities are not directly comparable across model families due to shifting background anisotropies and scale variations in multilingual hidden spaces~\cite{ethayarajh2019contextual, timkey-van-schijndel-2021-bark}. In highly anisotropic spaces, unrelated input representations can exhibit artificially high baseline similarity, obscuring relative cross-lingual distances (see Appendix~\ref{app:anisotropy} for baseline diagnostics). We counteract this by applying a layer-specific random cosine normalization. We estimate $\mu_{\ell}^{M}$ as the mean cosine over random source question pairs with non-overlapping answers, capturing pure directional bias rather than semantic overlap. The normalized similarity is computed as:

\begin{equation}
\tilde{s}_{\ell}(q_i,q_j)=\frac{s_{\ell}(q_i,q_j)-\mu_{\ell}^{M}}{1-\mu_{\ell}^{M}}.
\end{equation}
Using this normalized similarity, we define normalized layer-wise anchor bias as $\widetilde{\mathrm{AB}}_{\ell}(q_c)=\tilde{s}_{\ell}(q_s,q_c)-\tilde{s}_{\ell}(q_t,q_c)$.
To obtain a stable model-level estimate, we average $\widetilde{\mathrm{AB}}_{\ell}(q_c)$ over the upper half of transformer layers ($\mathcal{L}_{\mathrm{upper}}$):
\begin{equation}
\mathrm{AB}^{\mathrm{upper}}_{\mathrm{nrm}}(q_c)
=
\frac{1}{|\mathcal{L}_{\mathrm{upper}}|}
\sum_{\ell\in\mathcal{L}_{\mathrm{upper}}}
\widetilde{\mathrm{AB}}_{\ell}(q_c).
\label{eq:anchor_bias_upper}
\end{equation}
This restriction follows prior findings that upper layers encode more abstract, task-aligned semantic representations than lower ones \cite{peters2018dissecting, tenney2019bert}, a property extending to modern generative LLMs \cite{marks2024the}, thereby yielding a single robust anchor-bias score per condition.

\subsection{Frame-Dependent Anchoring Analysis}
\paragraph{Frame-Dependent Language Anchoring.}
As shown in \Cref{tab:main_contrast}, internal anchor bias consistently follows the input's grammatical backbone rather than lexical mix. GF-SRC (source-framed) exhibits a positive mean anchor bias of $+0.246$, confirming orientation toward the source anchor. Conversely, GF-TGT (target-framed) demonstrates a distinct target-ward shift with a negative mean bias of $-0.320$. 
This swing ($\Delta_{\mathrm{AB}}=+0.566$) shows that the grammatical frame substantially \emph{shifts} internal language anchoring, although it does not solely determine its direction: same-lexicon order counterfactuals show that lexical language sets the sign of AB while the frame changes its magnitude (Appendix~\ref{app:ko_english_order_anchor}).

\paragraph{Controlling for Token Proportion.}
GF-SRC and GF-TGT also differ in source-token proportion ($\rho{=}0.59$ vs.\ $0.40$; \Cref{fig:canvas_alpha}), so the contrast could be a token-ratio effect. Restricting to a balanced subset $\rho\in[0.40,0.60]$ shrinks the per-example separation but leaves it clearly positive: $\Delta_{\mathrm{AB}}=+0.58$ for Aya-Expanse-8B and $+0.72$ for Qwen3.5-9B (both $p<10^{-7}$; narrower bands in Appendix~\ref{app:rho_control}). The frame effect is therefore not reducible to how many tokens come from each language.


\paragraph{Behavioral QA Alignment of Anchor Bias.}
This frame-dependent representational shift corresponds to QA performance. 
As shown in \Cref{tab:main_contrast}, GF-SRC produces a smaller average F1 degradation relative to the monolingual source input ($-3.5$) than GF-TGT ($-6.2$), suggesting that closer proximity to the source anchor tends to preserve task performance better.
This behavioral alignment is consistent across individual model--condition pairs, where $\mathrm{AB}^{\mathrm{upper}}_{\mathrm{nrm}}$ correlates positively with $\Delta\mathrm{F1}$ ($\rho=+0.485$, $p<0.01$). 

\paragraph{Layer-Wise Robustness.}
We examine whether this anchoring shift persists across model depth. Specifically, we quantify the percentage of layers exhibiting positive $\mathrm{AB}^{\mathrm{upper}}_{\mathrm{nrm}}$ (i.e., source-anchored layers) for each model and grammar-forced condition. As shown in \Cref{tab:main_contrast}, GF-SRC yields a predominantly source-heavy trajectory across $75.9\%$ of layers on average, whereas GF-TGT drops sharply to $36.2\%$. This substantial gap demonstrates that the structural frame influences internal representation trajectories robustly throughout the model's depth rather than just at the final layers.

\begin{figure*}[t]
  \centering
  \includegraphics[width=0.9\linewidth]{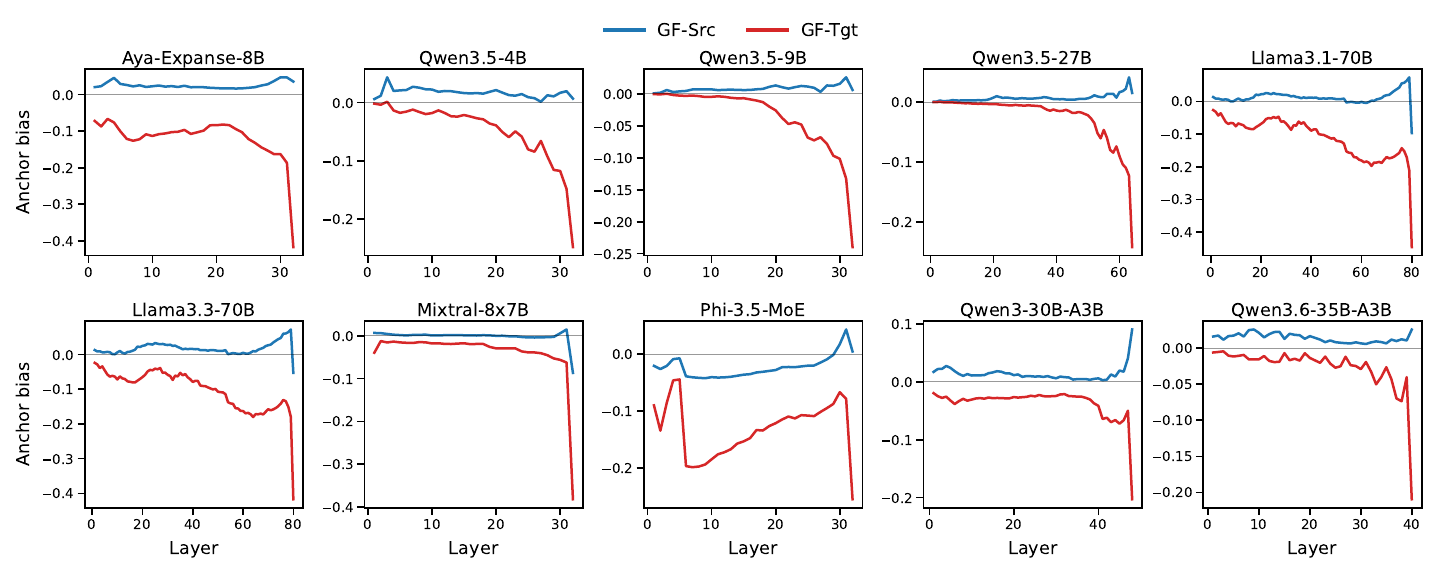}
  \caption{\textbf{Per-layer anchor bias across model depth.} Raw $\mathrm{AB}_\ell$ on question-content tokens, averaged across examples, plotted against layer index for each model. Positive: source-anchored; negative: target-anchored.}
  \label{fig:layer_anchor_depth}
\end{figure*}


\paragraph{Per-layer anchor trajectory.}
We further visualize how anchor bias evolves across model depth. 
Table~\ref{tab:main_contrast} reports normalized upper-layer anchor bias for cross-model comparison, whereas Figure~\ref{fig:layer_anchor_depth} plots raw $\mathrm{AB}_\ell$ at each layer for the two grammar-forced CS conditions: GF-SRC and GF-TGT. 
We use raw $\mathrm{AB}_\ell$ for this layer-wise visualization because normalization rescales $\mathrm{AB}_\ell$ by the positive factor $1/(1-\mu_\ell)$: it preserves the sign and within-layer ordering, but can visually over-amplify lower layers when $\mu_\ell$ is close to 1 due to anisotropy.
Figure~\ref{fig:layer_anchor_depth} confirms the same frame-dependent pattern observed in Table~\ref{tab:main_contrast}. 
GF-SRC stays mostly source-oriented, while GF-TGT shifts target-ward. 
This contrast is small in early layers and becomes clearer toward the top of the model, with the strongest target-ward anchoring concentrated in the final $30\%$ of layers. 
This depth profile motivates applying intervention at later layers, where the source--target anchoring contrast is more pronounced, rather than during earlier representation formation.

\subsection{Anchoring Beyond English as Source}
\label{sec:pivot_generalization}
Since English is the source throughout, the pattern could reflect an English-centric bias \cite{wendler-etal-2024-llamas,zhao2024largelanguagemodelshandle} rather than a frame effect. We therefore replace the source/frame-reference language while keeping the QA instances and GF construction fixed. With Spanish as source, the split replicates across four models (mean $\Delta_{\mathrm{AB}}=+0.541$; Appendix~\ref{sec:es_pivot}); translating the source tokens into Arabic, Hindi, Korean, and Chinese (Qwen3-235B-A22B) and evaluating Aya-Expanse-8B, $\Delta_{\mathrm{AB}}$ stays consistently positive across Arabic, Devanagari, Hangul, and Han scripts (\Cref{tab:cr_pivot_langs}). Anchoring thus tracks the language supplying the grammatical frame, not English identity. A within-language script control (the same Hindi questions in Devanagari vs.\ romanized form) shows the separation persists with language and content fixed ($\Delta_{\mathrm{AB}}=1.08$ vs.\ $3.74$), while its magnitude is script-sensitive (Appendix~\ref{app:hindi_script}).

\begin{table}[t]
\centering
\small
\setlength{\tabcolsep}{3.5pt}
\renewcommand{\arraystretch}{1.05}
\begin{adjustbox}{max width=\columnwidth}
\begin{tabular}{@{}lccccc@{}}
\toprule
\textbf{Source (frame) pivot} & \textsc{Es} & \textsc{Ar} & \textsc{Hi} & \textsc{Ko} & \textsc{Zh} \\
\midrule
Script & Latin & Arabic & Devanagari & Hangul & Han \\
$\Delta_{\mathrm{AB}}$ (GF-Src $-$ GF-Tgt) & $+0.70$ & $+0.29$ & $+0.32$ & $+0.28$ & $+0.36$ \\
\bottomrule
\end{tabular}
\end{adjustbox}
\caption{\textbf{Anchoring beyond English as source.}
Grammar-frame separation $\Delta_{\mathrm{AB}}$ with non-English source/frame-reference languages, evaluated with Aya-Expanse-8B on the same QA instances and GF-Src/GF-Tgt construction (English pivot: $+0.52$, Table~\ref{tab:main_contrast}). Full Spanish-pivot results are in Appendix~\ref{sec:es_pivot}.}
\label{tab:cr_pivot_langs}
\end{table}

\section{Aligning Code-Switched Representations with CANVAS}
\label{sec:canvas}

Motivated by our finding that target-ward inputs become increasingly target-anchored in the upper layers of target-framed code-switched inputs (\Cref{sec:anchor_bias}), we propose a precisely targeted intervention for CS inputs: \textbf{CANVAS} (Contextual Anchor-based Neural Vector Alignment Steering). CANVAS leverages this late-stage geometric window to extract an in-context source canvas from the input itself and softly steers target-language hidden states toward it during prefill. 
We apply this steering only in the designated upper layers, where the source--target anchoring contrast is most pronounced, and set its strength using the instance-specific controller introduced below.
\paragraph{Stage 1: Token Partition.}
CANVAS starts by separating the CS input into source, target, and shared spans, which are then used to estimate a source canvas and set the interpolation strength. 
Given a CS input $q^{\mathrm{CS}}$, we partition the input span as $C(q^{\mathrm{CS}})=C_{\mathrm{src}}\cup C_{\mathrm{tgt}}\cup C_{\mathrm{oth}}$, where $C_{\mathrm{src}}$ contains source-language tokens, $C_{\mathrm{tgt}}$ contains target-language tokens, and $C_{\mathrm{oth}}$ contains punctuation, numbers, whitespace artifacts, or other language-neutral tokens. We partition input tokens using lightweight token-level language tagging; implementation details are provided in Appendix~\ref{app:canvas_token_tagging}.


\paragraph{Stage 2: Source Canvas Construction.}
Using this partition, CANVAS runs an initial forward prefill pass to extract baseline hidden states $H_M(q^{\mathrm{CS}})\in\mathbb{R}^{L\times T\times d}$ from the templated user-query span (see Appendix~\ref{app:canvas_hidden_extraction} for the extraction protocol). We restrict intervention to the upper layers $\mathcal{L}_{\mathrm{ctrl}}=\{\lfloor0.7L\rfloor,\ldots, L-1\}$, isolating the final $30\%$ of model depth where abstract semantic anchoring stabilizes. These later layers provide more robust contextual targets for alignment than lower layers tied to surface forms. 
For each layer $\ell\in\mathcal{L}_{\mathrm{ctrl}}$, we compute two layer-specific anchors—the \emph{source canvas} ($\mathbf{c}^{\mathrm{src}}_{\ell}$) and the \emph{target anchor} ($\mathbf{a}^{\mathrm{tgt}}_{\ell}$)—via mean-pooling over their respective token sets:\label{eq:canvas_anchors}

\begin{equation}
\mathbf{c}^{\mathrm{src}}_{\ell}
=
\frac{1}{|C_{\mathrm{src}}|}
\sum_{t\in C_{\mathrm{src}}}
H_M(q^{\mathrm{CS}})[\ell,t,:],
\label{eq:canvas_src_canvas}
\end{equation}
and the \emph{target anchor},
\begin{equation}
\mathbf{a}^{\mathrm{tgt}}_{\ell}
=
\frac{1}{|C_{\mathrm{tgt}}|}
\sum_{t\in C_{\mathrm{tgt}}}
H_M(q^{\mathrm{CS}})[\ell,t,:].
\label{eq:canvas_tgt_anchor}
\end{equation}
The source canvas subsequently acts as the dynamic, in-context reference representation for aligning target-language hidden states.

\begin{table*}[t]
\centering
\small
\setlength{\tabcolsep}{2.1pt}
\renewcommand{\arraystretch}{1.02}
\begin{adjustbox}{max width=\textwidth}
\begin{tabular}{@{}l
!{\hspace{4pt}\color{GroupSep}\vrule width 0.3pt\hspace{4pt}}
cc
!{\hspace{4pt}\color{GroupSep}\vrule width 0.3pt\hspace{4pt}}
rrrc
!{\hspace{4pt}\color{GroupSep}\vrule width 0.3pt\hspace{4pt}}
rrrr@{}}
\toprule
& & &
\multicolumn{4}{c}{\textbf{Avg. F1}} &
\multicolumn{4}{c}{\textbf{$\Delta$F1 by CS condition}} \\
\cmidrule(lr){4-7} \cmidrule(lr){8-11}
\textbf{Model} &
\textbf{$L$} &
\textbf{$\mathcal{L}_{\mathrm{ctrl}}$} &
\textbf{\textsc{Base}} &
\textbf{Rand} &
\textbf{Adapt} &
\textbf{\textit{q}} &
\textbf{\textsc{GF-Src}} &
\textbf{\textsc{GF-Tgt}} &
\textbf{Random} &
\textbf{Selective} \\
\midrule

\modelwithlogo{meta}{Llama3.2-1B}
& 16 & $(11,15)$
& 2.57 & \underline{2.68} & \textbf{3.20} & $^{\ast\ast}$
& \canvasgain{+0.08} & \underline{\canvasgain{+0.82}} & \textbf{\canvasgain{+1.01}} & \canvasgain{+0.62} \\

\modelwithlogo{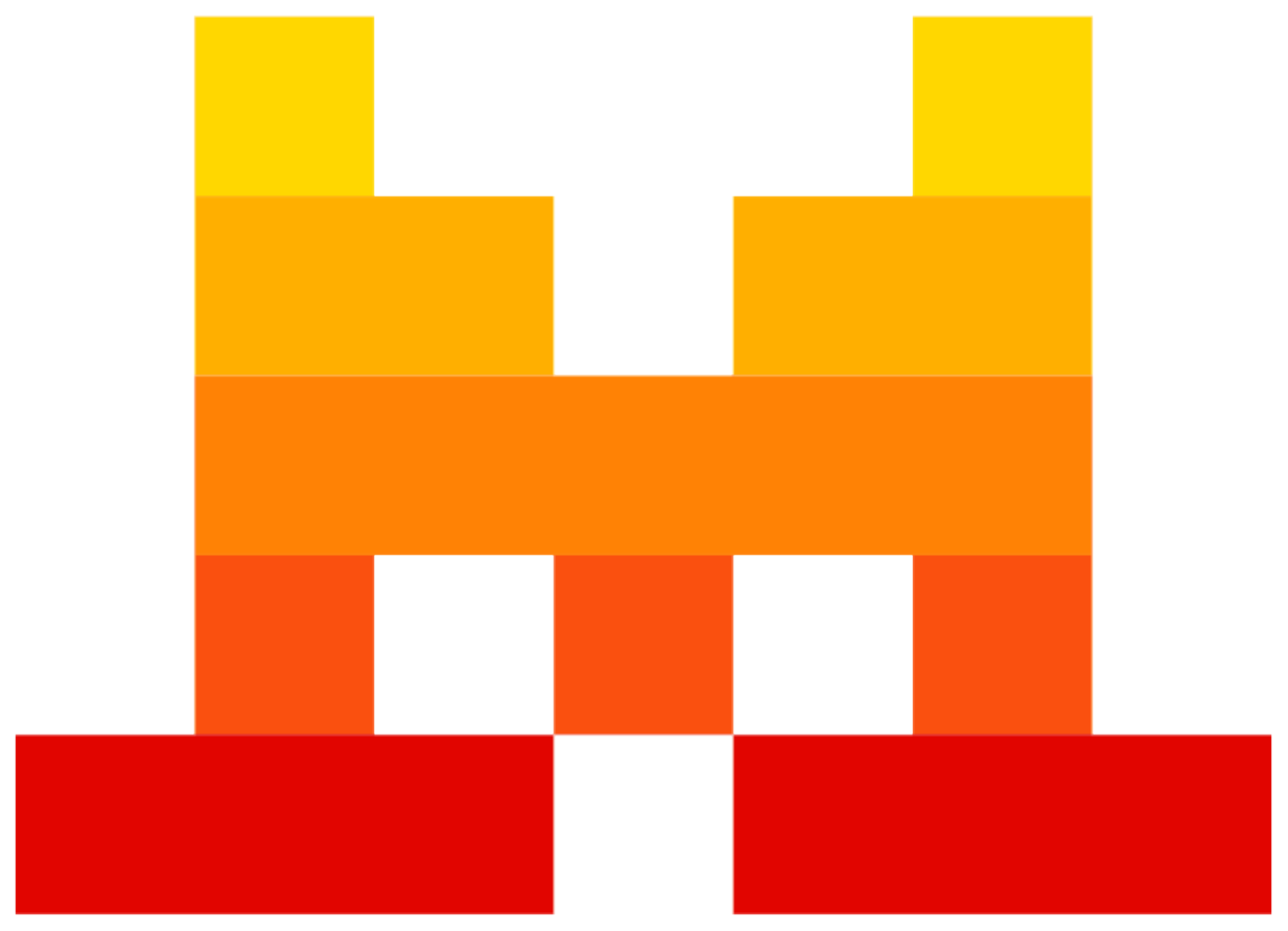}{Mistral-7B}
& 32 & $(22,31)$
& 4.53 & \underline{4.96} & \textbf{5.06} & $^{\ast\ast}$
& \canvasgain{+0.14} & \textbf{\canvasgain{+0.80}} & \underline{\canvasgain{+0.65}} & \canvasgain{+0.55} \\

\modelwithlogo{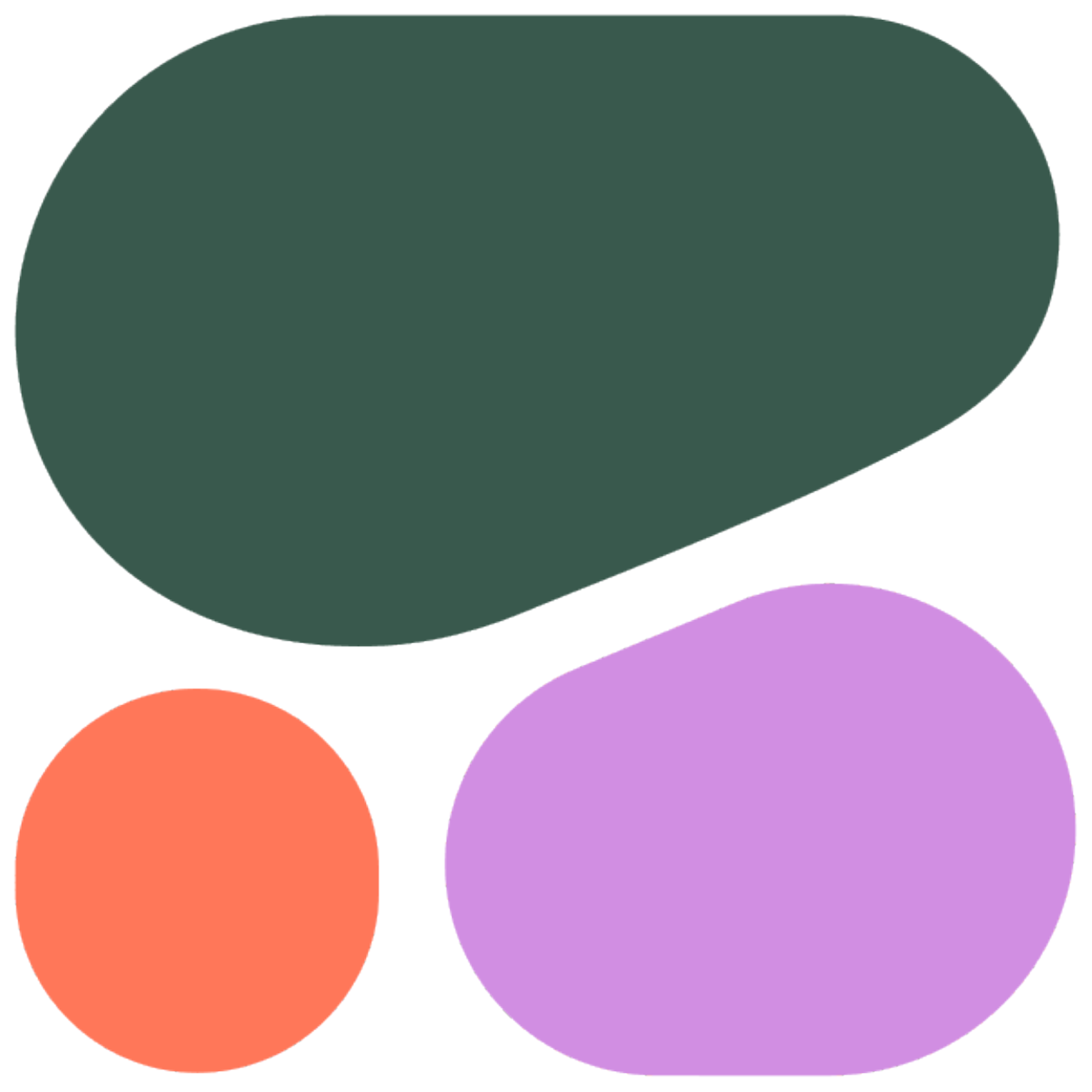}{Aya-Expanse-8B}
& 32 & $(22,31)$
& \underline{5.99} & \underline{5.99} & \textbf{6.88} & $^{\ast\ast}$
& \canvasgain{+0.62} & \textbf{\canvasgain{+1.41}} & \canvasgain{+0.63} & \underline{\canvasgain{+0.91}} \\

\modelwithlogo{meta}{Llama3.1-8B}
& 32 & $(22,31)$
& 6.21 & \underline{6.72} & \textbf{7.79} & $^{\ast\ast}$
& \canvasgain{+1.21} & \canvasgain{+0.64} & \underline{\canvasgain{+2.15}} & \textbf{\canvasgain{+2.34}} \\

\modelwithlogo{mistral}{Mixtral-8x7B}
& 32 & $(22,31)$
& 6.77 & \underline{6.82} & \textbf{6.91} & --
& \canvasgain{+0.16} & \underline{\canvasgain{+0.29}} & \textbf{\canvasgain{+0.38}} & \canvasloss{-0.29} \\

\modelwithlogo{qwen}{Qwen3-30B-A3B}
& 48 & $(33,47)$
& 6.91 & \underline{6.93} & \textbf{8.11} & $^{\ast\ast}$
& \canvasgain{+0.06} & \textbf{\canvasgain{+2.48}} & \canvasgain{+0.71} & \underline{\canvasgain{+1.55}} \\

\modelwithlogo{qwen}{Qwen3.5-4B}
& 36 & $(25,35)$
& \underline{7.58} & 7.35 & \textbf{8.14} & $^{\ast\ast}$
& \canvasgain{+0.37} & \underline{\canvasgain{+0.70}} & \canvasgain{+0.33} & \textbf{\canvasgain{+0.86}} \\

\modelwithlogo{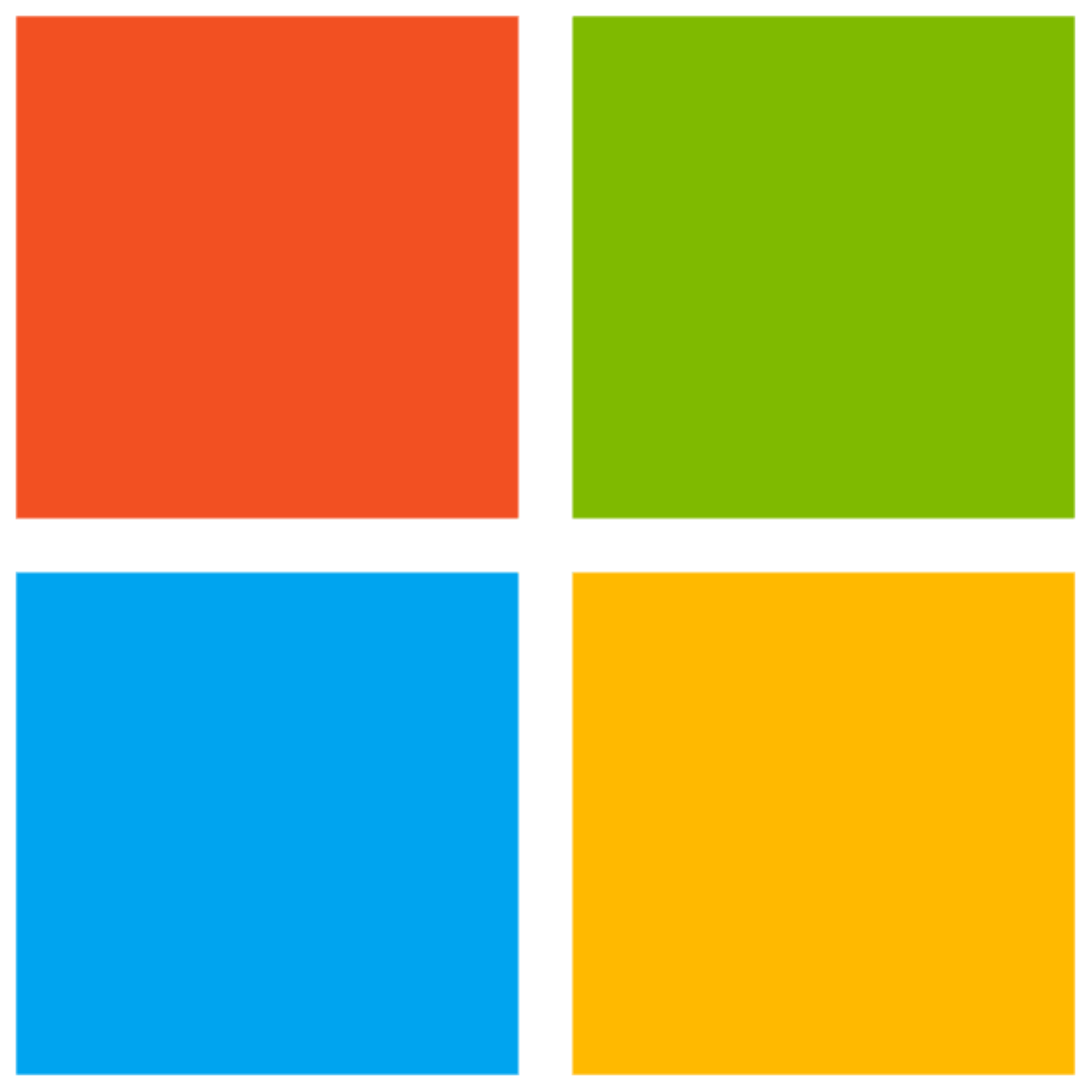}{Phi-3.5-MoE}
& 32 & $(22,31)$
& \underline{7.99} & 7.93 & \textbf{8.82} & $^{\ast\ast}$
& \canvasgain{+0.12} & \textbf{\canvasgain{+1.31}} & \canvasgain{+0.61} & \underline{\canvasgain{+1.26}} \\

\modelwithlogo{qwen}{Qwen3.5-9B}
& 40 & $(28,39)$
& 9.89 & \underline{10.01} & \textbf{10.24} & $^{\ast}$
& \canvasloss{-0.20} & \textbf{\canvasgain{+0.77}} & \canvasgain{+0.22} & \underline{\canvasgain{+0.61}} \\

\modelwithlogo{meta}{Llama3.1-70B}
& 80 & $(56,79)$
& 12.89 & \underline{12.98} & \textbf{13.20} & --
& \canvasloss{-0.52} & \underline{\canvasgain{+0.67}} & \textbf{\canvasgain{+0.73}} & \canvasgain{+0.35} \\

\modelwithlogo{meta}{Llama3.3-70B}
& 80 & $(56,79)$
& 12.96 & \underline{13.27} & \textbf{13.61} & $^{\ast}$
& \canvasgain{+0.29} & \textbf{\canvasgain{+1.66}} & \underline{\canvasgain{+0.61}} & \canvasgain{+0.04} \\

\modelwithlogo{qwen}{Qwen3.5-27B}
& 48 & $(33,47)$
& 14.14 & \underline{14.22} & \textbf{15.40} & $^{\ast\ast}$
& \canvasgain{+0.28} & \underline{\canvasgain{+1.74}} & \canvasgain{+0.96} & \textbf{\canvasgain{+2.03}} \\

\modelwithlogo{qwen}{Qwen3.6-35B-A3B}
& 40 & $(28,39)$
& 15.80 & \underline{16.07} & \textbf{17.53} & $^{\ast\ast}$
& \canvasgain{+0.66} & \textbf{\canvasgain{+3.58}} & \underline{\canvasgain{+1.41}} & \canvasgain{+1.28} \\

\bottomrule
\end{tabular}
\end{adjustbox}
\caption{\textbf{\textsc{CANVAS} intervention results.}
Avg. F1 is averaged over four CS conditions; condition-wise columns report Adapt--Base $\Delta$F1. Rand/Adapt denote random-direction/adaptive \textsc{CANVAS}. Bold/underline mark best/second-best values; $^{\ast}q<0.05$, $^{\ast\ast}q<0.01$ after BH--FDR correction.}
\label{tab:canvas_results}
\end{table*}

\paragraph{Stage 3: Canvas Alignment Scoring.}
To quantify the directional drift, we define the layer-wise \emph{canvas direction} as the normalized vector from the target anchor to the source canvas: $\mathbf{u}_{\ell}=(\mathbf{c}^{\mathrm{src}}_{\ell}-\mathbf{a}^{\mathrm{tgt}}_{\ell}) /\|\mathbf{c}^{\mathrm{src}}_{\ell}-\mathbf{a}^{\mathrm{tgt}}_{\ell}\|_2$. Let $C_{\mathrm{mix}}=C_{\mathrm{src}}\cup C_{\mathrm{tgt}}$. We compute the aggregated mixed representation $\mathbf{m}_{\ell}$ over all source- and target-language tokens:

\begin{equation}
\mathbf{m}_{\ell}
=
\frac{1}{|C_{\mathrm{mix}}|}
\sum_{t\in C_{\mathrm{mix}}}
H_M(q^{\mathrm{CS}})[\ell,t,:].
\label{eq:canvas_mixed_rep}
\end{equation}
The overall \emph{canvas alignment} score $\gamma(q^{\mathrm{CS}})$ is defined as the upper-layer average of the cosine similarity between the mixed representation and the canvas direction:
\begin{equation}
\gamma(q^{\mathrm{CS}})
=
\frac{1}{|\mathcal{L}_{\mathrm{ctrl}}|}
\sum_{\ell\in\mathcal{L}_{\mathrm{ctrl}}}
\cos\!\left(
\mathbf{m}_{\ell},
\mathbf{u}_{\ell}
\right).
\label{eq:canvas_alignment}
\end{equation}
A positive $\gamma$ means that the CS representation is aligned with the source canvas direction, while a negative $\gamma$ indicates target-side anchoring.

\paragraph{Stage 4: Adaptive Interpolation.}
Using the instance-specific score $\gamma_{\mathrm{CS}}=\gamma(q^{\mathrm{CS}})$, CANVAS dynamically calibrates the steering intensity $\alpha$:
\begin{equation}
\alpha
=
\operatorname{clip}_{[\alpha_{\min},\,\alpha_{\max}]}
\!\left(\alpha_0-\lambda\gamma_{\mathrm{CS}}\right),
\label{eq:canvas_alpha}
\end{equation}

where $\alpha_0$ represents the base steering intensity at neutral alignment, $\lambda>0$ controls the sensitivity to the alignment score, and $\alpha_{\min}, \alpha_{\max}$ define the clipped operating range. The negative-slope rule implements adaptive mitigation: if $\gamma_{\mathrm{CS}}>0$ (source-canvas aligned), the steering intensity decreases; if $\gamma_{\mathrm{CS}}<0$ (target-oriented), the steering intensity increases. 
This design applies stronger correction to more target-skewed CS states while keeping the update bounded under a single global schedule across all evaluated models and conditions. We provide hyperparameter validation and sensitivity analyses in Appendices~\ref{app:alpha_clip_validation} and~\ref{app:canvas_hyperparam_sensitivity}.
\paragraph{Stage 5: Online source-canvas interpolation.}
Following $\alpha$ determination, CANVAS initiates a second prefill pass using upper-layer activation hooks. For each target-language token position $t\in C_{\mathrm{tgt}}$ at layer $\ell\in\mathcal{L}_{\mathrm{ctrl}}$, we inject the source canvas via soft interpolation:

\begin{equation}
\widetilde{\mathbf{h}}_{\ell,t}
=
(1-\alpha)\,\mathbf{h}_{\ell,t}
+
\alpha\,\mathbf{c}^{\mathrm{src}}_{\ell}.
\end{equation}
This operation runs online, modifying only the latent vectors at target-token positions while leaving the surface text completely untouched. The adjusted hidden states are saved directly into the KV cache, from which the model greedily decodes the final answer.


\subsection{Intervention Protocol and Results}
\label{sec:intervention_results}
To evaluate the practical effectiveness of CANVAS, we compare its downstream CS performance against standard greedy decoding (\textbf{BASE}).
\paragraph{Evaluation protocol.}
We evaluate \textbf{BASE} decoding against two CANVAS configurations: (1) \textbf{Rand}, a baseline that injects an identical interpolation but utilizes a randomly sampled steering direction; and (2) \textbf{Adapt}, our full framework that dynamically sets the steering intensity $\alpha$ via the instance-specific score $\gamma$. Beyond the grammar-forced conditions (GF-SRC, GF-TGT), we expand our evaluation to the \emph{Random} and \emph{Selective} lexical switching conditions from CodeMixQA~\citep{winata2026can}, yielding four structural environments to test CANVAS's robustness.



\paragraph{Consistent QA Enhancement Across Conditions.}
As shown in \Cref{tab:canvas_results}, CANVAS (\textbf{Adapt}) improves average F1 across all evaluated MLLMs, supporting the view that source-canvas alignment can recover part of the CS performance loss.
The most substantial accuracy gains concentrate under the GF-TGT condition—the setting where our probing finds the strongest target-ward representation drift and the largest F1 degradation.
Furthermore, consistent gains under \textit{Random} and \textit{Selective} conditions demonstrate that source-canvas alignment operates as a generalizable linguistic correction mechanism. 
CANVAS further generalizes to a Spanish source anchor, shows significant paired gains for most models, and adds only modest overhead ($1.13\times$ latency, sub-second per question) without external model calls (Appendices~\ref{app:canvas_pivot_es}, \ref{sec:canvas-stats}, and~\ref{app:canvas_cost}). It therefore remains a lightweight method for broad CS settings. Model-wise variation in gain (e.g., Mixtral-8x7B, Llama-3.1-70B) mainly tracks each model's source-language \emph{oracle headroom} rather than scale or steering strength (Appendix~\ref{app:modelwise_canvas_repr}).

\begin{figure*}[t]
\centering
\includegraphics[width=0.72\textwidth]{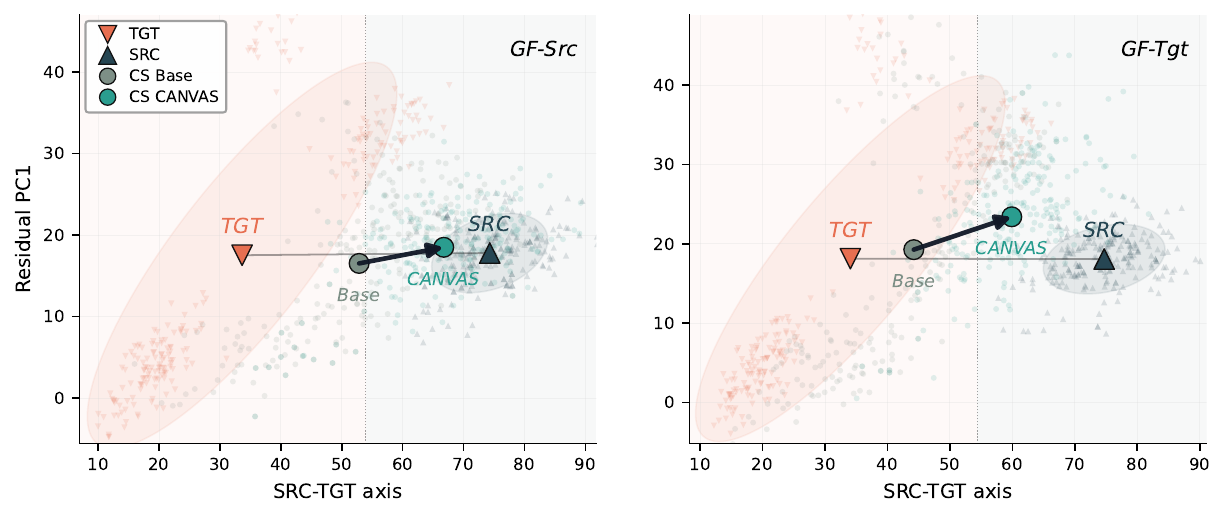}
\caption{\textbf{Latent displacement along the SRC--TGT axis.} We project token states onto a 2D subspace where the horizontal axis captures the global SRC--TGT direction. Faded points indicate individual instances, large markers denote condition centroids, and arrows show the directional shift induced by CANVAS.}

\label{fig:canvas_movement}
\end{figure*}

\begin{figure}[!hbp]
\centering
\includegraphics[width=0.88\linewidth]{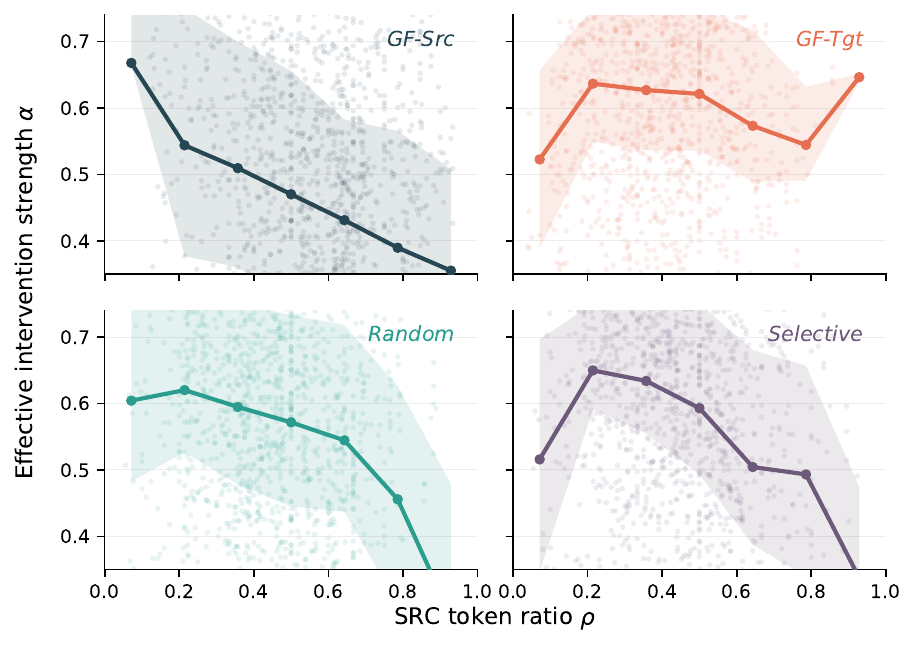}
\caption{\textbf{Adaptive intervention strength.}
CANVAS assigns larger interpolation strength $\alpha$ to more target-heavy inputs, measured by lower source-token ratio $\rho$.}
\label{fig:canvas_alpha}
\end{figure}

\paragraph{Comparison with Steering Baselines.}
Against ablated variants (fixed-$\alpha$, source-ratio controller, source-axis projection), controls (random or unrelated canvas), and the closest cross-lingual activation-steering methods \cite{kirtane2026language,mahmoud2026aligning,lim2025languagespecific}, CANVAS attains the highest CodeMixQA F1 (8.75 vs.\ 6.81 \textsc{Base}; Appendix~\ref{app:canvas_baselines}), and its advantage over the strongest baseline (mean-difference steering) is large and significant on higher-headroom multi-turn and RAG settings under paired McNemar tests (e.g., $72.0$ vs.\ $43.0$ accuracy on PingPong, $q<10^{-6}$; Appendix~\ref{app:canvas_rag_topic_llama}).
\subsection{Analysis and Geometric Diagnostics}

\label{sec:canvas_analysis}

\paragraph{Geometric Verification of Representational Steering.}
To check that the intervention produces the intended geometric correction, we visualize Qwen3.5-27B hidden states before and after steering, projected onto the SRC--TGT population vector (macro-average difference between clean source and target representations); the vertical axis is the first principal component of the residual after removing that direction. 
As shown in \Cref{fig:canvas_movement}, the representational centroids for both GF-SRC and GF-TGT undergo a substantial, rightward vector shift post-intervention, moving toward the clean source population space.
This geometric relocation supports the interpretation that CANVAS shifts the latent processing path toward the source-side anchor.

\paragraph{Adaptive Response to Structural Skew.}
To verify that the controller responds to the degree of drift estimated from internal representations, we aggregate all models and relate the instance-level source-token ratio $\rho$ to the runtime steering intensity $\alpha$. As illustrated in \Cref{fig:canvas_alpha}, GF-SRC queries keep a high source ratio and trigger mild alignment ($\rho=0.594$, $\gamma=-0.007$, $\alpha=0.453$), whereas target-heavy GF-TGT inputs elicit stronger adjustments ($\rho=0.404$, $\gamma=-0.139$, $\alpha=0.608$), without any manual thresholding. 
This monotonic pattern suggests that CANVAS allocates stronger intervention to inputs with greater target-ward drift, where source-side correction is more needed.

\paragraph{Disentangling Source Anchoring from Answer-Language Alignment.}
In the main setting the source language is also the required answer language, so the benefit of source-ward steering could reflect alignment with the \emph{output} language rather than correction of anchoring. We decouple the two by keeping English as the source/frame reference while requiring Qwen3.5-9B to answer in Korean or Spanish, scoring against gold answers translated into that language. CANVAS retains gains of $+0.41$ and $+0.49$ F1 in these non-English answer settings, comparable to the $+0.63$ and $+0.25$ obtained in the original English-answer setting for the same language pairs (\Cref{tab:app_answer_decoupling}, Appendix~\ref{app:language_canvas_mitigation}). Since the benefit persists when the answer language differs from the source, it follows source-frame alignment rather than output-language matching.

\paragraph{Robustness on Monolingual Inputs.}

CANVAS should also remain benign on monolingual or source-dominant inputs. The controller reduces the interpolation strength when the input shows little target-ward drift, making the update close to an identity operation; on pure monolingual and source-heavy inputs ($\rho_{\mathrm{SRC}}\ge0.80$) CANVAS preserves baseline performance (Appendix~\ref{app:source_heavy_checks}).

\begin{table*}[t]
\centering
\small
\setlength{\tabcolsep}{4pt}
\renewcommand{\arraystretch}{1.04}
\begin{adjustbox}{max width=\textwidth}
\begin{tabular}{@{}l rr rr c !{\hspace{4pt}\color{GroupSep}\vrule width 0.25pt\hspace{4pt}} rr rr c@{}}
\toprule
\multirow{3}{*}{\textbf{Model}} &
\multicolumn{5}{c}{\textbf{Bilingual CS}} &
\multicolumn{5}{c}{\textbf{Trilingual CS}} \\
\cmidrule(lr){2-6} \cmidrule(l){7-11}
& \multicolumn{2}{c}{\textsc{id-en}} &
  \multicolumn{2}{c}{\textsc{ha-en}} &
  \textbf{Bi.\ Mean} &
  \multicolumn{2}{c}{\textsc{jv-id-en}} &
  \multicolumn{2}{c}{\textsc{su-id-en}} &
  \textbf{Tri.\ Mean} \\
\cmidrule(lr){2-3} \cmidrule(lr){4-5} \cmidrule(lr){7-8} \cmidrule(lr){9-10}
& \textsc{Base} & \textsc{Canvas} & \textsc{Base} & \textsc{Canvas} & $\Delta$Acc &
  \textsc{Base} & \textsc{Canvas} & \textsc{Base} & \textsc{Canvas} & $\Delta$Acc \\
\midrule
\modelwithlogo{mistral}{Mistral-7B}        & 25.45 & \textbf{26.67} & 39.90 & \textbf{40.40} & \canvasgain{$+$0.86}          & \textbf{29.09} & 28.28 & 56.57 & 56.57 & \canvasloss{$-$0.41} \\
\modelwithlogo{meta}{Llama-3.1-8B}         & 20.81 & \textbf{23.64} & 60.61 & \textbf{65.66} & \canvasgain{$+$3.94}          & 37.37 & \textbf{43.64} & 78.38 & \textbf{79.39} & \canvasgain{$\mathbf{+3.64}$} \\
\modelwithlogo{cohere}{Aya-Expanse-8B}     & 16.16 & \textbf{18.59} & 44.44 & \textbf{50.00} & \canvasgain{$\mathbf{+3.99}$} & 28.28 & \textbf{30.91} & 84.65 & \textbf{85.25} & \canvasgain{$+$1.62} \\
\modelwithlogo{qwen}{Qwen3.5-27B}          & 25.25 & \textbf{28.28} & 90.40 & 90.40          & \canvasgain{$+$1.52}          & \textbf{59.60} & 58.38 & 86.46 & \textbf{86.67} & \canvasloss{$-$0.51} \\
\midrule
\rowcolor{MeanGray}
\textbf{\textit{Mean}} & 21.92 & \textbf{24.30} & 58.84 & \textbf{61.62} & \canvasgain{$+$2.58} & 38.59 & \textbf{40.30} & 76.52 & \textbf{76.97} & \canvasgain{$+$1.09} \\
\bottomrule
\end{tabular}
\end{adjustbox}
\caption{\textbf{\textsc{CANVAS} on PingPong multi-turn CS QA.} Five-option multiple-choice accuracy (\%) for \textsc{Base} and \textsc{Canvas}. Bi.\ / Tri.\ Mean report the average $\Delta$Acc for bilingual and trilingual CS settings.}
\label{tab:app_pingpong}
\end{table*}

\subsection{Generalization to Multi-Turn CS Dialog.}
We test whether CANVAS generalizes from controlled single-turn prompts to noisy, long-context multi-turn dialog using the answerable QA subset of PingPong~\citep{farhansyah2026pingpongnaturalbenchmarkmultiturn}, a human-authored multi-party benchmark covering Indonesian--English (id-en), Hausa--English (ha-en), Javanese--Indonesian--English (jv-id-en), and Sundanese--Indonesian--English (su-id-en).
Using the same 30\% upper-layer schedule, CANVAS transfers effectively to this setting (Table~\ref{tab:app_pingpong}).
It yields $+2.58$ accuracy on bilingual pairs (id-en, ha-en) across all tested architectures and, even under trilingual settings (jv-id-en, su-id-en) where a binary partition is lossy, a positive $+1.09$ Acc. 
Appendix~\ref{app:canvas_rag_topic_llama} further shows positive gains on retrieval-augmented QA and topic classification. 
In-context source alignment thus generalizes beyond controlled short-answer QA.

\section{Related Work}
\label{sec:related}
\subsection{Multilingual Consistency in LLMs}

Multilingual LLMs do not behave uniformly across languages, showing fairness and hallucination gaps, RAG failures, and cross-language preferences \cite{piqueras2022pretrained,ramesh2023fairness,chataigner2024multilingual,wu2024not,park2025investigating}, as well as uneven cross-lingual factual consistency \cite{qi-etal-2023-cross,xu2023language}. Closely related work suggests latent language preferences or English-centered internal processing \cite{wendler-etal-2024-llamas,zhao2024largelanguagemodelshandle,sharma-etal-2025-faux}. We extend this analysis to code-switched inputs, asking whether a single mixed-language question is internally anchored toward the source or target side.

\subsection{Code-Switching in LLMs}

Code-switching is a structured phenomenon shaped by grammatical, discourse, and interactional constraints \cite{dogruoz-etal-2021-survey,winata-etal-2023-decades}. Recent work uses CS to evaluate robustness, translation, generation, and reasoning \cite{laureano-de-leon-etal-2024-code,huzaifah2024evaluating,abdaljalil2025evaluating,mohamed2025lost,heredia2025conditioning,winata2026can}, and controlled CS probes reveal syntactic generalization, language-specific knowledge access, output-language alignment, and consistency effects \cite{sterner2025minimal,sterner2025code,kim-etal-2025-code,oh2026ola}. We follow this probing perspective but focus on internal anchoring and whether it predicts QA degradation.

Related work also steers non-English processing toward a high-performing language in activation space via learned or contrastive vectors \cite{mahmoud2026aligning,lim2025languagespecific} or source--target activation offsets \cite{kirtane2026language}. These fit corpus-level directions from parallel data and are not designed for mixed-language inputs, whereas CANVAS derives an instance-specific canvas from the CS input itself and scales the update by that instance's measured anchoring (Appendix~\ref{app:canvas_baselines}).

\section{Conclusion}
\label{sec:conclusion}
We study how MLLMs process code-switched inputs through internal language anchoring. Anchor bias measures whether a CS representation is closer to its source- or target-language counterpart, and grammatical frames consistently shift CS representations toward the corresponding anchor, robustly to token proportion and source language. CANVAS uses the same signal to softly align target-language token states toward an in-context source canvas, improving CS QA across models, conditions, and answer languages.

\section*{Limitations}
\label{sec:limitations}
Our analysis is limited in that it models code-switched processing through paired source and target anchors. This controlled axis enables matched comparisons with monolingual siblings and links representation shifts to downstream behavior, but it also compresses richer multilingual dynamics into a source--target contrast. Natural code-switching may involve local syntax, borrowed terms, named entities, discourse context, and language-specific factual familiarity that cannot be fully captured by a single anchor direction. We therefore treat anchor bias as a measurable and actionable component of CS processing, rather than a complete account of multilingual degradation.

\section*{Ethics Statement}
\label{sec:ethics}
We conduct all experiments using publicly available multilingual datasets and models that are distributed under documented research licenses, and we follow the usage terms set by the original providers. Our short-answer factoid QA setting is based on the CodeMixQA construction protocol, and our multi-turn QA setting is based on PingPong. Target-language counterparts are produced by a strong neural translator (Qwen3-235B-A22B), and grammar-forced code-switched variants are utilized following the CodeMixQA protocol from prior work. No new human-subject data are collected, and no personally identifying information is introduced by our pipeline.

\section*{Acknowledgement}
This work was supported by the Institute of Information \& Communications Technology Planning \& Evaluation (IITP) grant funded by the Korea government (MSIT) (No. RS-2021-II211341; Artificial Intelligence Graduate School Program at Chung-Ang University) and (No. RS-2026-25546026; Leading Generative AI Human Resources Development).
This work was also supported by the National Research Foundation of Korea (NRF) grant funded by the Korean government (MSIT) (RS-2026-25494299).

\bibliography{custom}

\clearpage
\appendix

\section{Appendix Table of Contents}
\label{app:toc}
\begin{enumerate}[leftmargin=*, itemsep=2pt, topsep=2pt]
    \item Section~\ref{app:llm} describes the use of LLMs during manuscript preparation.

    \item Section~\ref{app:implementation} provides \textbf{implementation details}, including the OpenRouter pipeline, local decoding setup, hidden-state extraction, quantization fallback, and hardware resources.

    \item Section~\ref{app:moe_routing} provides \textbf{complementary analyses for MoE models}, including MoE routing extraction, routing-profile similarity, routing configuration, Cohen's $d$ for routing separation, and the relationship between expert routing and representation-level anchor bias.

    \item Section~\ref{sec:es_pivot} tests \textbf{source-language generalization beyond English}, covering the Spanish-pivot anchor-bias replication, additional Arabic/Hindi/Korean/Chinese pivots, and Spanish-anchor \textsc{CANVAS} results.

    \item Section~\ref{app:dataset_details} describes the \textbf{dataset construction and matched comparison statistics}, caveats on translated target-language questions, additional code-switching conditions, and prompt templates.

    \item Section~\ref{app:anchor_bias_details} provides additional \textbf{anchor-bias} analyses, including anisotropy diagnostics, language-wise breakdowns, a within-language script control, metric variants, frontier-model scaling validation, same-lexicon order counterfactuals that disentangle lexical language from word order, and a token-proportion control.

    \item Section~\ref{app:canvas_details} provides detailed analyses of \textbf{CANVAS}, including the adaptive $\alpha$ schedule, layer-wise effects, source-heavy checks, token-level language tagging, hidden-state extraction, layer-span ablations, comparison with steering variants and baselines, language-wise mitigation and answer-language decoupling, case studies, model-wise movement and headroom diagnostics, cost analysis, additional movement analyses, cumulative recovery trajectories, statistical tests, detector sensitivity, hyperparameter sensitivity, and task extensions to retrieval-augmented QA, multi-turn dialog QA, and topic classification with paired comparisons against the strongest steering baseline.
\end{enumerate}

\section{The Use of Large Language Models}
\label{app:llm}
We write the manuscript ourselves, and an LLM (ChatGPT) is used solely for refinement---style, clarity, and grammar. It is not used for ideation or content generation.

\section{Implementation Details}
\label{app:implementation}

\paragraph{OpenRouter pipeline.}
Translation of SimpleQA questions and answer generation by the high-capacity reference model Qwen3-235B-A22B-2507~\cite{yang2025qwen3}, accessed via the OpenRouter OpenAI-compatible chat-completion API, with \texttt{DeepInfra} as the sole underlying provider for all calls (selected for consistent latency, to avoid provider switching, and to improve reproducibility). The maximum translation length is 256 tokens and the maximum answer length is 64 tokens; sampling temperature is set to $0$ (greedy).

\paragraph{Local model decoding.}
All dense models (Aya-Expanse-8B, Qwen3.5-4B/9B/27B, Llama-3.1-70B-Instruct, Llama-3.3-70B-Instruct) and all MoE models (Mixtral-8x7B-Instruct, Phi-3.5-MoE-Instruct, Qwen3-30B-A3B, Qwen3.6-35B-A3B) are run locally with deterministic greedy decoding. Specifically, we set \texttt{do\_sample=False} and use each model's default end-of-sequence token. For Qwen-3 family models, we set \texttt{enable\_thinking=False} where the tokenizer supports it. 

\paragraph{Hidden-state extraction.}
For representation analysis, we apply each model's chat template to the question, run a forward pass with \texttt{output\_hidden\_states=True}, and pool representations over the question-content region. We exclude shared prompt tokens (system instruction, assistant prefix, chat-template markers) from the pooled representation so that cosine-based comparisons focus on the question itself rather than on identical instruction text.

\paragraph{Quantization fallback.}
Models are loaded with automatic device mapping in bfloat16 where possible. If bf16 OOMs on the available hardware, we fall back to 4-bit NF4 quantization via \texttt{bitsandbytes} (used for all 70B-class dense models and the largest MoE checkpoints) or, for intermediate cases, to 8-bit. The CANVAS hidden-state interpolation is applied after dequantization at each control layer, so the intervention does not interact with the quantization scheme.

\paragraph{Hardware.}
All local inference runs use a single SLURM node with $4\times$ NVIDIA RTX 6000 Ada Generation GPUs (48~GB each), connected via PCIe. Per-model runtime is dominated by the prefill pass with hidden-state extraction. A full evaluation pass for one model and one CS condition completes within roughly $0.05$--$0.3$~h for models up to $8$B parameters, $0.4$--$0.8$~h for $27$--$35$B models, and $0.6$--$1.1$~h for $70$B-class dense models and the largest MoE checkpoints (Mixtral-8x7B), measured as the sum of per-example base, oracle, and CANVAS generation wall-clock times in the result logs.

\section{Code-Switching for MoE Models}
\label{app:moe_routing}

\subsection{Routing Similarity in MoE Models}
The representation-level results show where grammar-forced CS inputs are placed in the hidden-state space. However, hidden-state geometry is only one axis of how MLLMs process language. In mixture-of-experts (MoE) models, language processing can also be examined through the sparse computational pathway selected by the router: each token is assigned to a small subset of feed-forward experts. Recent work on multilingual MoE models shows that different languages can induce distinct routing patterns and partially specialized experts \cite{bandarkar2026multilingual,chen2026understanding,zheng2026unveilinglanguageroutingisolation}. This makes routing a complementary axis for analyzing language understanding in MLLMs. While prior analyses mainly focus on monolingual or language-separated inputs, CS provides a more fine-grained setting: a single question contains material from two languages, but its grammatical frame may be dominated by one of them. We therefore ask whether the language anchor observed in hidden-state geometry also appears in the model's expert-routing profile.

\begin{table*}[t]
\centering
\small
\setlength{\tabcolsep}{2.0pt}
\renewcommand{\arraystretch}{1.04}
\begin{adjustbox}{max width=\textwidth}
\begin{tabular}{l c c c c c c c c c c c c}
\toprule
\multirow{3}{*}{\raisebox{-1.3ex}{\textbf{Model}}} &
\multicolumn{6}{c}{\textbf{Cosine similarity over expert-frequency vectors}} &
\multicolumn{6}{c}{\textbf{Jaccard overlap over active expert sets}} \\
\cmidrule(lr){2-7} \cmidrule(lr){8-13}
&
\multicolumn{3}{c}{$\cos(\mathrm{SRC},\mathrm{CS})$} &
\multicolumn{3}{c}{$\cos(\mathrm{TGT},\mathrm{CS})$} &
\multicolumn{3}{c}{$\mathrm{Jacc}(\mathrm{SRC},\mathrm{CS})$} &
\multicolumn{3}{c}{$\mathrm{Jacc}(\mathrm{TGT},\mathrm{CS})$} \\
\cmidrule(lr){2-4} \cmidrule(lr){5-7}
\cmidrule(lr){8-10} \cmidrule(lr){11-13}
&
\cellcolor{SRCAnchor!18}\textsc{GF-Src} & \textsc{GF-Tgt} & $d$ &
\textsc{GF-Src} & \cellcolor{TGTAnchor!18}\textsc{GF-Tgt} & $d$ &
\cellcolor{SRCAnchor!18}\textsc{GF-Src} & \textsc{GF-Tgt} & $d$ &
\textsc{GF-Src} & \cellcolor{TGTAnchor!18}\textsc{GF-Tgt} & $d$ \\
\midrule
\modelwithlogo{qwen}{Qwen3-30B-A3B}
& 0.924 & 0.879 & +0.64
& 0.912 & 0.943 & -0.49
& 0.910 & 0.886 & +0.94
& 0.897 & 0.917 & -0.68 \\

\modelwithlogo{qwen}{Qwen3.6-35B-A3B}
& 0.933 & 0.905 & +0.63
& 0.925 & 0.944 & -0.40
& 0.646 & 0.586 & +0.56
& 0.596 & 0.653 & -0.53 \\

\modelwithlogo{microsoft}{Phi-3.5-MoE}
& 0.896 & 0.843 & +0.55
& 0.905 & 0.942 & -0.51
& 0.969 & 0.960 & +0.65
& 0.960 & 0.967 & -0.42 \\
\bottomrule
\end{tabular}
\end{adjustbox}
\caption{\textbf{Routing similarity under grammar-forced CS.}
We compare CS routing profiles against source and target routes. Highlighted columns indicate the frame-matched route.}
\label{tab:routing}
\end{table*}

\paragraph{Routing profiles.}
We collect top-$k$ expert selections from router or gate modules during the same forward pass for MoE models. We aggregate expert selections over the question-content region and compare whether the CS routing profile is closer to the source or target-language routing profile.
For each matched question set, we run separate forward passes for the source, CS, and target-language variants under the same prompt format. At each MoE layer, the router selects top-$k$ experts for every question-content token. Let $z_{\ell,t,j}(q)$ denote the $j$-th selected expert for token $t$ at layer $\ell$ when processing question variant $q$, where $j\in\{1,\dots,k\}$. We aggregate these selections into a layer-wise expert-frequency vector $\mathbf{f}_{\ell}(q)$, whose $e$-th component is

\vspace{-4mm}
\begingroup
\small
\begin{equation}
f_{\ell,e}(q)
=
\frac{1}{k|C(q)|}
\sum_{t\in C(q)}
\sum_{j=1}^{k}
\mathbb{I}\!\left[z_{\ell,t,j}(q)=e\right].
\label{eq:routing_freq}
\end{equation}
\endgroup
This vector records how often each expert is selected for a given input condition. Thus, source, CS, and target-language questions each induce their own routing-frequency profile.

\paragraph{Routing similarity.}
We compare each CS routing profile against the source and target-language routing profiles. First, we compute cosine similarity between expert-frequency vectors:
\begin{equation}
S^{\mathrm{cos}}_{\ell}(q_i,q_j)
=
\cos\!\left(\mathbf{f}_{\ell}(q_i),\mathbf{f}_{\ell}(q_j)\right).
\end{equation}
Second, we compute the Jaccard overlap between active expert sets. Let
\begin{equation}
A_{\ell}(q)=\{e \mid f_{\ell,e}(q)>0\}
\end{equation}
be the set of experts activated by question $q$ at layer $\ell$. We define
\begin{equation}
S^{\mathrm{jac}}_{\ell}(q_i,q_j)
=
\frac{|A_{\ell}(q_i)\cap A_{\ell}(q_j)|}
{|A_{\ell}(q_i)\cup A_{\ell}(q_j)|}.
\end{equation}
For each CS condition, we report its routing alignment with the source route, $S(q_s,q_c)$, and with the target route, $S(q_t,q_c)$, averaging each score over matched examples and MoE layers. We additionally report Cohen's $d$ as a standardized effect size between the \textsc{GF-Src} and \textsc{GF-Tgt} routing-score distributions; positive values indicate larger scores for \textsc{GF-Src}, while negative values indicate larger scores for \textsc{GF-Tgt}. Details are provided in Appendix~\ref{app:cohens_d}.

\paragraph{Routing shifts toward the frame language.}
\Cref{tab:routing} shows that the grammatical frame also shapes the sparse routing path selected by MoE models. Source-framed CS inputs are routed more like the source-language question: their expert-frequency vectors and active expert sets are closer to the SRC routing profile. Conversely, target-framed CS inputs are routed more like the target-language question, showing higher alignment with the TGT routing profile. This pattern holds consistently across both cosine similarity over expert-frequency vectors and Jaccard overlap over active expert sets. The Cohen's $d$ values further show that the two grammar-forced conditions induce clearly separated routing distributions. Beyond this directional pattern, routing behavior also aligns with the representation-level anchor signal. At the example level, routing bias, defined as source-route alignment minus target-route alignment, correlates strongly with anchor bias for cosine routing profiles ($\rho={+}0.703$, $p<0.001$) and moderately for Jaccard routing profiles ($\rho={+}0.464$, $p<0.001$). This suggests that expert-routing shifts track the same internal anchoring reflected in hidden states and are behaviorally relevant.

\subsubsection{MoE Routing Configuration}
\label{app:routing_config}

We report the routing configuration for the MoE models used in the expert-selection analysis. These details are important because routing similarity depends on the number of available experts, the number of experts selected per token, and the number of MoE layers from which routing traces are collected. Models with different expert counts and top-$k$ values can have different baseline overlap patterns, so these metadata help interpret the routing-similarity results in \Cref{tab:routing}.

\begin{table}[!htbp]
\centering
\small
\setlength{\tabcolsep}{4pt}
\renewcommand{\arraystretch}{1.05}
\begin{tabular}{lrrr}
\toprule
\textbf{Model} & \textbf{Experts} & top-$k$ & \textbf{Routing layers} \\
\midrule
\modelwithlogo{microsoft}{Phi-3.5-MoE}        & 16  & 2 & 32 \\
\modelwithlogo{qwen}{Qwen3-30B-A3B}           & 128 & 8 & 48 \\
\modelwithlogo{qwen}{Qwen3.6-35B-A3B}         & 256 & 8 & 40 \\
\bottomrule
\end{tabular}
\label{tab:routing_config_app}
\caption{\textbf{MoE routing configuration.}
We compute routing similarity from the top-$k$ expert selections over question-content tokens at each MoE layer.}
\end{table}

For each model, we collect expert selections over the question-content region after applying the same prompt format used in the representation analysis. We then aggregate these token-level selections into layer-wise expert-frequency vectors. This makes the routing analysis parallel to the hidden-state analysis: both compare matched EN, TGT, and CS variants over the same question region.

\subsection{Cohen's $d$ for Routing Separation}
\label{app:cohens_d}

We use Cohen's $d$ to quantify how strongly the routing-score distributions differ between \textsc{GF-Src} and \textsc{GF-Tgt}. For a given routing metric, let $a$ and $b$ denote the per-example routing scores under \textsc{GF-Src} and \textsc{GF-Tgt}, respectively. We compute
\begin{equation}
d
=
\frac{\mu(a)-\mu(b)}
{\sqrt{(\sigma_a^2+\sigma_b^2)/2}},
\end{equation}
where $\mu(\cdot)$ and $\sigma^2(\cdot)$ denote the mean and variance of each distribution. The sign of $d$ gives the direction of the effect: $d>0$ means that \textsc{GF-Src} has larger routing scores, while $d<0$ means that \textsc{GF-Tgt} has larger routing scores. The magnitude $|d|$ measures standardized separation; as a common rule of thumb, values around $0.2$, $0.5$, and $0.7$ are often interpreted as small, medium, and large effects. We use $d$ only as an effect-size summary, not as a statistical significance test.

\subsection{Expert Routing Mirrors Anchor Bias}
\label{app:routing_bias_anchor}

Table~\ref{tab:routing} in the main text shows that grammar-forced CS inputs route toward the frame-matched language route. Here, we provide a compact mechanism check linking this expert-level signal to representation-level anchor bias, without repeating the main routing table.

\begin{figure}[t]
  \centering
  \includegraphics[width=\linewidth]{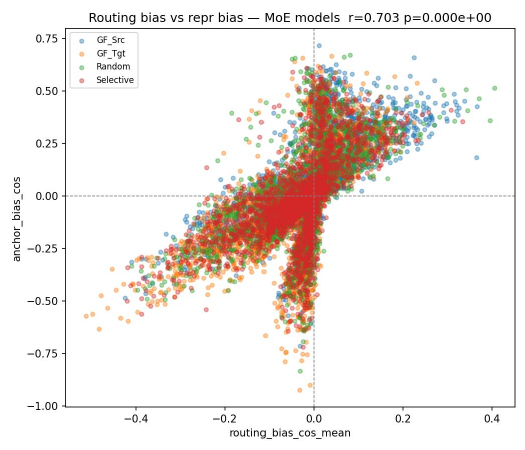}
  \caption{\textbf{Routing bias tracks representation anchor bias.}
  Each point is an example-level CS input from an MoE model. The horizontal axis reports SRC-vs-TGT expert-routing bias, and the vertical axis reports representation anchor bias. The positive association indicates that expert allocation shifts in the same SRC/TGT direction as representation-space anchor bias.}
  \label{fig:app_routing_similarity}
\end{figure}

Each point in Figure~\ref{fig:app_routing_similarity} is an example-level CS input from an MoE model. The $x$-axis is routing bias, computed as the SRC-vs-TGT difference in CS routing similarity, and the $y$-axis is representation anchor bias computed from hidden states. Positive values indicate SRC-oriented behavior, whereas negative values indicate TGT-oriented behavior. Thus, points in the upper-right and lower-left quadrants indicate examples where expert routing and representation geometry agree in their language orientation. The positive association suggests that language anchoring is reflected not only in hidden-state geometry but also in sparse expert selection.

\section{Source-Language Generalization Beyond English}
\label{sec:es_pivot}

\subsection{Anchor Bias}
\paragraph{Experimental setup.}
We replace English with Spanish as the source language and build eight CS pairs: \textsc{es}$\leftrightarrow$\{Toba Batak, Bengali, French, Hindi, Korean, Marathi, Urdu, Chinese\}. Questions, system prompts, and gold answers are in Spanish; CS inputs follow the same GF-Src / GF-Tgt grammar-forcing protocol as the main experiments. We evaluate Aya-Expanse-8B, Qwen3.5-4B, Qwen3.5-9B, and Qwen3.5-27B. Anchor bias uses the same upper-half aggregation and per-layer random cosine normalization as Table~\ref{tab:main_contrast}.

\paragraph{Results.}
Table~\ref{tab:es_pivot} shows the same frame-dependent split with Spanish as the source. GF-Src is consistently ES-anchored ($+0.213$ to $+0.358$), while GF-Tgt is consistently TGT-anchored ($-0.190$ to $-0.337$). The mean gap is $+0.541$, larger than the English-pivot upper-half gap of $+0.384$ computed over the same four models. QA follows the same ordering: GF-Src has a smaller mean F1 drop than GF-Tgt ($-3.4$ vs.\ $-5.9$ pp), and SRC-anchored layer ratios are much higher for GF-Src (${\sim}82\%$) than GF-Tgt (${\sim}35\%$).

\paragraph{Interpretation.}
The Spanish pivot rules out an English-specific explanation: the model tracks the language that supplies the grammatical frame, not English identity itself. The effect is also not weakened when the source language is less dominant in pre-training. We therefore interpret anchor bias as a general source-language anchoring phenomenon: upper-layer representations follow the grammatical backbone of the input, whether that source is English or Spanish. MoE routing diagnostics are not repeated for this pivot.

\paragraph{Additional pivots.}
Because Spanish is high-resource and shares the Latin script with several targets, we also use Arabic, Hindi, Korean, and Chinese as source/frame-reference languages (\Cref{tab:cr_pivot_langs} in the main text). For each pivot we translate the source-language tokens with Qwen3-235B-A22B while keeping the same QA instances and GF-Src/GF-Tgt construction, and evaluate Aya-Expanse-8B. The GF-Src$-$GF-Tgt separation stays positive for every pivot ($+0.29$ Arabic, $+0.32$ Hindi, $+0.28$ Korean, $+0.36$ Chinese), so the pattern is not specific to English, Spanish, or Latin-script sources.

\begin{table*}[t]
\centering
\small
\setlength{\tabcolsep}{2.7pt}
\renewcommand{\arraystretch}{1.06}
\begin{adjustbox}{max width=\textwidth}
\begin{tabular}{lccc!{\hspace{4pt}\color{GroupSep}\vrule width 0.25pt\hspace{4pt}}ccc!{\hspace{4pt}\color{GroupSep}\vrule width 0.25pt\hspace{4pt}}ccc}
\toprule
\multirow{2}{*}{\raisebox{-0.55ex}{\textbf{Model}}} &
\multicolumn{3}{c}{$\mathrm{AB}^{\mathrm{upper}}_{\mathrm{nrm}}$\;(\textit{ES as source})} &
\multicolumn{3}{c}{\textbf{QA behavior (F1)}} &
\multicolumn{3}{c}{\textbf{SRC-anchored layers (\%)}} \\
\cmidrule(lr){2-4} \cmidrule(lr){5-7} \cmidrule(lr){8-10}
& \textsc{GF-Src} & \textsc{GF-Tgt} & $\Delta_{\mathrm{AB}}$
& \textsc{SRC} & \textsc{GF-Src} & \textsc{GF-Tgt}
& \textsc{GF-Src} & \textsc{GF-Tgt} & $\Delta_{\mathrm{layer}}$ \\
\midrule
\modelwithlogo{cohere}{Aya-Expanse-8B}
& \srcanchor{27}{+0.358} & \tgtanchor{25}{-0.337} & \gapcell{+0.695}
& 14.8 & \fdrop{11.2}{-3.6} & \fdrop{8.3}{-6.5}
& 83.8 & 40.5 & +43.3 \\
\modelwithlogo{qwen}{Qwen3.5-4B}
& \srcanchor{20}{+0.261} & \tgtanchor{19}{-0.256} & \gapcell{+0.517}
& 7.4 & \fdrop{4.7}{-2.7} & \fdrop{2.5}{-4.9}
& 79.8 & 32.7 & +47.1 \\
\modelwithlogo{qwen}{Qwen3.5-9B}
& \srcanchor{21}{+0.281} & \tgtanchor{20}{-0.270} & \gapcell{+0.550}
& 12.4 & \fdrop{9.5}{-2.9} & \fdrop{6.5}{-5.9}
& 80.7 & 33.5 & +47.2 \\
\modelwithlogo{qwen}{Qwen3.5-27B}
& \srcanchor{16}{+0.213} & \tgtanchor{14}{-0.190} & \gapcell{+0.403}
& 17.1 & \fdrop{12.7}{-4.4} & \fdrop{10.7}{-6.4}
& 82.4 & 35.0 & +47.4 \\
\midrule
\rowcolor{MeanGray}
\textbf{\textit{Mean}}
& \srcanchor{21}{+0.278} & \tgtanchor{20}{-0.263} & \gapcell{+0.541}
& 12.9 & \fdrop{9.5}{-3.4} & \fdrop{7.0}{-5.9}
& 81.7 & 35.4 & +46.3 \\
\bottomrule
\end{tabular}
\end{adjustbox}
\caption{\textbf{Source-language anchor bias with Spanish (ES) as source.}
Metrics follow Table~\ref{tab:main_contrast}: $\mathrm{AB}^{\mathrm{upper}}_{\mathrm{nrm}}$ averages normalized anchor bias over the upper half of layers, and SRC-anchored layers count the fraction of layers with $\mathrm{AB}_{\mathrm{nrm}}>0$.}
\label{tab:es_pivot}
\end{table*}

\subsection{CANVAS}
\label{app:canvas_pivot_es}

\paragraph{Setup.} We test whether \textsc{CANVAS} generalizes beyond English as the source anchor. We construct a Spanish-pivot CS dataset by pairing each TGT language with Spanish in place of English and translating the English gold answers to Spanish via Qwen3.6-35B-A3B. We evaluate four conditions per example: \textsc{ES-oracle} (pure-Spanish question, upper bound), \textsc{Base} (Spanish--TGT CS, no intervention), \textsc{CANVAS} (CS with steering toward the Spanish anchor), and \textsc{TGT-oracle} (pure-TGT question, lower bound). We pair Spanish with eight TGT languages, giving $1{,}280$ examples per model ($8$ pairs $\times$ $2$ CS conditions $\times$ $80$ items).

\paragraph{Pipeline.} We keep the adaptive interpolation schedule used in the main runs ($\alpha=\textrm{clip}_{[0.05,0.75]}(0.45-1.5\gamma)$, $30\%$ layer budget) and replace only the SRC anchor with Spanish. For non-Latin TGT scripts (Bengali, Devanagari, Korean, Chinese), we tag tokens by character-level script: Latin-script tokens are treated as SRC and non-Latin tokens as TGT. For Latin-script TGT languages (French, Toba Batak), we fall back to the \texttt{lingua} language detector. The system prompt declares Spanish as the answer language.

\paragraph{Result.} We report results in Table~\ref{tab:app_canvas_pivot_es}. We observe a positive $\Delta\textrm{F1}_{\textsc{CANVAS}}$ on all five models, with a mean improvement of $+0.67$ F1 over direct CS answering. Llama-3.1-8B benefits the most ($+1.47$), and the two smaller Qwen variants gain $+0.75$ and $+0.85$. Even Mistral-7B ($+0.15$) and Qwen3.5-27B ($+0.13$) show no regression and stay well above their corresponding TGT-oracle baselines.

\begin{table*}[t]
\centering
\small
\setlength{\tabcolsep}{5pt}
\renewcommand{\arraystretch}{1.04}
\begin{tabular}{@{}lrrrrr@{}}
\toprule
\textbf{Model} & \textbf{ES-ora} & \textbf{Base} & \textbf{CANVAS} & \textbf{TGT-ora} & $\Delta\mathrm{F1}_{\mathrm{CANVAS}}$ \\
\midrule
\modelwithlogo{meta}{Llama3.1-8B}   & 10.54 &  7.06 &  8.53 & 2.57 & \canvasgain{+1.47} \\
\modelwithlogo{mistral}{Mistral-7B} &  8.00 &  3.46 &  3.61 & 1.16 & \canvasgain{+0.15} \\
\modelwithlogo{qwen}{Qwen3.5-4B}    &  7.36 &  3.60 &  4.35 & 2.17 & \canvasgain{+0.75} \\
\modelwithlogo{qwen}{Qwen3.5-9B}    & 12.39 &  8.03 &  8.88 & 3.85 & \canvasgain{+0.85} \\
\modelwithlogo{qwen}{Qwen3.5-27B}   & 17.10 & 11.68 & 11.81 & 9.17 & \canvasgain{+0.13} \\
\midrule
\rowcolor{MeanGray}
\textbf{\textit{Mean}} & 11.08 & 6.77 & 7.44 & 3.78 & \canvasgain{+0.67} \\
\bottomrule
\end{tabular}
\caption{\textbf{CANVAS with a Spanish anchor.}
F1 (\%) on the Spanish-pivot CS dataset. ES-ora and TGT-ora are pure-Spanish and pure-target baselines, respectively. \textsc{CANVAS} improves over direct CS answering for all five models, showing that the source-anchor intervention is not English-specific.}
\label{tab:app_canvas_pivot_es}
\end{table*}

\paragraph{Script-dependent behaviour.} We find a clear script-dependent pattern when aggregating per-pair gains across the five models. The six non-Latin TGT pairs (\textsc{ben, hin, kor, mar, urd, zho}) all gain $\Delta\textrm{F1}$ in the range $+0.52$ to $+1.35$ (mean $+0.95$), while the two Latin-script TGT pairs (\textsc{bbc-esp} $+0.03$ and \textsc{fra-esp} $-0.37$) average $-0.17$. This pattern is consistent with the tagger design: script-based tagging is unambiguous for non-Latin TGT languages, whereas the \texttt{lingua} fallback used for Latin-script TGT languages can struggle to separate Spanish from morphologically similar Romance-language tokens, most notably French. The non-Latin sub-pool, therefore, provides the cleanest evaluation of the language-agnostic claim, and its Spanish-anchor gains match the magnitude of the English-anchor \textsc{CANVAS} improvements reported in the main paper.

\paragraph{Takeaway.} \textsc{CANVAS} produces positive $\Delta\textrm{F1}$ across all five evaluated models when Spanish replaces English as the source anchor, with the strongest effect on non-Latin TGT languages where script-based tagging is most reliable. This result argues against the alternative explanation that the English-anchor gains depend on a privileged role of English in the representation space: the same steering mechanism transfers to another high-resource source language with only the runtime language-detection backend changed. The Latin-script TGT sub-pool further suggests that the main practical bottleneck for non-English anchors is SRC/TGT tagging precision, rather than the steering mechanism itself.

\section{Dataset Details}
\label{app:dataset_details}
\subsection{Dataset Statistics}
\label{app:dataset_statistics}

We provide dataset statistics to clarify the scale and balance of the controlled CS evaluation. Our setup starts from English SimpleQA questions and pairs each question with a target-language counterpart and a code-switched variant. Because each CS example is evaluated together with its matched EN and TGT forms, the resulting evaluation is organized around EN--TGT--CS comparisons rather than independent monolingual examples.

\begin{table}[!htbp]
\centering
\small
\setlength{\tabcolsep}{4pt}
\renewcommand{\arraystretch}{1.05}
\begin{tabular}{lr}
\toprule
\textbf{Item} & \textbf{Count} \\
\midrule
Unique English source questions & 955 \\
Target languages & 9 \\
CS examples per language & 320 \\
Examples per language--condition & 80 \\
Matched EN--TGT--CS comparisons & 2,880 \\
CS conditions & 4 \\
Evaluated variants per CS example & 3 \\
Total evaluated input instances & 8,640 \\
\bottomrule
\end{tabular}
\caption{\textbf{Dataset statistics.}
Each CS example is evaluated with matched EN, TGT, and CS inputs under the same reference answer. The four CS conditions are balanced within each language pair.}
\label{tab:dataset_statistics}
\end{table}

As shown in \Cref{tab:dataset_statistics}, each target language contributes 320 CS examples, with 80 examples for each CS condition. Since each CS example is evaluated with its matched English and target-language counterparts, the full evaluation contains 8,640 input instances. This balanced construction ensures that differences between \textsc{GF-Src} and \textsc{GF-Tgt} are not driven by unequal language or condition coverage. We use nine target languages: Bengali, Spanish, French, Hindi, Korean, Marathi, Toba Batak, Urdu, and Chinese.

\subsection{Caveat on Translated Target Questions}
\label{app:translation_caveat}

In our matched set $\mathcal{Q}(x)=\{q^{\mathrm{SRC}},q^{\mathrm{TGT}},q^{\mathrm{GF\text{-}SRC}},q^{\mathrm{GF\text{-}TGT}}\}$, $q^{\mathrm{SRC}}$ and the two grammar-forced CS variants come from CodeMixQA~\citep{winata2026can}, but $q^{\mathrm{TGT}}$ is generated by Qwen3-235B-A22B from $q^{\mathrm{SRC}}$ using the prompt in \Cref{app:prompts}. For short, factoid SimpleQA queries (e.g., \textit{``Who wrote Pride and Prejudice?''}~$\to$~\textit{``누가 오만과 편견을 썼나요?''}) the translation direction is essentially deterministic and the resulting $q^{\mathrm{TGT}}$ has the same syntactic frame as $q^{\mathrm{GF\text{-}TGT}}$. For more complex queries, however, the translator can pick a paraphrase whose syntactic frame differs from the CodeMixQA CS construction (e.g., choosing an active vs.\ passive frame, or a different connective ordering). Because of this, we treat $q^{\mathrm{TGT}}$ as a target-language reference for representation-level anchor comparison rather than as a strict word-order-matched twin of the CS variants; all behavioral claims in the paper are reported either per condition or as deltas relative to $q^{\mathrm{SRC}}$, so they are unaffected by this mismatch. The cosine-based anchor-bias measure is also robust to surface-syntactic variation because it pools over question-content tokens within each variant separately rather than aligning tokens across variants. 

\subsection{Additional Code-Switching Conditions}
\label{app:additional_cs}

In addition to the two grammar-forced conditions used in the main analysis, we also examine random and selective code-switching conditions from CodeMixQA~\cite{winata2026can}. These conditions provide useful controls because they introduce bilingual lexical mixing without imposing the same explicit grammatical-frame contrast as \textsc{GF-Src} and \textsc{GF-Tgt}. Random switching replaces tokens without a strong linguistic selection criterion, while selective switching targets more informative lexical items. As a result, these conditions help separate the general effect of lexical mixing from the more specific effect of grammatical frame.

\begin{table}[!htbp]
\centering
\small
\setlength{\tabcolsep}{4pt}
\renewcommand{\arraystretch}{1.05}
\begin{tabular}{lrr}
\toprule
\textbf{Condition} & \textbf{Anchor bias} & $\boldsymbol{\Delta}$\textbf{F1} \\
\midrule
\textsc{Random} & $-0.026$ & $-5.19$ \\
\textsc{Selective}  & $-0.103$ & $-4.37$ \\
\bottomrule
\end{tabular}
\caption{\textbf{Additional CS controls.}
Random and selective switching degrade performance even without imposing the same grammar-frame contrast as \textsc{GF-Src}/\textsc{GF-Tgt}. These controls show that arbitrary lexical mixing can be harmful, while the grammar-forced pair isolates the stronger structural anchoring effect.}
\label{tab:additional_cs_controls}
\end{table}

As shown in \Cref{tab:additional_cs_controls}, both random and selective switching reduce QA performance, confirming that code-switched lexical mixing itself can be challenging for MLLMs. However, their anchor-bias values remain much closer to zero than the grammar-forced contrast reported in the main results. This suggests that arbitrary or lexically selective mixing does not by itself create the strong directional anchoring observed under \textsc{GF-Src} and \textsc{GF-Tgt}. Instead, the main contrast is driven by the grammatical frame that organizes the mixed-language question.

\subsection{Prompt Templates}
\label{app:prompts}

We provide the prompt templates used for constructing target-language questions and generating short answers. The templates are intentionally restrictive: the translation prompt preserves the original information needed, and the answer prompt asks for a short English answer without explanations.

\paragraph{Target-language question construction.}
For each English SimpleQA question, we use Qwen3-235B-A22B to obtain a target-language counterpart. The prompt asks the model to preserve the original information needed and output only the translated question.

\begin{tcolorbox}[promptbox,title={Target-language question construction}]
\role{System} You are a professional translator.

\vspace{1mm}
\role{User} Translate the following English question into \{target language\}.

\vspace{1mm}
\textbf{Constraints}
\begin{itemize}[leftmargin=*,nosep]
    \item Preserve the meaning exactly.
    \item Preserve named entities unless they have a standard form in the target language.
    \item Do not add explanations.
    \item Output only the translated question.
\end{itemize}

\vspace{1mm}
\textbf{Input question}\\
\{question\}
\end{tcolorbox}

\paragraph{English short-answer generation.}
For answer generation, the model receives one question at a time and is instructed to return a short English answer. We use deterministic decoding and a maximum output length of 64 tokens.

\begin{tcolorbox}[promptbox,title={English short-answer generation}]
\role{System} You are a short-answer QA system. Return only the final answer as a short phrase. Answer must be in English. No explanation.

\vspace{1mm}
\role{User} \{question\}
\end{tcolorbox}

\section{Anchor Bias Details}
\label{app:anchor_bias_details}
\subsection{Anisotropy Diagnostics}
\label{app:anisotropy}

Cosine similarities in hidden-state spaces can be inflated by representation anisotropy: even unrelated inputs may occupy similar directions and therefore show high background cosine similarity. This is especially important for cross-model comparisons, because different model families can have very different random-cosine baselines. To quantify this effect, we estimate a random baseline cosine for each model from unrelated question pairs.

\begin{table}[!htbp]
\centering
\small
\setlength{\tabcolsep}{5.0pt}
\renewcommand{\arraystretch}{1.05}
\begin{tabular}{llrr}
\toprule
\textbf{Model} & \textbf{Type} & \textbf{Random cos.} & $\mathbf{1-\mu}$ \\
\midrule
\modelwithlogo{cohere}{Aya-Expanse-8B}        & Dense & 0.7944 & 0.2056 \\
\modelwithlogo{meta}{Llama3.3-70B}            & Dense & 0.8729 & 0.1271 \\
\modelwithlogo{meta}{Llama3.1-70B}            & Dense & 0.8775 & 0.1225 \\
\modelwithlogo{qwen}{Qwen3.5-4B}              & Dense & 0.9112 & 0.0888 \\
\modelwithlogo{qwen}{Qwen3.5-27B}             & Dense & 0.9145 & 0.0855 \\
\modelwithlogo{qwen}{Qwen3.5-9B}              & Dense & 0.9320 & 0.0680 \\
\midrule
\modelwithlogo{mistral}{Mixtral-8x7B}         & MoE   & 0.8862 & 0.1138 \\
\modelwithlogo{microsoft}{Phi-3.5-MoE}        & MoE   & 0.8698 & 0.1302 \\
\modelwithlogo{qwen}{Qwen3.6-35B-A3B}         & MoE   & 0.9379 & 0.0621 \\
\modelwithlogo{qwen}{Qwen3-30B-A3B}           & MoE   & 0.9791 & 0.0209 \\
\bottomrule
\end{tabular}
\caption{\textbf{Random baseline cosine for anisotropy correction.}
We estimate the average cosine similarity between representations of unrelated questions for each model. High random cosine values indicate strong background alignment in the hidden-state space; $1-\mu$ is the denominator used in the random cosine normalization.}
\label{tab:anisotropy_baseline}
\end{table}

As shown in Table~\ref{tab:anisotropy_baseline}, random baseline cosine values are high across all models and exceed $0.90$ for several Qwen-family models. This means that raw cosine similarity can overstate representational closeness even for unrelated inputs. We therefore subtract the model- and layer-specific random baseline before comparing EN--CS and TGT--CS similarities in the main anchor-bias. We use a random-pair baseline rather than the SRC--TGT cosine because SRC--TGT cosine is itself a content-anchored quantity we want to measure relative to chance; layer-wise min--max scaling would similarly entangle normalization with the very anchor distance under study.

\subsection{Language-wise Anchor Bias}
\label{app:language_anchor_bias}
\paragraph{Setup.}
We group the matched SRC/CS/TGT triples by target language and CS condition, then compute the same normalized anchor-bias metric used in Table~\ref{tab:main_contrast}. For each language--condition cell, we average over examples and models. We also report CS F1 in the same point scale as the main tables so that representation movement and QA behavior can be read together.

\begin{table*}[t]
\centering
\small
\setlength{\tabcolsep}{2.4pt}
\renewcommand{\arraystretch}{1.02}
\begin{adjustbox}{max width=\textwidth}
\begin{tabular}{@{}l!{\hspace{4pt}\color{GroupSep}\vrule width 0.25pt\hspace{4pt}}rrrr!{\hspace{4pt}\color{GroupSep}\vrule width 0.25pt\hspace{4pt}}rrrr@{}}
\toprule
\multirow{2}{*}{\raisebox{-0.55ex}{\textbf{Language}}} &
\multicolumn{4}{c}{\textbf{$\mathrm{AB}^{\mathrm{upper}}_{\mathrm{nrm}}$ by CS condition}} &
\multicolumn{4}{c}{\textbf{CS F1 by CS condition}} \\
\cmidrule(lr){2-5} \cmidrule(lr){6-9}
& \textsc{GF-Src} & \textsc{GF-Tgt} & Random & Selective & \textsc{GF-Src} & \textsc{GF-Tgt} & Random & Selective \\
\midrule
Toba Batak & \canvasgain{+4.40} & \canvasgain{+4.00} & \canvasgain{+4.03} & \canvasgain{+4.02} & 10.1 & 10.6 & 8.9 & 12.6 \\
Hindi & \canvasgain{+2.30} & \canvasgain{+0.53} & \canvasgain{+1.11} & \canvasgain{+0.57} & 15.2 & 14.1 & 16.0 & 15.4 \\
Chinese (Mandarin) & \canvasgain{+1.00} & \canvasloss{-1.65} & \canvasloss{-0.05} & \canvasloss{-0.21} & 13.9 & 9.0 & 13.7 & 16.1 \\
Spanish & \canvasgain{+0.21} & \canvasloss{-1.22} & \canvasloss{-0.58} & \canvasloss{-0.78} & 15.8 & 12.2 & 16.7 & 16.5 \\
Urdu & \canvasgain{+0.57} & \canvasloss{-1.56} & \canvasloss{-0.74} & \canvasloss{-0.77} & 12.8 & 13.5 & 13.1 & 14.8 \\
French & \canvasgain{+0.18} & \canvasloss{-1.25} & \canvasloss{-0.91} & \canvasloss{-0.96} & 13.5 & 12.7 & 10.3 & 9.9 \\
Bengali & \canvasloss{-0.28} & \canvasloss{-2.04} & \canvasloss{-1.06} & \canvasloss{-1.51} & 8.9 & 9.1 & 13.6 & 8.9 \\
Korean & \canvasloss{-0.26} & \canvasloss{-1.97} & \canvasloss{-1.68} & \canvasloss{-1.44} & 15.6 & 8.6 & 10.5 & 11.0 \\
Marathi & \canvasloss{-0.29} & \canvasloss{-2.16} & \canvasloss{-1.27} & \canvasloss{-1.68} & 13.6 & 12.0 & 10.0 & 15.3 \\
\midrule
\rowcolor{MeanGray}
\textbf{\textit{Mean}} & \canvasgain{+0.87} & \canvasloss{-0.81} & \canvasloss{-0.13} & \canvasloss{-0.31} & 13.3 & 11.3 & 12.5 & 13.4 \\
\bottomrule
\end{tabular}
\end{adjustbox}
\caption{\textbf{Language-wise anchor bias by CS condition.} We group examples by target language and CS condition and report the same normalized anchor-bias family used in Table~\ref{tab:main_contrast}. F1 is shown on the same point scale as the main tables.}
\label{tab:app_language_anchor_bias}
\end{table*}

\paragraph{Results.}
Table~\ref{tab:app_language_anchor_bias} shows that the anchor signal is not only a model-level effect. The GF-Src column is generally more SRC-oriented than GF-Tgt, but the strength of this split differs by language. This language-level variation explains why the aggregate AB--F1 pattern should be interpreted as a structured multilingual phenomenon rather than as a single global scalar.

\subsection{Within-Language Script Control}
\label{app:hindi_script}
Prior work reports that code-switching between the same languages in different scripts can yield distinct activation patterns. Beyond the cross-language trend in \Cref{tab:app_language_anchor_bias} (weaker separation for Latin-script targets such as Spanish and French than for Korean and Chinese), we run a controlled within-language comparison: the same Hindi CS questions are evaluated in their original Devanagari form and in a romanized form, holding content, English spans, GF condition, model (Aya-Expanse-8B), and evaluation protocol fixed. We romanize all 96 available questions with Qwen3-235B-A22B and keep the 49 pairs that pass format, content-preservation, and quality checks. \Cref{tab:app_hindi_script} shows that GF-Src and GF-Tgt remain clearly separated under both scripts ($\Delta_{\mathrm{AB}}=1.08$ in Devanagari, $3.74$ romanized). The increase of $+2.66$ is itself significant and is driven mainly by GF-Tgt becoming more TGT-ward after romanization. The anchoring pattern thus persists when language and meaning are fixed, while its magnitude is sensitive to script.

\begin{table}[t]
\centering
\small
\setlength{\tabcolsep}{5pt}
\renewcommand{\arraystretch}{1.04}
\begin{adjustbox}{max width=\columnwidth}
\begin{tabular}{@{}lcccc@{}}
\toprule
\textbf{Hindi script} & \textbf{GF-Src AB} & \textbf{GF-Tgt AB} & $\Delta_{\mathrm{AB}}$ & $p$ \\
\midrule
Devanagari & $+0.07$ & $-1.02$ & $\mathbf{1.08}$ & $0.0019$ \\
Romanized  & $+0.71$ & $-3.03$ & $\mathbf{3.74}$ & $0.0006$ \\
\bottomrule
\end{tabular}
\end{adjustbox}
\caption{\textbf{Within-language script control.} The same 49 Hindi CS questions in Devanagari and romanized form (Aya-Expanse-8B); content, English spans, and GF condition are held fixed.}
\label{tab:app_hindi_script}
\end{table}

\subsection{Additional Anchor-Bias Variants}
\label{app:ab_variants}

Table~\ref{tab:anchor_bias_variants} reports four variants of the anchor-bias metric to verify that the GF-Src$\,{>}\,0$ / GF-Tgt$\,{<}\,0$ sign pattern is not an artefact of a specific pooling or normalization choice. $\textsc{AB}_{raw}$ is the raw question-content cosine difference; $\textsc{AB}_{nrm}$ normalizes by $1{-}b_m$ (the model's random-cosine dynamic range, shown in the second column); $\textsc{AB}_{tok}$ applies per-token L2 normalization before pooling; and $\textsc{AB}_{tok,nrm}$ combines both. All four variants agree on the sign for every model–condition pair: GF-Src is positive (SRC-anchored) for models with positive main-table bias, GF-Tgt is universally negative (TGT-anchored), and all Random and Selective conditions are negative. The normalized variants ($\textsc{AB}_{nrm}$, $\textsc{AB}_{tok,nrm}$) amplify absolute magnitudes substantially for models with small dynamic range (e.g., Llama3.2-1B, $1{-}b_m{=}0.0041$) but do not change the ordering or direction of any entry. We use $\textsc{AB}_{nrm}$ (normalized mean-pool) as the primary metric.

\begin{table*}[t]
\centering
\small
\setlength{\tabcolsep}{2.35pt}
\renewcommand{\arraystretch}{0.98}
\begin{adjustbox}{max width=\textwidth}
\begin{tabular}{@{}l!{\hspace{3pt}\color{GroupSep}\vrule width 0.25pt\hspace{3pt}}cc!{\hspace{3pt}\color{GroupSep}\vrule width 0.25pt\hspace{3pt}}cccc@{}}
\toprule
\multirow{2}{*}{\raisebox{-0.55ex}{\textbf{Language}}} &
\multicolumn{2}{c}{\textbf{Avg. F1}} &
\multicolumn{4}{c}{\textbf{$\Delta$F1 by CS condition}} \\
\cmidrule(lr){2-3} \cmidrule(lr){4-7}
& \textbf{\textsc{Base}} & \textbf{\textsc{CANVAS}} & \textbf{\textsc{GF-Src}} & \textbf{\textsc{GF-Tgt}} & \textbf{Random} & \textbf{Selective} \\
\midrule
Hindi & 10.47 & \canvasbest{11.43} & \canvasgain{+0.65} & \canvasgain{+1.51} & \canvasgain{+1.23} & \canvasgain{+0.44} \\
Marathi & 8.98 & \canvasbest{11.25} & \canvasgain{+1.43} & \canvasgain{+3.28} & \canvasgain{+2.26} & \canvasgain{+2.13} \\
Spanish & 10.37 & \canvasbest{10.55} & \canvasloss{-0.10} & \canvasgain{+0.20} & \canvasgain{+0.69} & \canvasloss{-0.06} \\
Urdu & 9.49 & \canvasbest{10.22} & \canvasgain{+0.15} & \canvasgain{+0.96} & \canvasgain{+1.01} & \canvasgain{+0.81} \\
Simplified Chinese & 9.44 & \canvasbest{9.91} & \canvasgain{+0.18} & \canvasgain{+0.98} & \canvasgain{+0.29} & \canvasgain{+0.42} \\
Korean & 7.83 & \canvasbest{9.48} & \canvasgain{+1.49} & \canvasgain{+2.10} & \canvasgain{+1.51} & \canvasgain{+1.52} \\
French & 8.69 & \canvasbest{8.79} & \canvasgain{+0.02} & \canvasloss{-0.50} & \canvasgain{+0.39} & \canvasgain{+0.51} \\
Bengali & 7.37 & \canvasbest{8.61} & \canvasgain{+0.65} & \canvasgain{+1.52} & \canvasgain{+1.23} & \canvasgain{+1.54} \\
Toba Batak & 6.94 & \canvasbest{7.01} & \canvasloss{-0.33} & \canvasloss{-0.05} & \canvasgain{+0.37} & \canvasgain{+0.31} \\
\midrule
\rowcolor{MeanGray}
\textbf{\textit{Mean}} & 8.84 & \canvasbest{9.70} & \canvasgain{+0.46} & \canvasgain{+1.11} & \canvasgain{+1.00} & \canvasgain{+0.85} \\
\bottomrule
\end{tabular}
\end{adjustbox}
\caption{\textbf{Language-wise CANVAS mitigation.} We use the same reporting convention as Table~\ref{tab:canvas_results}: Avg. F1 is averaged over four CS conditions, and condition-wise columns report CANVAS--Base F1 changes.}
\label{tab:app_language_canvas_mitigation}
\end{table*}

\subsection{Frontier Model Scaling Validation}
\label{app:frontier}

\paragraph{Setup.}
The Layer-AB and Mitigation pools used in the main paper are restricted to models whose weights are publicly available, so that the anchor-bias analysis and \textsc{CANVAS} mitigation can be performed on internal hidden states. A natural concern is whether code-switched degradation also persists at the closed-source frontier scale, where models substantially outperform our open pool on general QA. We address this concern by evaluating three frontier API models on the same $2{,}879$-row CS dataset used throughout the paper: DeepSeek-V3.2~\cite{liu2024deepseek}, Gemini 2.5 Flash~\cite{comanici2025gemini}, and GPT-5.4-mini~\cite{singh2025openai}. All three models are queried via OpenRouter under identical decoding settings (greedy, $\texttt{max\_new\_tokens}=48$) and the same system prompt as the open pool. Each model produces predictions for the EN, CS, and TGT conditions; we report F1 against the same gold answers.

\paragraph{Frontier CS degradation.}
Table~\ref{tab:app_frontier_scaling} reports aggregate F1 across the three conditions. Two findings stand out. First, all three frontier models exhibit CS degradation: $\Delta\textrm{F1}_{\text{CS}-\text{EN}}$ ranges from $-1.88$ (GPT-5.4-mini) to $-7.64$ (Gemini 2.5 Flash). Although the magnitude varies, the direction is consistent — \emph{none} of the frontier models is immune to code-switching. Second, the source-language anchoring ordering $\textrm{F1}_{\text{TGT}}<\textrm{F1}_{\text{CS}}<\textrm{F1}_{\text{EN}}$ that we report for the open pool is preserved in every frontier model. The TGT$-$EN gap ranges from $-3.54$ to $-19.07$, indicating that source-language preference is not a small-model artifact but a property that scales with model capacity.

\begin{table}[t]
\centering
\small
\setlength{\tabcolsep}{5pt}
\renewcommand{\arraystretch}{1.04}
\begin{adjustbox}{max width=\linewidth}
\begin{tabular}{@{}l rrr rr@{}}
\toprule
\textbf{Model} & \textbf{EN} & \textbf{CS} & \textbf{TGT} &
$\Delta\textrm{F1}_{\text{CS-EN}}$ & $\Delta\textrm{F1}_{\text{TGT-EN}}$ \\
\midrule
\modelwithlogo{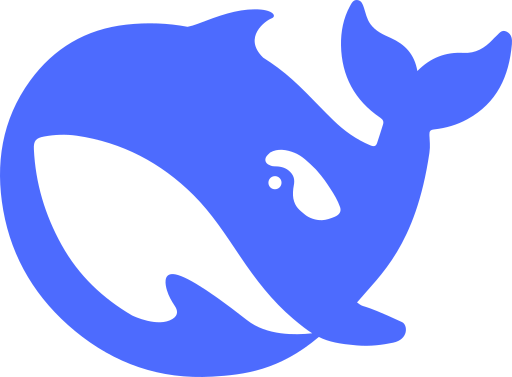}{DeepSeek-V3.2}  & 19.97 & 16.86 & 11.28 & \canvasloss{$-3.11$} & \canvasloss{$-8.69$}  \\
\modelwithlogo{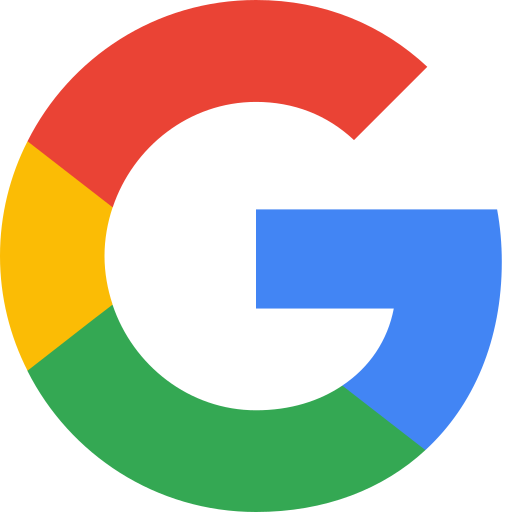}{Gemini 2.5 Flash} & 28.42 & 20.78 &  9.34 & \canvasloss{$-7.64$} & \canvasloss{$-19.07$} \\
\modelwithlogo{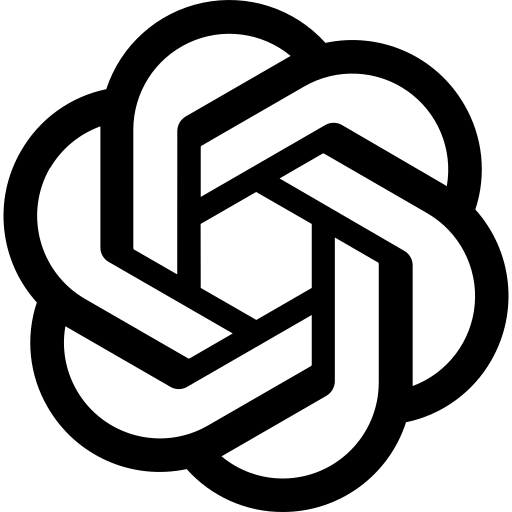}{GPT-5.4-mini}     & 19.59 & 17.71 & 16.05 & \canvasloss{$-1.88$} & \canvasloss{$-3.54$}  \\
\bottomrule
\end{tabular}
\end{adjustbox}
\caption{\textbf{Frontier model scaling check.} Three closed-source frontier models evaluated via OpenRouter on the same $2{,}879$-row CS dataset. CS degradation ($\Delta\textrm{F1}_{\text{CS}-\text{EN}}<0$) and the source-language preference ordering $\textrm{F1}_{\text{TGT}}<\textrm{F1}_{\text{CS}}<\textrm{F1}_{\text{EN}}$ both hold for every frontier model.}
\label{tab:app_frontier_scaling}
\end{table}

\paragraph{Open vs.\ frontier CS performance.}
A useful cross-check is the actual CS performance gap relative to model scale. Our largest open model, Qwen3.6-35B-A3B ($35\mathrm{B}$ parameters, $\sim\!3\mathrm{B}$ active per token), reaches CS F1 $15.80$ under direct CS answering and $17.53$ with \textsc{CANVAS} applied (Table~\ref{tab:app_canvas_cost}). The frontier results in Table~\ref{tab:app_frontier_scaling} sit at $16.86$ (DeepSeek-V3.2), $17.71$ (GPT-5.4-mini), and $20.78$ (Gemini 2.5 Flash). \textsc{CANVAS} therefore brings the open MoE model within $0.18$ F1 of GPT-5.4-mini and ahead of DeepSeek-V3.2; only Gemini 2.5 Flash retains a meaningful $3.25$ F1 lead. Given that DeepSeek-V3.2 is roughly an order of magnitude larger in total parameter count than Qwen3.6-35B-A3B, the practical CS F1 gap is small relative to the underlying capacity difference: scaling alone does not close the CS gap, while a representation-level intervention applied to a moderate-capacity open model achieves comparable CS performance.

\subsection{Lexical Language versus Word Order}
\label{app:ko_english_order_anchor}

\paragraph{Setup.}
We test whether word order alone can move an input across the source--target anchor axis. Starting from the Korean-target examples, we construct two same-lexicon order diagnostics using OpenRouter Qwen3-235B-A22B only to change the order of the existing lexical material. \textsc{SRC-TGT}$_{\mathrm{order}}$ keeps source-language lexical material but arranges it in target-like order, while \textsc{TGT-SRC}$_{\mathrm{order}}$ keeps target-language lexical material but arranges it in source-like order. We compare these probes with pure source text (\textsc{SRC}), pure target text (\textsc{TGT}), and the two grammar-forced CS conditions. We evaluate three local 7--8B-level models and report normalized anchor bias, where positive values indicate source-oriented representations and negative values indicate target-oriented representations.

\paragraph{Results.}
Table~\ref{tab:app_ko_english_order_anchor_probe} shows that word order does affect the strength of anchoring, even though it does not fully determine the anchor direction. When we keep source-language words but impose target-like order, \textsc{SRC-TGT}$_{\mathrm{order}}$ moves away from the pure \textsc{SRC} endpoint: its mean $\mathrm{AB}_{\mathrm{nrm}}$ drops from $+1.000$ to $+0.647$. This means that changing only the order of the source sentence already reduces source anchoring and shifts the representation toward the target side. Conversely, when we keep target-language words but impose source-like order, \textsc{TGT-SRC}$_{\mathrm{order}}$ moves slightly away from the pure \textsc{TGT} endpoint: its mean $\mathrm{AB}_{\mathrm{nrm}}$ increases from $-1.000$ to $-0.876$. This movement is smaller than the source-to-target-order case on average, although Aya-Expanse-8B shows a relatively large shift.

The two order-counterfactual probes therefore occupy intermediate positions rather than collapsing exactly to either monolingual endpoint. \textsc{SRC-TGT}$_{\mathrm{order}}$ stays between \textsc{SRC} and the CS conditions on the source side, while \textsc{TGT-SRC}$_{\mathrm{order}}$ stays between \textsc{TGT} and the CS conditions on the target side. This confirms that grammatical frame is not a superficial formatting detail: changing only the order of the same lexical material is enough to move the internal representation away from its original monolingual anchor.

\begin{table*}[t]
\centering
\small
\setlength{\tabcolsep}{5.0pt}
\renewcommand{\arraystretch}{1.04}
\begin{adjustbox}{max width=\textwidth}
\begin{tabular}{@{}lrrrrrr@{}}
\toprule
\textbf{Model} &
\textbf{\textsc{SRC}} &
\textbf{\textsc{SRC-TGT}$_{\mathrm{order}}$} &
\textbf{\textsc{CS GF-Src}} &
\textbf{\textsc{CS GF-Tgt}} &
\textbf{\textsc{TGT-SRC}$_{\mathrm{order}}$} &
\textbf{\textsc{TGT}} \\
\midrule
\modelwithlogo{cohere}{Aya-Expanse-8B}
& 1.000 & 0.445 & 0.313 & -0.606 & -0.765 & -1.000 \\
\modelwithlogo{meta}{Llama3.1-8B}
& 1.000 & 0.692 & 0.032 & -0.641 & -0.910 & -1.000 \\
\modelwithlogo{qwen}{Qwen3.5-4B}
& 1.000 & 0.804 & 0.076 & -0.627 & -0.954 & -1.000 \\
\midrule
\rowcolor{MeanGray}
\textbf{\textit{Mean}}
& 1.000 & 0.647 & 0.140 & -0.625 & -0.876 & -1.000 \\
\bottomrule
\end{tabular}
\end{adjustbox}
\caption{\textbf{Lexical language versus word order on Korean-target examples.}
We report normalized anchor bias. \textsc{SRC-TGT}$_{\mathrm{order}}$ keeps source-language words but places them in target-like order; \textsc{TGT-SRC}$_{\mathrm{order}}$ keeps target-language words but places them in source-like order. Positive values indicate source-oriented representations; negative values indicate target-oriented representations.}
\label{tab:app_ko_english_order_anchor_probe}
\end{table*}

\subsection{Token-Proportion Control}
\label{app:rho_control}
GF-Src and GF-Tgt differ not only in grammatical frame but also in source-token proportion ($\rho{=}0.59$ vs.\ $0.40$ on average). To check that the frame contrast is not a token-ratio artifact, we restrict to examples whose $\rho$ lies in progressively narrower bands around $0.5$ and recompute the per-example $\mathrm{AB}^{\mathrm{upper}}_{\mathrm{nrm}}$ separation. \Cref{tab:app_rho_control} shows that the separation decreases as the band narrows, confirming that token proportion contributes, but remains clearly positive and highly significant even in the tightest band ($\rho\in[0.45,0.55]$: $+0.38$ for Aya-Expanse-8B, $+0.55$ for Qwen3.5-9B). The grammatical frame therefore shifts anchoring over and above token proportion, consistent with the softened claim in \Cref{sec:anchor_bias}.

\begin{table}[t]
\centering
\small
\setlength{\tabcolsep}{3.5pt}
\renewcommand{\arraystretch}{1.05}
\begin{adjustbox}{max width=\columnwidth}
\begin{tabular}{@{}llcc@{}}
\toprule
\textbf{Subset} & $\rho$ \textbf{band} & \textbf{Aya-Expanse-8B} $\Delta_{\mathrm{AB}}$ ($p$) & \textbf{Qwen3.5-9B} $\Delta_{\mathrm{AB}}$ ($p$) \\
\midrule
Full       & --            & $+1.30$ ($4{\times}10^{-83}$) & $+1.74$ ($8{\times}10^{-67}$) \\
Restricted & $[0.35,0.65]$ & $+0.75$ ($3{\times}10^{-25}$) & $+0.94$ ($1{\times}10^{-15}$) \\
Restricted & $[0.40,0.60]$ & $+0.58$ ($7{\times}10^{-12}$) & $+0.72$ ($2{\times}10^{-8}$) \\
Restricted & $[0.45,0.55]$ & $+0.38$ ($7{\times}10^{-4}$)  & $+0.55$ ($9{\times}10^{-4}$) \\
\bottomrule
\end{tabular}
\end{adjustbox}
\caption{\textbf{Frame effect under balanced token proportion.}
Per-example $\mathrm{AB}^{\mathrm{upper}}_{\mathrm{nrm}}$ separation (GF-Src minus GF-Tgt) on the full set and on subsets whose source-token ratio $\rho=|C_{\mathrm{src}}|/(|C_{\mathrm{src}}|{+}|C_{\mathrm{tgt}}|)$ lies in the given band.}
\label{tab:app_rho_control}
\end{table}

\section{CANVAS Details}
\label{app:canvas_details}
\subsection{Adaptive Alpha Schedule and Clip Range}
\label{app:alpha_clip_validation}
\paragraph{Setup.}
CANVAS uses the adaptive interpolation rule from the main method section:
\begin{equation*}
\alpha
=
\mathrm{clip}_{[\alpha_{\min},\alpha_{\max}]}
\left(\alpha_0-\lambda\gamma_{\mathrm{CS}}\right),
\end{equation*}
where $\gamma_{\mathrm{CS}}$ is the representation-alignment score computed from the clean prefill over the question span. The term $-\lambda\gamma_{\mathrm{CS}}$ is intentional: if the CS state is already source-canvas aligned ($\gamma_{\mathrm{CS}}>0$), CANVAS applies a weaker correction; if it is target-oriented ($\gamma_{\mathrm{CS}}<0$), CANVAS increases the interpolation strength. The main paper setting is $(\alpha_0,\lambda,\alpha_{\min},\alpha_{\max})=(0.45,1.5,0.05,0.75)$.

\paragraph{Schedule choice.}
We use this single global schedule because it balances three constraints: it improves average CS F1, preserves the intended monotonic behavior, and keeps the intervention bounded. The center $\alpha_0=0.45$ gives a moderate interpolation when the alignment score is near neutral, while $\lambda=1.5$ creates a visible separation between source-aligned and target-oriented inputs before clipping. A smaller $\lambda$ would flatten that adaptivity; a much larger $\lambda$ would make moderate alignment changes saturate the clip bounds more often. We choose the clip range $[0.05,0.75]$ to keep the lower regime close to an identity update and the upper regime short of target-state replacement. We do not tune language-specific, condition-specific, or model-specific constants.

\paragraph{Parameter interpretation.}
The four values control different aspects of the same bounded controller. Raising $\alpha_0$ strengthens the default update for inputs with $\gamma_{\mathrm{CS}}\approx0$, whereas lowering it makes neutral cases closer to direct CS answering. Raising $\alpha_{\min}$ forces more interpolation even for strongly source-aligned cases, while lowering it makes the low end more nearly identity-like. Raising $\alpha_{\max}$ permits stronger source-canvas mixing for target-oriented cases, while lowering it caps those corrections earlier; at our upper bound, at least one quarter of the original target-token representation is preserved. Table~\ref{tab:app_alpha_schedule} summarizes these roles.

\begin{figure*}[t]
\centering
\includegraphics[width=0.88\textwidth]{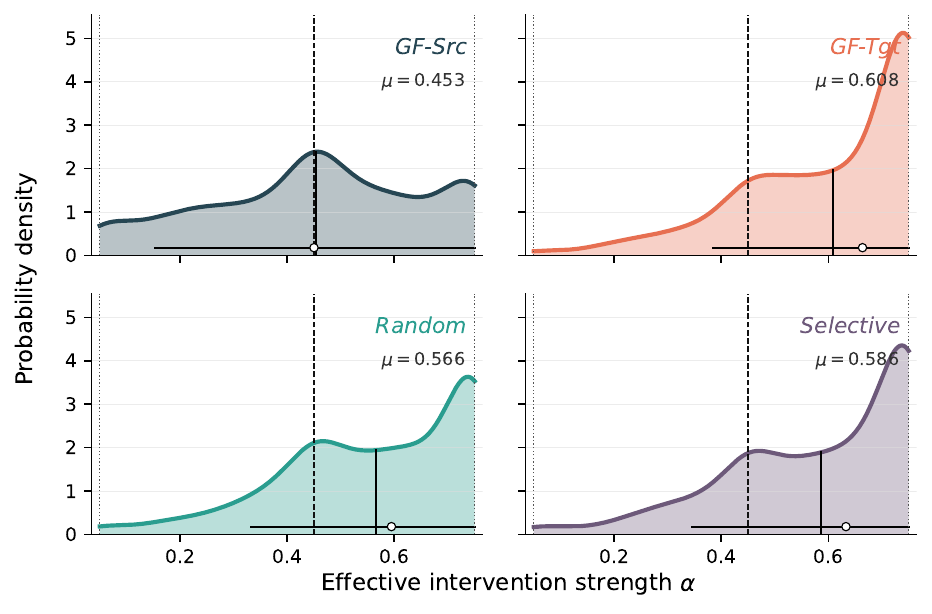}
\caption{\textbf{Effective $\alpha$ distribution under the validation-selected CANVAS rule.}
We plot the area-normalized probability density of the effective interpolation strength used by CANVAS. We mark the validation-selected center $\alpha_0=0.45$ and clip range $[0.05,0.75]$ with dashed and dotted vertical lines, respectively. We assign weaker updates to SRC-heavy \textsc{GF-Src} examples and stronger updates to TGT-heavy \textsc{GF-Tgt} examples, while random and selective switching remain between these two endpoints.}
\label{fig:app_alpha_distribution}
\end{figure*}

\paragraph{Observed distribution.}
We visualize the effective $\alpha$ distribution from the CANVAS representation-analysis dump in Figure~\ref{fig:app_alpha_distribution}. We find that the learned behavior matches the intended controller: \textsc{GF-Src} has a higher source-token ratio and receives a weaker intervention on average ($\alpha=0.453$), while \textsc{GF-Tgt} is more target-heavy and receives a stronger intervention ($\alpha=0.608$). We also observe that random and selective switching falls between these two endpoints.

\begin{table*}[t]
\centering
\small
\setlength{\tabcolsep}{3.5pt}
\renewcommand{\arraystretch}{1.16}
\begin{tabularx}{\linewidth}{@{}>{\raggedright\arraybackslash}p{0.29\linewidth}c>{\raggedright\arraybackslash}X@{}}
\toprule
Component & Value & Interpretation \\
\midrule
Base strength $\alpha_{\mathrm{base}}$ &
$0.45$ &
Sets $\alpha$ when $\gamma_{\mathrm{CS}}\!\approx\!0$. Larger values strengthen neutral cases. \\
\addlinespace[2pt]
Adaptivity scale $\lambda$ &
$1.5$ &
Scales $-\lambda\gamma_{\mathrm{CS}}$. Larger values separate SRC-/TGT-oriented cases more sharply. \\
\addlinespace[2pt]
Lower clip $\alpha_{\min}$ &
$0.05$ &
Keeps strongly SRC-aligned cases near identity. Larger values force a stronger minimum update. \\
\addlinespace[2pt]
Upper clip $\alpha_{\max}$ &
$0.75$ &
Prevents full target-state replacement. Larger values allow stronger source-canvas mixing. \\
\bottomrule
\end{tabularx}
\caption{\textbf{CANVAS adaptive-$\alpha$ schedule.}
We use one global bounded controller for all languages, conditions.}
\label{tab:app_alpha_schedule}
\end{table*}

\subsection{Layer-wise CANVAS Analysis}
\label{app:layerwise_canvas}

\paragraph{Setup.}
We analyze where the CANVAS-induced representation shift appears across depth. CANVAS intervenes only after the control-layer boundary, so the expected pattern is layer-local: lower layers should remain close to the base CS trajectory, while upper layers should show an increasing source-directed shift after entering $\mathcal{L}_{\mathrm{ctrl}}$.

\begin{figure*}[t]
  \centering
  \includegraphics[width=0.82\textwidth]{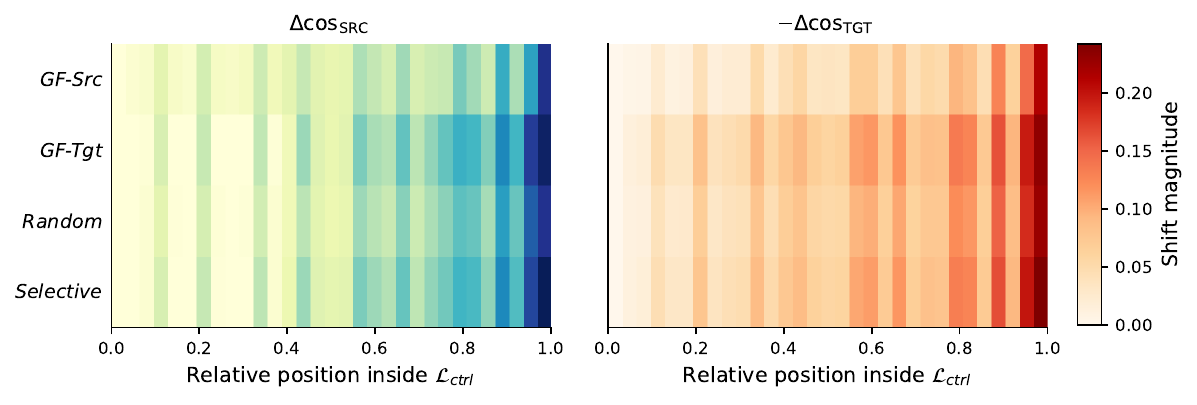}
  \caption{\textbf{Layer-wise CANVAS effect.}
  We plot the magnitude of the source-ward shift across the relative position inside $\mathcal{L}_{\mathrm{ctrl}}$. Darker cells indicate larger $\Delta\cos_{\mathrm{SRC}}$ or larger $-\Delta\cos_{\mathrm{TGT}}$. The effect is strongest near the later control layers, matching the layer-wise anchoring analysis.}
  \label{fig:app_canvas_layerwise}
\end{figure*}

\paragraph{Results.}
Figure~\ref{fig:app_canvas_layerwise} shows this trajectory as a heatmap over the relative position inside $\mathcal{L}_{\mathrm{ctrl}}$. The left panel reports $\Delta\cos_{\mathrm{SRC}}$, and the right panel reports $-\Delta\cos_{\mathrm{TGT}}$, so darker colors in both panels indicate a stronger source-ward correction. The shift becomes larger toward the later part of the control interval, supporting the mechanism implied by the anchor-bias analysis: CANVAS redirects the late-layer CS state rather than globally rewriting the entire prompt representation.

\subsection{Source-heavy Checks}
\label{app:source_heavy_checks}

\paragraph{CANVAS preserves source-heavy inputs.}
We next check cases that are already mostly source-language text. We select examples with a detected source-token ratio $\rho_{\mathrm{SRC}}\ge0.80$ from the full CANVAS runs and split them into three diagnostic rows. ``All source-heavy'' is the full selected set. ``Steered source-heavy'' keeps only examples where the detector still finds at least one target-language token span ($|C_{\mathrm{tgt}}|>0$), so CANVAS actually has hidden states to modify. ``Identity source-heavy'' contains examples with no detected target span ($|C_{\mathrm{tgt}}|=0$), where CANVAS should leave the prompt effectively unchanged. Table~\ref{tab:app_high_src_ratio_canvas} shows that the full source-heavy set uses a moderate mean $\alpha$ (0.445) and preserves performance: average F1 changes from 9.08 to 9.12, with 97.3\% of examples showing no F1 drop. Even in the stricter steered subset, where CANVAS actually intervenes, the average change remains small ($\Delta$F1=-0.21), and 96.0\% of examples show no drop.

\begin{table*}[t]
\centering
\small
\setlength{\tabcolsep}{3.4pt}
\renewcommand{\arraystretch}{1.04}
\begin{adjustbox}{max width=\textwidth}
\begin{tabular}{@{}llrrrrrrr@{}}
\toprule
\textbf{Subset} & \textbf{Definition} & Mean $\rho_{\mathrm{SRC}}$ & Mean $\alpha$ & \textbf{Base F1} & \textbf{CANVAS F1} & $\Delta$F1 & No-drop (\%) & Fixed / Lost \\
\midrule
All source-heavy & $\rho_{\mathrm{SRC}}\ge0.80$ & 0.991 & 0.445 & 9.08 & 9.12 & \canvasgain{+0.04} & 97.3 & 265/246 \\
Steered source-heavy & $\rho_{\mathrm{SRC}}\ge0.80,\ |C_{\mathrm{tgt}}|>0$ & 0.861 & 0.366 & 15.04 & 14.83 & \canvasloss{-0.21} & 96.0 & 18/26 \\
Identity source-heavy & $\rho_{\mathrm{SRC}}\ge0.80,\ |C_{\mathrm{tgt}}|=0$ & 1.000 & 0.450 & 8.68 & 8.74 & \canvasgain{+0.06} & 97.4 & 247/220 \\
\bottomrule
\end{tabular}
\end{adjustbox}
\caption{\textbf{CANVAS on source-heavy CS inputs.}
We split high-source-ratio examples into all cases, cases with at least one target-token span that CANVAS can steer, and cases with no detected target span where CANVAS should behave as an identity operation. No-drop reports the percentage of examples where CANVAS F1 is at least direct CS F1. Fixed/Lost counts base-zero to CANVAS-positive cases and base-positive to CANVAS-zero cases.}
\label{tab:app_high_src_ratio_canvas}
\end{table*}

\subsection{Token-Level Language Tagging for \textsc{CANVAS}}
\label{app:canvas_token_tagging}

\textsc{CANVAS} requires identifying which question tokens belong to the source-language span and which belong to the target-language span. We therefore apply lightweight token-level language tagging before computing the source canvas and target anchor. Each token in the question-content span is assigned one of three labels: \textsc{Src}, \textsc{Tgt}, or \textsc{Other}. \textsc{Src} denotes source-language tokens, which correspond to English tokens in our main English--target-language setting. \textsc{Tgt} denotes target-language tokens, and \textsc{Other} denotes punctuation, numbers, whitespace artifacts, or language-neutral tokens.

Because chat templates include system instructions and assistant-prefix tokens that are shared across all inputs, we first locate the raw user question inside the full templated prompt. When tokenizer offset mappings are available, we use them to identify the token positions overlapping with the question span. Tokens outside this span are not used for canvas construction, alignment estimation, or interpolation. This ensures that the \textsc{CANVAS} signal is computed from the code-switched question itself rather than from shared prompt text.
For non-Latin target languages, we use Unicode script information. Tokens containing only Hangul (U+AC00--U+D7A3), CJK (U+4E00--U+9FFF), Devanagari (U+0900--U+097F), Arabic, or Bengali-script characters are labeled as \textsc{Tgt}. Tokens containing only Latin-script alphabetic characters are labeled as \textsc{Src}, except for Latin-script target languages. For Spanish, French, and Toba Batak, script information alone is insufficient because both English and the target language use Latin characters. In these cases, we use the Lingua probabilistic language detector\footnote{\url{https://github.com/pemistahl/lingua-py}} restricted to the \{English, target\} language pair to assign each Latin-script token to \textsc{Src} or \textsc{Tgt}. Tokens with mixed Latin and non-Latin characters are conservatively labeled as \textsc{Tgt}, while punctuation and empty or non-alphabetic surfaces are labeled as \textsc{Other}. 

For BPE tokenizers, raw token strings may not directly reflect the original Unicode surface form. We therefore decode each token ID back into its surface string before applying script-based tagging. If either the source or target span contains too few tokens, \textsc{CANVAS} skips the intervention for that example and falls back to ordinary decoding.

\subsection{Hidden-State Extraction for CANVAS}
\label{app:canvas_hidden_extraction}

\textsc{CANVAS} computes its source canvas, target anchor, and alignment score from the hidden states of the user question, not from the full chat-templated input. For each example, the model input is the concatenation of system-prompt tokens, user-question tokens, and assistant-prefix tokens:
\begin{equation}
\mathbf{x}
=
[\mathbf{x}_{\mathrm{sys}};\mathbf{x}_{\mathrm{q}};\mathbf{x}_{\mathrm{asst}}],
\end{equation}
where $\mathbf{x}_{\mathrm{sys}}$ contains the shared QA instruction, $\mathbf{x}_{\mathrm{q}}$ contains the actual code-switched question, and $\mathbf{x}_{\mathrm{asst}}$ contains model-specific generation-template tokens such as assistant role markers. Since $\mathbf{x}_{\mathrm{sys}}$ and $\mathbf{x}_{\mathrm{asst}}$ are shared across inputs and can obscure the representation of the CS question itself, all \textsc{CANVAS} signals are computed only from token positions overlapping with $\mathbf{x}_{\mathrm{q}}$.

Concretely, we run a clean forward pass with hidden-state outputs enabled. For a model with $L$ layers, sequence length $T$, and hidden size $d$, this gives
\begin{equation}
H_M(q^{\mathrm{CS}})\in\mathbb{R}^{L\times T\times d}.
\end{equation}
We then identify the token positions corresponding to the user question using tokenizer offsets when available. These question positions are further tagged as source, target, or other, producing $C_{\mathrm{src}}$, $C_{\mathrm{tgt}}$, and $C_{\mathrm{oth}}$.

For each selected upper layer $\ell\in\mathcal{L}_{\mathrm{ctrl}}$, we gather the hidden states at source-language positions and average them along the token dimension to obtain one layer-specific source canvas:
\begin{equation}
\mathbf{c}^{\mathrm{src}}_{\ell}
=
\frac{1}{|C_{\mathrm{src}}|}
\sum_{t\in C_{\mathrm{src}}}
H_M(q^{\mathrm{CS}})[\ell,t,:].
\end{equation}
The target anchor is computed analogously:
\begin{equation}
\mathbf{a}^{\mathrm{tgt}}_{\ell}
=
\frac{1}{|C_{\mathrm{tgt}}|}
\sum_{t\in C_{\mathrm{tgt}}}
H_M(q^{\mathrm{CS}})[\ell,t,:].
\end{equation}
Thus, each selected layer has its own source canvas and target anchor. For example, if two source-token hidden states at a layer are
\[
[a,b,c,d]
\quad\text{and}\quad
[e,f,g,h],
\]
their mean source canvas is
\[
\left[
\frac{a+e}{2},
\frac{b+f}{2},
\frac{c+g}{2},
\frac{d+h}{2}
\right].
\]
This layer-specific averaging is repeated independently for every $\ell\in\mathcal{L}_{\mathrm{ctrl}}$, so \textsc{CANVAS} aligns target-token states to the source canvas of the same layer rather than to a single global vector.

\subsection{CANVAS Layer-span Ablation}
\label{app:canvas_layer_ablation}

\paragraph{Evaluation protocol.}
We compare direct code-switched answering (\textsc{BASE}) with CANVAS on the same short-answer QA setting used in the previous sections. For the intervention study, we evaluate MLLMs across four CS conditions: \textsc{GF-SRC}, \textsc{GF-TGT}, random switching, and selective switching. We report average F1 over the four CS conditions and condition-wise F1 changes relative to \textsc{BASE}. We also include a random-direction baseline, \textsc{CANVAS}$_{\mathrm{rand}}$, which keeps the same adaptive $\alpha$ controller and 30\% layer budget as CANVAS but replaces the source--target representation direction with a random interpolation direction.

\paragraph{Setup.}
We vary the intervention span while keeping the current adaptive schedule fixed, $\alpha=\mathrm{clip}_{[0.05,0.75]}(0.45-1.5\gamma_{\mathrm{CS}})$. \textsc{CANVAS}-$k$ applies the intervention only to the last $k\%$ of layers. We choose the 30\% span as the main setting because it matches the upper-layer region where the anchor-bias analysis shows stable language-frame commitment, while still leaving the lower and middle layers to encode the original CS question without intervention. Thus, the ablation tests whether CANVAS should act as a late anchor correction or as a broad representation rewrite.

\paragraph{Results.}
Table~\ref{tab:app_canvas_layer_ablation} supports this choice. \textsc{CANVAS}-30 gives the best average F1, while \textsc{CANVAS}-50 performs nearly on par and can be viewed as a stronger, wider-span variant. We use \textsc{CANVAS}-30 as the main setting because it achieves the same level of improvement with a narrower intervention span, making it the minimal effective intervention: it targets the late anchor-commitment region while leaving more of the lower and middle-layer CS composition untouched. In contrast, performance drops for \textsc{CANVAS}-70 and collapses for \textsc{CANVAS}-90. This pattern suggests that increasing the span beyond the late control region does not simply strengthen the same correction. Once the intervention reaches too far into lower layers, it perturbs lexical and compositional processing before the model has formed a stable high-level CS state; the source-direction update then competes with question encoding rather than correcting a late anchor. The random-direction baseline is also lower than \textsc{CANVAS}-30 for every evaluated model, so the gain does not come from arbitrary hidden-state mixing; it depends on applying the source--target anchor direction in the late control region identified by the analysis.

\begin{table*}[t]
\centering
\small
\setlength{\tabcolsep}{3.8pt}
\renewcommand{\arraystretch}{1.03}
\begin{adjustbox}{max width=\textwidth}
\begin{tabular}{lrrrrrr}
\toprule
\textbf{Model} &
\textbf{\textsc{CANVAS}-10} &
\textbf{\textsc{CANVAS}-30} &
\textbf{\textsc{CANVAS}-50} &
\textbf{\textsc{CANVAS}-70} &
\textbf{\textsc{CANVAS}-90} &
\textbf{\textsc{CANVAS}$_{\mathrm{rand}}$} \\
\midrule
\modelwithlogo{cohere}{Aya-Expanse-8B}
& 6.03 & 6.30 & \textbf{7.11} & 6.70 & 4.91 & 5.99 \\
\modelwithlogo{meta}{Llama3.1-8B}
& 7.11 & 7.68 & \textbf{8.03} & 7.25 & 5.28 & 6.72 \\
\modelwithlogo{mistral}{Mistral-7B}
& 4.92 & 5.13 & \textbf{5.15} & 4.27 & 3.38 & 4.96 \\
\modelwithlogo{microsoft}{Phi-3.5-MoE}
& 7.96 & 8.58 & \textbf{8.81} & 8.30 & 6.17 & 7.93 \\
\modelwithlogo{qwen}{Qwen3.5-27B}
& 14.93 & \textbf{15.03} & 14.45 & 12.85 & 8.95 & 14.22 \\
\modelwithlogo{qwen}{Qwen3.5-4B}
& 7.80 & \textbf{7.98} & 7.81 & 7.03 & 4.83 & 7.35 \\
\modelwithlogo{qwen}{Qwen3.5-9B}
& 10.04 & \textbf{10.24} & 9.52 & 8.72 & 5.56 & 10.01 \\
\modelwithlogo{qwen}{Qwen3-30B-A3B}
& 7.17 & \textbf{8.13} & 8.10 & 7.33 & 6.00 & 6.93 \\
\midrule
\rowcolor{MeanGray}
\textbf{\textit{Mean}}
& 8.24 & \textbf{8.64} & 8.62 & 7.81 & 5.64 & 8.01 \\
\bottomrule
\end{tabular}
\end{adjustbox}
\caption{\textbf{Layer-span ablation for \textsc{CANVAS}.}
We report average F1 over the four CS conditions. \textsc{CANVAS}-$k$ applies the adaptive intervention to the last $k\%$ of layers. \textsc{CANVAS}$_{\mathrm{rand}}$ uses the same adaptive $\alpha$ and the same 30\% layer span as \textsc{CANVAS}-30, but replaces the source--target representation direction with a random interpolation direction.}
\label{tab:app_canvas_layer_ablation}
\end{table*}

\subsection{Comparison with Steering Variants and Baselines}
\label{app:canvas_baselines}
\Cref{tab:cr_baselines} compares CANVAS with ablated variants, controls, and the closest cross-lingual activation-steering methods, all run with the same model, prompts, control layers, and budget. Removing the instance-specific controller (fixed-$\alpha$), replacing $\gamma$ with the source-token ratio, or projecting onto the source axis each lowers F1, and both controls (a norm-matched random direction and a shuffled, unrelated canvas) fall below the full method, so the in-context canvas and the alignment-based controller both contribute. CANVAS also outperforms corpus-level methods that fit language vectors from parallel data \cite{kirtane2026language,mahmoud2026aligning,lim2025languagespecific} while requiring no such fitting; the gap to the strongest of these (mean-difference steering, 8.67) is small here because headroom on SimpleQA-derived CodeMixQA is limited, which motivates the paired comparison in higher-headroom settings in Appendix~\ref{app:canvas_rag_topic_llama}.

\begin{table}[t]
\centering
\small
\setlength{\tabcolsep}{3pt}
\renewcommand{\arraystretch}{1.04}
\begin{adjustbox}{max width=\columnwidth}
\begin{tabular}{@{}llr@{}}
\toprule
\textbf{Type} & \textbf{Method} & \textbf{F1} \\
\midrule
No interv. & \textsc{Base} & 6.81 \\
\midrule
\multirow{4}{*}{Variant}
 & Fixed-$\alpha$ interpolation & 8.30 \\
 & Source-axis projection & 8.32 \\
 & Source-ratio $\alpha$ controller & 8.56 \\
 & \textbf{\textsc{CANVAS} (ours)} & \textbf{8.75} \\
\midrule
\multirow{2}{*}{Control}
 & Norm-matched random direction & 8.18 \\
 & Shuffled/unrelated canvas & 8.65 \\
\midrule
\multirow{3}{*}{Prior steering}
 & \cite{lim2025languagespecific} & 7.72 \\
 & \cite{mahmoud2026aligning} & 8.48 \\
 & \cite{kirtane2026language} & 8.67 \\
\bottomrule
\end{tabular}
\end{adjustbox}
\caption{\textbf{Comparison with \textsc{CANVAS} variants, controls, and prior cross-lingual activation steering} \cite{lim2025languagespecific,mahmoud2026aligning,kirtane2026language} on CodeMixQA (Qwen3-30B-A3B; macro F1 over the four CS conditions, identical prompts, control layers, and budget across rows).}
\label{tab:cr_baselines}
\end{table}

\subsection{Language-wise CANVAS Mitigation}
\label{app:language_canvas_mitigation}
\paragraph{Setup.}
We reuse the main CANVAS outputs and group examples by target language. Following Table~\ref{tab:canvas_results}, Avg. F1 is the mean over the four CS conditions, and each condition column reports CANVAS--Base F1 change.

\paragraph{Results.}
Table~\ref{tab:app_language_canvas_mitigation} shows that CANVAS improves most language groups, with the largest gains on languages where the direct CS baseline is weaker. The gains are not restricted to a single CS construction: improvements appear across grammar-forced, random, and selective switching, although some high-resource or already-stable cases show smaller or slightly negative changes.

\paragraph{Answer-language decoupling.}
In the main setting English is both the source/frame reference and the required answer language. To separate the two, we keep the CS inputs and the frame reference fixed and require Qwen3.5-9B to answer in Korean or Spanish instead, scoring against gold answers translated into that language. \Cref{tab:app_answer_decoupling} reports the per-condition breakdown. CANVAS retains positive gains in every decoupled cell (mean $+0.41$ for Korean answers, $+0.49$ for Spanish answers), comparable to the English-answer results for the same language pairs ($+0.63$, $+0.25$). The benefit therefore follows source-frame alignment rather than a match between source and answer language.

\begin{table}[t]
\centering
\small
\setlength{\tabcolsep}{3pt}
\renewcommand{\arraystretch}{1.04}
\begin{adjustbox}{max width=\columnwidth}
\begin{tabular}{@{}llccc@{}}
\toprule
\textbf{Setting} & \textbf{CS target / answer lang.} & \textbf{GF-Src} & \textbf{GF-Tgt} & \textbf{Mean} \\
\midrule
Original  & Korean CS / English   & \canvasloss{$-0.63$} & \canvasgain{$+1.88$} & \canvasgainbest{$+0.63$} \\
Original  & Spanish CS / English  & \canvasgain{$+0.49$} & $+0.00$ & \canvasgainbest{$+0.25$} \\
\midrule
Decoupled & All CS / Korean   & \canvasgain{$+0.63$} & \canvasgain{$+0.19$} & \canvasgainbest{$+0.41$} \\
Decoupled & All CS / Spanish  & \canvasgain{$+0.41$} & \canvasgain{$+0.57$} & \canvasgainbest{$+0.49$} \\
\bottomrule
\end{tabular}
\end{adjustbox}
\caption{\textbf{Answer-language decoupling.} $\Delta$F1 (\textsc{CANVAS}$-$\textsc{Base}) for Qwen3.5-9B when English remains the source/frame reference but the required answer language is Korean or Spanish (gold translated accordingly). \emph{Original} rows are the English-answer results of the corresponding language groups in Table~\ref{tab:app_language_canvas_mitigation}'s protocol.}
\label{tab:app_answer_decoupling}
\end{table}

\definecolor{CanvasSrc}{HTML}{2D6A4F}
\definecolor{CanvasTgt}{HTML}{7A4EAB}
\definecolor{CanvasBase}{HTML}{B23A48}
\definecolor{CanvasGood}{HTML}{1B7F5A}
\definecolor{CanvasBg}{HTML}{F7F8FA}

\subsection{CANVAS Case Studies}
\label{app:canvas_case_study}

\paragraph{Overview.}
We pair one recovery with one failure for the same Llama-3.1-8B model. Each figure combines the CS question and answer with layer-wise cosine similarity to the SRC and TGT anchors and the resulting anchor bias. The pair shows what CANVAS changes internally and also bounds the claim: source-ward movement is useful on average, but it is not an answer verifier.

\paragraph{Success Case}
\label{app:canvas_success_case}

\begin{figure*}[t]
\centering
\begin{tcolorbox}[
  enhanced,
  width=0.96\textwidth,
  colback=CanvasBg,
  colframe=black!18,
  boxrule=0.5pt,
  arc=1.5mm,
  left=2mm,right=2mm,top=1.5mm,bottom=1.5mm,
  fonttitle=\bfseries,
  coltitle=black,
  title={Success case: CANVAS fixes a CS answer (Llama-3.1-8B, Korean--English)}
]
\small
\begin{tabularx}{\linewidth}{@{}>{\bfseries}p{0.18\linewidth}X@{}}
Model        & Llama-3.1-8B-Instruct \\[1pt]
CS input     & S\"{u}leymaniye는 대화재에서 몇 년도에 damage되었나요? \\[1pt]
             & \textit{(``In what year was the S\"{u}leymaniye [Mosque] damaged in the Great Fire?'')} \\[1pt]
Gold answer  & \textbf{1660} \\
\end{tabularx}

\vspace{1.5mm}
\begin{tabularx}{\linewidth}{@{}p{0.22\linewidth}p{0.22\linewidth}p{0.52\linewidth}@{}}
\cellcolor{CanvasBase!9}\textbf{Base CS answer}
& \cellcolor{CanvasGood!9}\textbf{CANVAS answer}
& \cellcolor{black!5}\textbf{Diagnostics} \\[2pt]
\cellcolor{CanvasBase!9}\large\textcolor{CanvasBase}{2019}
& \cellcolor{CanvasGood!9}\large\textcolor{CanvasGood}{1660}
& \cellcolor{black!5}$\alpha{=}0.75$;\
  $\gamma{=}{-}0.396$;\
  EN tokens$\,{=}\,6$;\
  KO tokens$\,{=}\,11$;\
  steered layers$\,{=}\,10$ (layers 22--31 of 32) \\
\end{tabularx}

\vspace{1mm}
\footnotesize
\textbf{Token breakdown.}
The question contains 11 Korean tokens (TGT) and only 6 English tokens (SRC),
giving $\gamma{=}{-}0.396$ (strongly TGT-anchored).
CANVAS detects this imbalance and applies $\alpha{=}0.75$ interpolation
toward the English anchor in the top 10 layers.
\end{tcolorbox}

\vspace{2mm}

\includegraphics[width=0.96\textwidth]{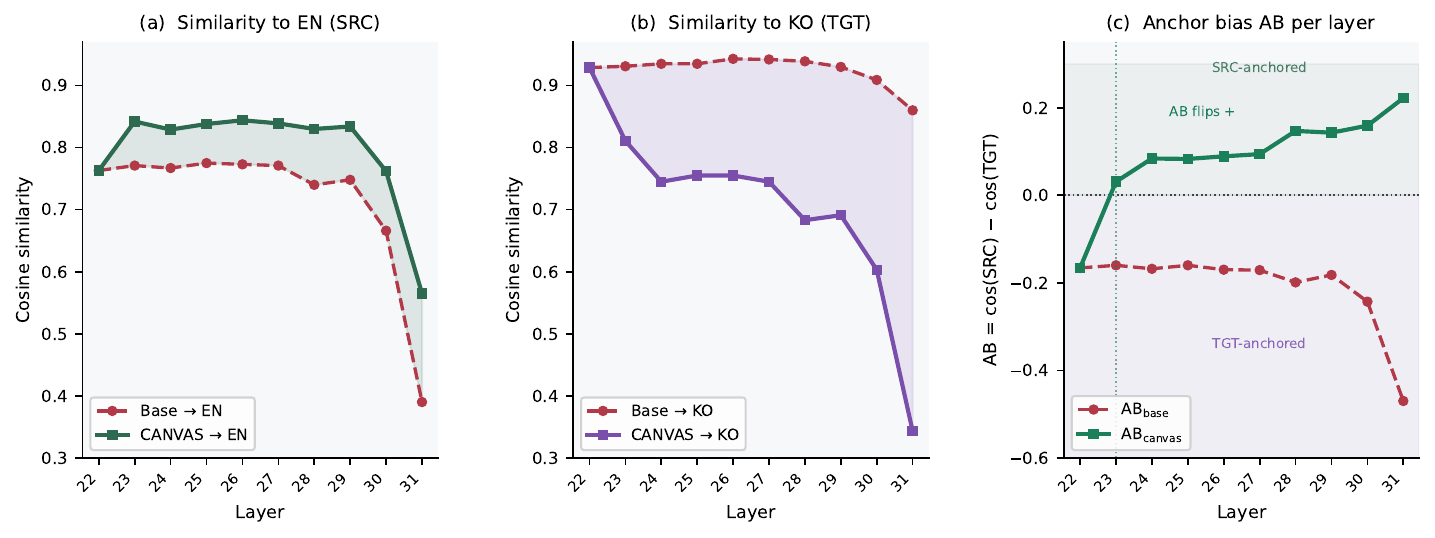}

\caption{%
  \textbf{CANVAS success case: representation trajectory.}
  \textit{Top}: qualitative example showing the wrong base answer (2019) and the
  correct CANVAS answer (1660) for a Korean--English CS question.
  \textit{Bottom (a--c)}: layer-by-layer hidden-state cosine similarity to the
  EN anchor (SRC) and KO anchor (TGT) for the CS prompt, and the resulting
  anchor bias $\mathrm{AB}{=}\cos(\mathbf{h}_\mathrm{EN},\mathbf{h}_\mathrm{CS})
  -\cos(\mathbf{h}_\mathrm{KO},\mathbf{h}_\mathrm{CS})$.
  In the base run, $\mathrm{AB}$ stays negative throughout (KO-anchored).
  CANVAS flips $\mathrm{AB}$ to positive from layer~23 onward (EN-anchored),
  which is sufficient for the model to recover the correct answer.%
}
\label{fig:app_canvas_case_study}
\end{figure*}

The success case uses a Korean--English CS question with 11 Korean tokens and 6 English tokens. Its alignment score is strongly target-oriented ($\gamma{=}{-}0.396$), so CANVAS selects $\alpha{=}0.75$ and steers Korean-token hidden states in layers 22--31. Figure~\ref{fig:app_canvas_case_study} shows that direct CS answering returns \textcolor{CanvasBase}{\textbf{2019}}, while CANVAS recovers the gold year \textcolor{CanvasGood}{\textbf{1660}}.

The lower panels show the internal correction behind the answer change. Similarity to the EN anchor rises under CANVAS, similarity to the KO anchor falls across the steered layers, and anchor bias flips from negative to positive from layer~23 onward. This example, therefore, connects the intended source-ward intervention to a concrete answer recovery.

\paragraph{Failure Case}
\label{app:canvas_failure_case}

\begin{figure*}[t]
\centering
\begin{tcolorbox}[
  enhanced,
  width=0.96\textwidth,
  colback=CanvasBg,
  colframe=black!18,
  boxrule=0.5pt,
  arc=1.5mm,
  left=2mm,right=2mm,top=1.5mm,bottom=1.5mm,
  fonttitle=\bfseries,
  coltitle=black,
  title={Failure case: CANVAS changes a correct CS answer (Llama-3.1-8B, Spanish--English)}
]
\small
\begin{tabularx}{\linewidth}{@{}>{\bfseries}p{0.18\linewidth}X@{}}
Model        & Llama-3.1-8B-Instruct \\[1pt]
CS input     & En qu\'e a\~no la dise\~nadora de moda Mowalola Ogunlesi abandon\'o Central Saint Martins? \\[1pt]
             & \textit{(``In what year did fashion designer Mowalola Ogunlesi drop out of Central Saint Martins?'')} \\[1pt]
Gold answer  & \textbf{2018} \\
\end{tabularx}

\vspace{1.5mm}
\begin{tabularx}{\linewidth}{@{}p{0.22\linewidth}p{0.22\linewidth}p{0.52\linewidth}@{}}
\cellcolor{CanvasGood!9}\textbf{Base CS answer}
& \cellcolor{CanvasBase!9}\textbf{CANVAS answer}
& \cellcolor{black!5}\textbf{Diagnostics} \\[2pt]
\cellcolor{CanvasGood!9}\large\textcolor{CanvasGood}{2018}
& \cellcolor{CanvasBase!9}\large\textcolor{CanvasBase}{2017}
& \cellcolor{black!5}$\alpha{=}0.75$;\
  $\gamma{=}{-}0.234$;\
  EN tokens$\,{=}\,6$;\
  ES tokens$\,{=}\,14$;\
  steered layers$\,{=}\,10$ (layers 22--31 of 32) \\
\end{tabularx}

\vspace{1mm}
\footnotesize
\textbf{Token breakdown.}
The question contains 14 Spanish tokens (TGT) and 6 English tokens (SRC),
so CANVAS applies its strongest source-ward interpolation.
Here the direct CS answer is already correct, but the intervention changes the
generated year.
\end{tcolorbox}

\vspace{2mm}

\includegraphics[width=0.96\textwidth]{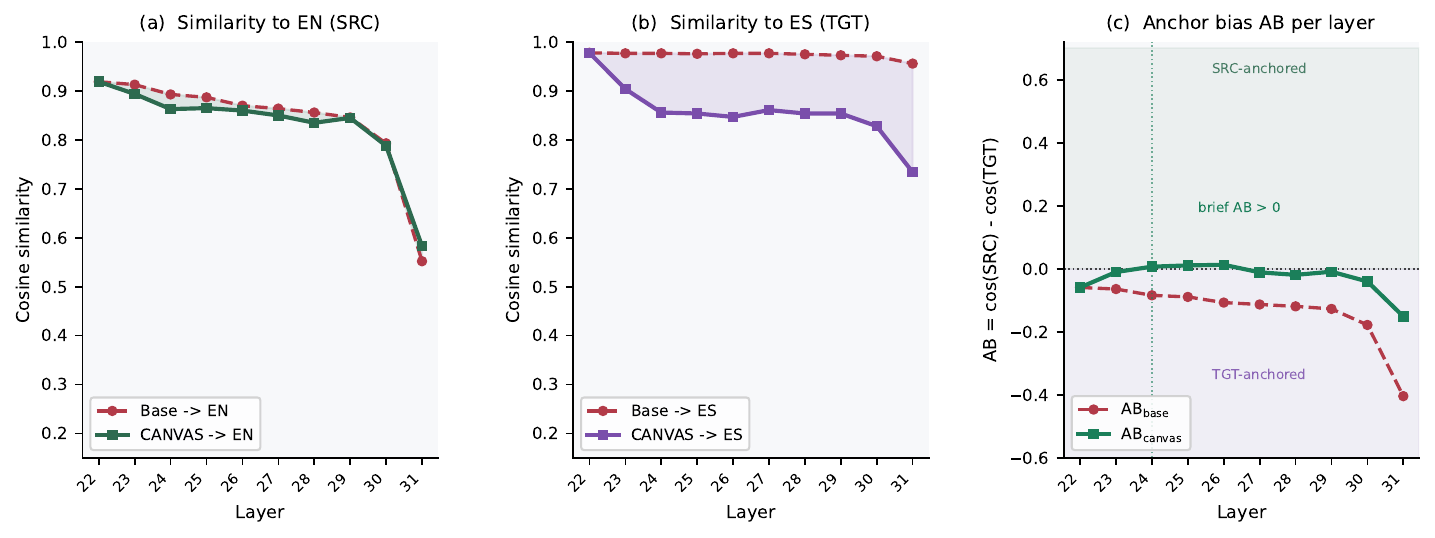}

\caption{%
  \textbf{CANVAS failure case: representation trajectory.}
  \textit{Top}: qualitative example where direct CS answering returns the gold
  year (2018), while CANVAS outputs a wrong year (2017).
  \textit{Bottom (a--c)}: layer-by-layer cosine similarity to the EN anchor
  (SRC), similarity to the Spanish anchor (TGT), and anchor bias
  $\mathrm{AB}{=}\cos(\mathbf{h}_\mathrm{EN},\mathbf{h}_\mathrm{CS})
  -\cos(\mathbf{h}_\mathrm{ES},\mathbf{h}_\mathrm{CS})$.
  CANVAS moves the CS state toward the SRC side but does not produce a stable
  final source-oriented state, while also changing an already correct answer.%
}
\label{fig:app_canvas_failure_case_study}
\end{figure*}

The failure case uses a Spanish--English GF-Tgt question with 14 Spanish tokens and 6 English tokens. CANVAS sees a target-oriented alignment score ($\gamma{=}{-}0.234$) and applies its maximal interpolation ($\alpha{=}0.75$). In Figure~\ref{fig:app_canvas_failure_case_study}, direct CS answering already returns the gold year \textcolor{CanvasGood}{\textbf{2018}}, but CANVAS changes the answer to \textcolor{CanvasBase}{\textbf{2017}}.

The layer trace explains the failure boundary. CANVAS reduces Spanish similarity and pushes anchor bias toward the source side, but the shift is incomplete at the top of the stack: anchor bias briefly crosses zero in the early steered layers and returns negative by the final layer. At the same time, the direct CS state is already sufficient for the correct answer. The strong source-ward correction, therefore, perturbs an already correct answer state without producing a stable final source-oriented state. This case shows why source-anchor correction improves CS answering on average, but is not a correctness certificate for every example.

\subsection{Model-wise CANVAS Analysis}
\label{app:modelwise_canvas_repr}
\paragraph{Setup.}
We group the CANVAS representation analyses by model. For each model, we compare the code-switched representation before and after intervention against two reference anchors: the source-language question representation and the target-language question representation. We compute the metrics over the control-layer interval $\mathcal{L}_{\mathrm{ctrl}}$ and mean-pool question-content tokens, so the comparison focuses on the question state rather than the generated answer.

\paragraph{Metrics.}
Table~\ref{tab:app_canvas_modelwise_repr} reports the model-wise movement signals. $\Delta\mathrm{AB}$ measures whether CANVAS increases source-oriented anchor bias. $\Delta\cos_{\mathrm{SRC}}$ measures the change in similarity to the SRC and TGT anchors. The projection ratio $\eta$ measures how much of the intervention displacement lies along the SRC--TGT axis; a positive $\eta$ means that the movement is in the SRC direction.

\begin{figure*}[t]
  \centering
  \includegraphics[width=0.95\textwidth]{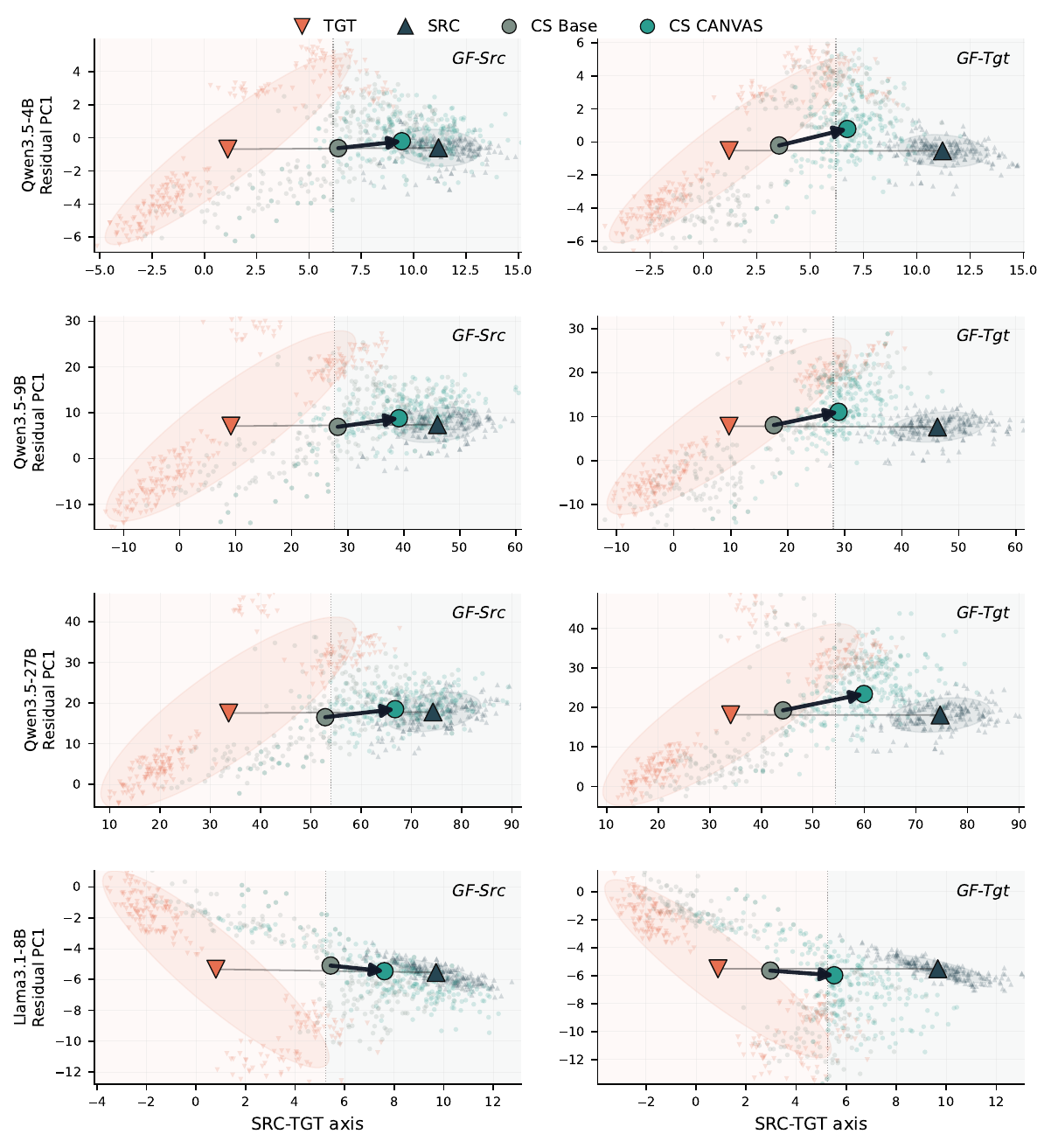}
  \caption{\textbf{Model-wise CANVAS representation movement.}
  We visualize the same representation analysis for four representative models. Each row is a model and each column is a grammar-forced CS condition. Triangles mark SRC/TGT anchors, gray points mark base CS states, teal points mark CANVAS states, and arrows connect the base and CANVAS centroids. Across models, CANVAS shifts the CS centroid toward the source-side region of the SRC--TGT axis.}
  \label{fig:app_canvas_modelwise_repr}
\end{figure*}

\paragraph{Results.}
Across analyzed models, CANVAS consistently increases AB, increases cosine similarity to the SRC anchor, and yields positive projection ratios on the SRC--TGT axis. Figure~\ref{fig:app_canvas_modelwise_repr} shows the same analysis as a model-wise representation grid. Each row corresponds to one model, and each column corresponds to a grammar-forced CS condition. The arrows summarize the movement from the base CS centroid to the CANVAS centroid, making it possible to compare whether the same source-directed shift appears across model families.

\paragraph{Headroom and gain variability.}
Models differ in how much they gain from CANVAS (e.g., Mixtral-8x7B and Llama-3.1-70B are not individually significant in \Cref{tab:canvas_results}). Using each model's full predictions, we relate its gain to three quantities (\Cref{tab:app_headroom}): \emph{oracle headroom}, defined as the F1 of the same question presented entirely in the source language minus direct CS F1 before intervention; initial alignment $\gamma$; and mean steering strength $\alpha$. Gain is most strongly associated with headroom ($\rho=0.49$), whereas initial alignment ($\rho=0.22$) and $\alpha$ ($\rho=-0.21$) are weakly predictive; Llama-3.1-70B, for instance, has far less headroom ($+3.30$) than Llama-3.1-8B ($+16.88$) or Qwen3.5-27B ($+12.96$). Model-wise variability thus mainly reflects how much useful source-side behavior there is to recover, not model scale or steering strength. It also explains why models with similar mean gains can differ in paired significance: for Mixtral-8x7B and Llama-3.1-70B the improvement is concentrated in a few large per-question gains rather than spread consistently across questions.

\begin{table}[t]
\centering
\small
\setlength{\tabcolsep}{4pt}
\renewcommand{\arraystretch}{1.04}
\begin{adjustbox}{max width=\columnwidth}
\begin{tabular}{@{}lccc@{}}
\toprule
\textbf{Model} & \textbf{Oracle headroom} & \textbf{Initial alignment $\gamma$} & \textbf{Mean $\alpha$} \\
\midrule
Mixtral-8x7B  & $+9.31$  & $-0.08$ & 0.56 \\
Llama-3.1-70B & $+3.30$  & $-0.07$ & 0.50 \\
Llama-3.1-8B  & $+16.88$ & $-0.11$ & 0.54 \\
Qwen3.5-27B   & $+12.96$ & $-0.08$ & 0.53 \\
\bottomrule
\end{tabular}
\end{adjustbox}
\caption{\textbf{Model-wise diagnostics of \textsc{CANVAS} gains.} Oracle headroom is the F1 of the pure source-language question minus direct CS F1 before intervention.}
\label{tab:app_headroom}
\end{table}

\begin{figure*}[t]
  \centering
  \includegraphics[width=0.92\textwidth]{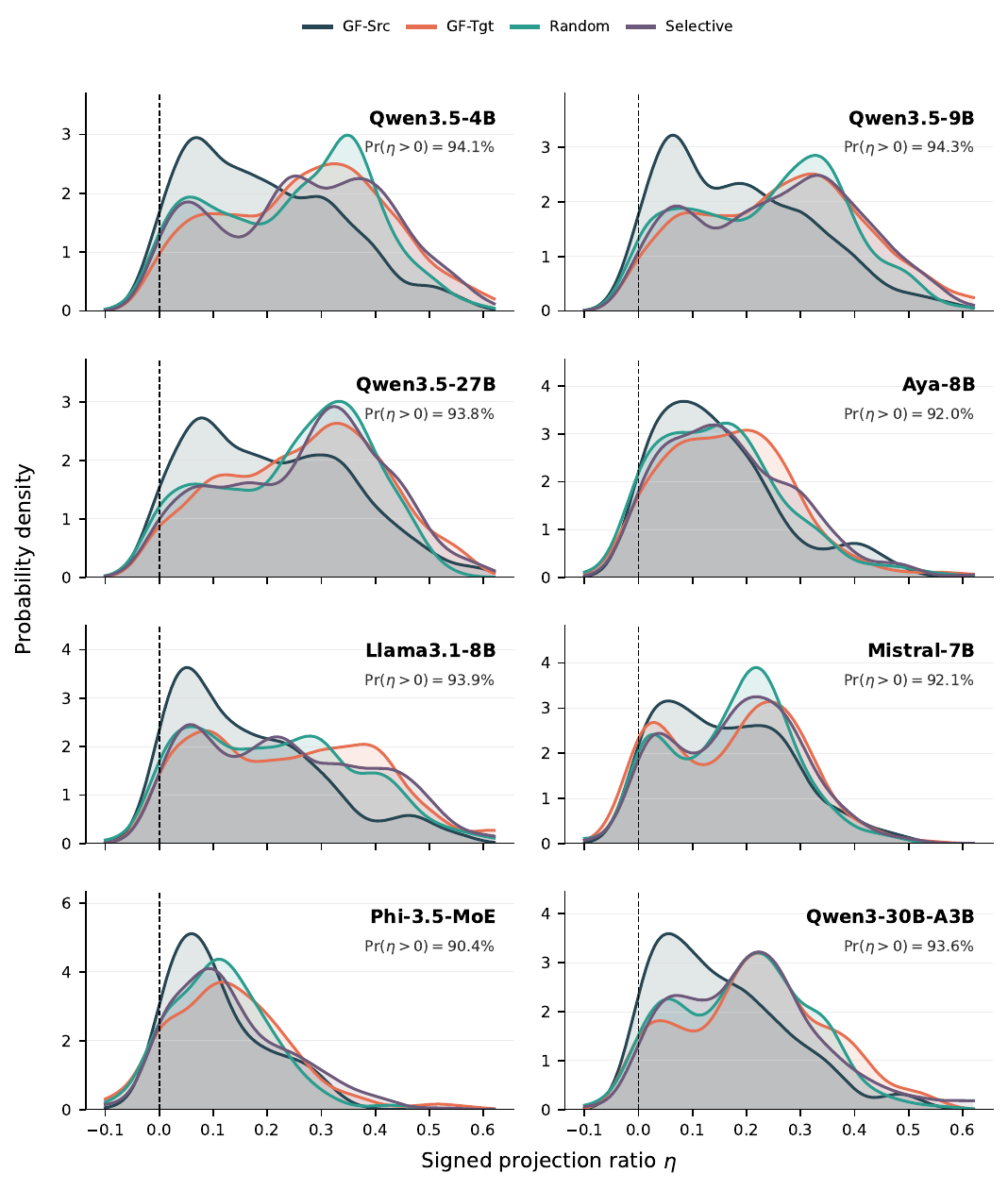}
  \caption{\textbf{Model-wise source-ward representation movement.}
  We plot the signed projection ratio $\eta$ of the CANVAS displacement onto the SRC--TGT anchor axis for each analyzed model. Positive values indicate movement toward the source anchor. Each panel reports the percentage of examples with $\eta>0$.}
  \label{fig:app_canvas_eta_modelwise}
\end{figure*}

\begin{figure*}[t]
  \centering
  \includegraphics[width=0.92\textwidth]{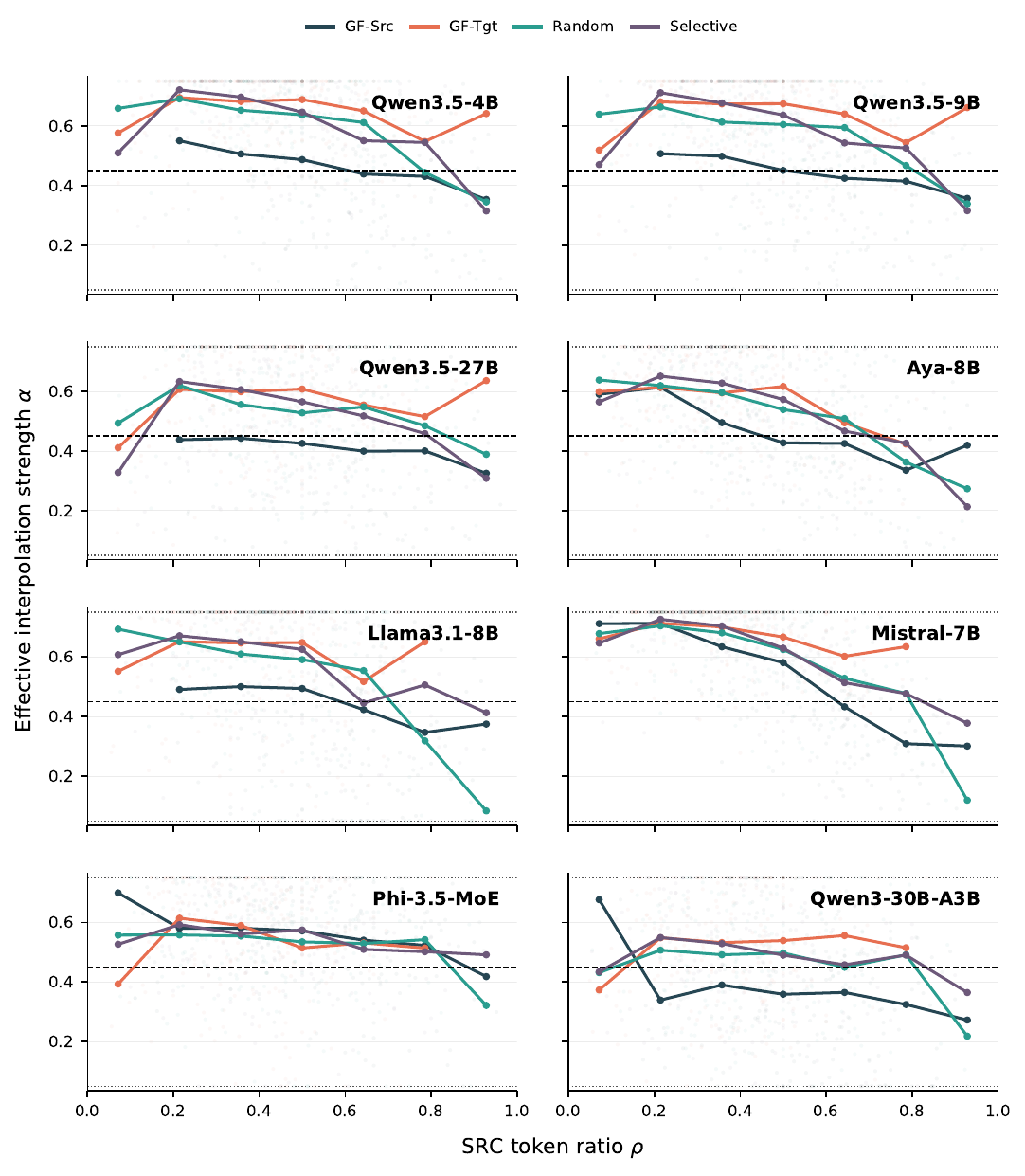}
  \caption{\textbf{Model-wise adaptive intervention strength.}
  We compare the source-token ratio $\rho$ with the effective interpolation strength $\alpha$ selected by CANVAS for each analyzed model. Lower $\rho$ corresponds to more target-heavy inputs, and the binned trend lines show how the adaptive rule changes intervention strength within each model.}
  \label{fig:app_canvas_alpha_modelwise}
\end{figure*}

\paragraph{Model-wise diagnostics.}
We provide model-wise versions of the two main CANVAS diagnostics. Figure~\ref{fig:app_canvas_eta_modelwise} decomposes the signed projection ratio $\eta$ by model and CS condition. The mass of each density remains mostly to the right of zero, and $\Pr(\eta>0)$ stays above 90\% for every analyzed model. Figure~\ref{fig:app_canvas_alpha_modelwise} decomposes the adaptive controller by model. Within each model, a lower source-token ratio $\rho$ generally receives a larger effective interpolation strength $\alpha$, showing that the validation-selected schedule remains adaptive.

\begin{table*}[t]
\centering
\small
\setlength{\tabcolsep}{4pt}
\renewcommand{\arraystretch}{1.03}
\begin{tabular}{lrrrrr}
\toprule
\textbf{Model} &
$\Delta$AB &
$\Delta\cos_{\mathrm{SRC}}$ &
$\eta$ &
$\Pr(\eta>0)$ &
Mean $\alpha$ \\
\midrule
\modelwithlogo{cohere}{Aya-Expanse-8B} 
& 0.143 & 0.036 & 0.160 & 92.0 & 0.542 \\

\modelwithlogo{meta}{Llama3.1-8B} 
& 0.159 & 0.059 & 0.214 & 93.9 & 0.562 \\

\modelwithlogo{mistral}{Mistral-7B} 
& 0.224 & 0.087 & 0.171 & 92.1 & 0.613 \\

\modelwithlogo{microsoft}{Phi-3.5-MoE} 
& 0.071 & 0.026 & 0.121 & 90.4 & 0.548 \\

\modelwithlogo{qwen}{Qwen3.5-27B} 
& 0.111 & 0.032 & 0.248 & 93.8 & 0.523 \\

\modelwithlogo{qwen}{Qwen3.5-4B} 
& 0.156 & 0.050 & 0.243 & 94.1 & 0.594 \\

\modelwithlogo{qwen}{Qwen3.5-9B} 
& 0.139 & 0.047 & 0.238 & 94.3 & 0.576 \\

\modelwithlogo{qwen}{Qwen3-30B-A3B} 
& 0.088 & 0.029 & 0.192 & 93.6 & 0.469 \\
\bottomrule
\end{tabular}
\caption{\textbf{Model-wise representation movement under \textsc{CANVAS}.}
$\Delta\mathrm{AB}$ measures the increase in source-oriented anchor bias, and $\Delta\cos_{\mathrm{SRC}}$ measures the change in similarity to the source-language anchor. $\eta$ measures the signed projection of the displacement on the SRC--TGT axis, and $\Pr(\eta>0)$ reports the percentage of examples whose displacement has a source-ward component.}
\label{tab:app_canvas_modelwise_repr}
\end{table*}

\begin{table*}[t]
\centering
\scriptsize
\setlength{\tabcolsep}{2.15pt}
\renewcommand{\arraystretch}{1.02}
\begin{adjustbox}{max width=\textwidth}
\begin{tabular}{@{}lrrr!{\hspace{3pt}\color{GroupSep}\vrule width 0.25pt\hspace{3pt}}rrr!{\hspace{3pt}\color{GroupSep}\vrule width 0.25pt\hspace{3pt}}rrr!{\hspace{3pt}\color{GroupSep}\vrule width 0.25pt\hspace{3pt}}rr@{}}
\toprule
\multirow{2}{*}{\textbf{Model}} &
\multicolumn{3}{c}{\textbf{Avg. F1}} &
\multicolumn{3}{c}{\textbf{Mean latency}} &
\multicolumn{3}{c}{\textbf{Generation diagnostics}} &
\multicolumn{2}{c}{\textbf{Tail latency}} \\
\cmidrule(lr){2-4} \cmidrule(lr){5-7} \cmidrule(lr){8-10} \cmidrule(lr){11-12}
& \textbf{\textsc{Base}} & \textbf{\textsc{CANVAS}} & $\Delta$F1
& Base ms & CANVAS ms & Cost$\times$
& Base len. & CANVAS len. & Faster ex.
& Base p95 & CANVAS p95 \\
\midrule
\modelwithlogo{cohere}{Aya-Expanse-8B} & 5.99 & 6.88 & \canvasgain{+0.89} & 334.8 & 404.2 & 1.21 & 4.3 & 5.0 & 31.1\% & 1216 & 1554 \\
\modelwithlogo{mistral}{Mistral-7B} & 4.53 & 5.06 & \canvasgain{+0.54} & 644.9 & 583.2 & 0.90 & 8.1 & 7.6 & 37.5\% & 1633 & 1576 \\
\modelwithlogo{qwen}{Qwen3.5-4B} & 7.58 & 8.14 & \canvasgain{+0.56} & 444.3 & 528.7 & 1.19 & 2.6 & 2.4 & 26.7\% & 862 & 888 \\
\modelwithlogo{qwen}{Qwen3.5-9B} & 9.89 & 10.24 & \canvasgain{+0.35} & 442.1 & 544.7 & 1.23 & 2.4 & 2.3 & 18.7\% & 777 & 898 \\
\modelwithlogo{qwen}{Qwen3.5-27B} & 14.14 & 15.40 & \canvasgain{+1.25} & 975.1 & 1166.4 & 1.20 & 2.8 & 2.7 & 17.1\% & 1855 & 2100 \\
\modelwithlogo{meta}{Llama3.2-1B} & 2.57 & 3.20 & \canvasgain{+0.63} & 137.4 & 138.3 & 1.01 & 3.3 & 3.1 & 38.9\% & 460 & 399 \\
\modelwithlogo{meta}{Llama3.1-8B} & 6.21 & 7.79 & \canvasgain{+1.59} & 279.2 & 288.6 & 1.03 & 3.5 & 3.1 & 27.8\% & 755 & 700 \\
\modelwithlogo{meta}{Llama3.1-70B} & 12.89 & 13.20 & \canvasgain{+0.31} & 1207.7 & 1518.0 & 1.26 & 2.4 & 2.4 & 13.8\% & 2212 & 2622 \\
\modelwithlogo{meta}{Llama3.3-70B} & 12.96 & 13.61 & \canvasgain{+0.65} & 1157.1 & 1466.7 & 1.27 & 2.3 & 2.3 & 13.3\% & 1974 & 2345 \\
\midrule
\modelwithlogo{mistral}{Mixtral-8x7B} & 6.77 & 6.91 & \canvasgain{+0.13} & 3167.0 & 1853.3 & 0.59 & 13.3 & 12.6 & 86.2\% & 5762 & 3467 \\
\modelwithlogo{microsoft}{Phi-3.5-MoE} & 7.99 & 8.82 & \canvasgain{+0.83} & 757.0 & 824.6 & 1.09 & 4.2 & 4.1 & 11.5\% & 2337 & 2331 \\
\modelwithlogo{qwen}{Qwen3-30B-A3B} & 6.91 & 8.11 & \canvasgain{+1.20} & 653.2 & 905.0 & 1.39 & 2.9 & 2.9 & 5.4\% & 1416 & 1784 \\
\modelwithlogo{qwen}{Qwen3.6-35B-A3B} & 15.80 & 17.53 & \canvasgain{+1.73} & 756.7 & 1047.7 & 1.38 & 2.3 & 2.2 & 2.3\% & 1177 & 1619 \\
\midrule
\rowcolor{MeanGray} \textbf{\textit{Mean}} & 8.79 & 9.61 & \canvasgain{+0.82} & 842.8 & 866.9 & 1.13 & 4.2 & 4.1 & 25.4\% & 1726 & 1714 \\
\bottomrule
\end{tabular}
\end{adjustbox}
\caption{\textbf{CANVAS cost analysis.}
We measure empirical wall-clock latency per example for direct CS answering and CANVAS on the same evaluation set. Cost$\times$ is CANVAS latency divided by direct CS latency. Length is a lightweight token proxy computed from generated answer strings, Faster ex. is the percentage of examples where CANVAS is faster than direct CS answering, and p95 reports the 95th-percentile latency in milliseconds.}
\label{tab:app_canvas_cost}
\end{table*}

\begin{table*}[t]
\caption{\textbf{Anchor-bias metric variants.} $\textsc{AB}_{raw}$ is the question-content-only cosine anchor bias ($\cos(\mathbf{h}_\mathrm{SRC},\mathbf{h}_\mathrm{CS})-\cos(\mathbf{h}_\mathrm{TGT},\mathbf{h}_\mathrm{CS})$, mean-pooled over question tokens). $\textsc{AB}_{nrm}$ divides $\textsc{AB}_{raw}$ by $1{-}b_m$, the model-specific random-cosine dynamic range. $\textsc{AB}_{tok}$ L2-normalizes each token vector before pooling; $\textsc{AB}_{tok,nrm}$ additionally applies the same normalization. Main tables report $\textsc{AB}_{nrm}$. All values are verified against source data.}
\centering
\small
\setlength{\tabcolsep}{2.35pt}
\renewcommand{\arraystretch}{0.98}
\begin{adjustbox}{max width=\textwidth}
\begin{tabular}{@{}l r c!{\hspace{3pt}\color{GroupSep}\vrule width 0.25pt\hspace{3pt}}rr!{\hspace{3pt}\color{GroupSep}\vrule width 0.25pt\hspace{3pt}}rr@{}}
\toprule
\multirow{2}{*}{\raisebox{-0.55ex}{\textbf{Model}}} &
\multirow{2}{*}{\raisebox{-0.55ex}{$1{-}b_m$}} &
\multirow{2}{*}{\raisebox{-0.55ex}{\textbf{Cond.}}} &
\multicolumn{2}{c}{\textbf{Mean-pool}} &
\multicolumn{2}{c}{\textbf{Token-norm}} \\
\cmidrule(lr){4-5} \cmidrule(lr){6-7}
& & & $\textsc{AB}_{raw}$ & $\textsc{AB}_{nrm}$ & $\textsc{AB}_{tok}$ & $\textsc{AB}_{tok,nrm}$ \\
\midrule
\multirow{4}{*}{\modelwithlogo{qwen}{Qwen3.5-4B}} & \multirow{4}{*}{0.0636}
 & \textsc{GF-Src} & \canvasgain{+0.038} & \canvasgain{+0.59}   & \canvasgain{+0.042} & \canvasgain{+0.66} \\
 & & \textsc{GF-Tgt}  & \canvasloss{$-$0.279} & \canvasloss{$-$4.39}  & \canvasloss{$-$0.276} & \canvasloss{$-$4.33} \\
 & & \textsc{Rand}    & \canvasloss{$-$0.142} & \canvasloss{$-$2.24}  & \canvasloss{$-$0.138} & \canvasloss{$-$2.17} \\
 & & \textsc{Sel}     & \canvasloss{$-$0.196} & \canvasloss{$-$3.08}  & \canvasloss{$-$0.193} & \canvasloss{$-$3.03} \\
\midrule
\multirow{4}{*}{\modelwithlogo{mistral}{Mistral-7B}} & \multirow{4}{*}{0.0454}
 & \textsc{GF-Src} & \canvasloss{$-$0.175} & \canvasloss{$-$3.86}  & \canvasloss{$-$0.160} & \canvasloss{$-$3.51} \\
 & & \textsc{GF-Tgt}  & \canvasloss{$-$0.450} & \canvasloss{$-$9.90}  & \canvasloss{$-$0.440} & \canvasloss{$-$9.69} \\
 & & \textsc{Rand}    & \canvasloss{$-$0.342} & \canvasloss{$-$7.52}  & \canvasloss{$-$0.328} & \canvasloss{$-$7.23} \\
 & & \textsc{Sel}     & \canvasloss{$-$0.366} & \canvasloss{$-$8.06}  & \canvasloss{$-$0.354} & \canvasloss{$-$7.79} \\
\midrule
\multirow{4}{*}{\modelwithlogo{meta}{Llama3.1-8B}} & \multirow{4}{*}{0.0254}
 & \textsc{GF-Src} & \canvasloss{$-$0.011} & \canvasloss{$-$0.43}  & \canvasloss{$-$0.015} & \canvasloss{$-$0.60} \\
 & & \textsc{GF-Tgt}  & \canvasloss{$-$0.381} & \canvasloss{$-$14.97} & \canvasloss{$-$0.385} & \canvasloss{$-$15.12} \\
 & & \textsc{Rand}    & \canvasloss{$-$0.235} & \canvasloss{$-$9.25}  & \canvasloss{$-$0.239} & \canvasloss{$-$9.40} \\
 & & \textsc{Sel}     & \canvasloss{$-$0.277} & \canvasloss{$-$10.87} & \canvasloss{$-$0.281} & \canvasloss{$-$11.03} \\
\midrule
\multirow{4}{*}{\modelwithlogo{cohere}{Aya-Expanse-8B}} & \multirow{4}{*}{0.0719}
 & \textsc{GF-Src} & \canvasgain{+0.030} & \canvasgain{+0.42}   & \canvasgain{+0.041} & \canvasgain{+0.57} \\
 & & \textsc{GF-Tgt}  & \canvasloss{$-$0.434} & \canvasloss{$-$6.04}  & \canvasloss{$-$0.427} & \canvasloss{$-$5.94} \\
 & & \textsc{Rand}    & \canvasloss{$-$0.273} & \canvasloss{$-$3.79}  & \canvasloss{$-$0.262} & \canvasloss{$-$3.65} \\
 & & \textsc{Sel}     & \canvasloss{$-$0.316} & \canvasloss{$-$4.40}  & \canvasloss{$-$0.308} & \canvasloss{$-$4.28} \\
\midrule
\multirow{4}{*}{\modelwithlogo{qwen}{Qwen3.5-9B}} & \multirow{4}{*}{0.0211}
 & \textsc{GF-Src} & \canvasgain{+0.036} & \canvasgain{+1.72}   & \canvasgain{+0.037} & \canvasgain{+1.75} \\
 & & \textsc{GF-Tgt}  & \canvasloss{$-$0.272} & \canvasloss{$-$12.93} & \canvasloss{$-$0.267} & \canvasloss{$-$12.68} \\
 & & \textsc{Rand}    & \canvasloss{$-$0.138} & \canvasloss{$-$6.57}  & \canvasloss{$-$0.135} & \canvasloss{$-$6.40} \\
 & & \textsc{Sel}     & \canvasloss{$-$0.187} & \canvasloss{$-$8.86}  & \canvasloss{$-$0.183} & \canvasloss{$-$8.69} \\
\midrule
\multirow{4}{*}{\modelwithlogo{qwen}{Qwen3.5-27B}} & \multirow{4}{*}{0.0206}
 & \textsc{GF-Src} & \canvasgain{+0.018} & \canvasgain{+0.87}   & \canvasgain{+0.019} & \canvasgain{+0.94} \\
 & & \textsc{GF-Tgt}  & \canvasloss{$-$0.290} & \canvasloss{$-$14.10} & \canvasloss{$-$0.284} & \canvasloss{$-$13.76} \\
 & & \textsc{Rand}    & \canvasloss{$-$0.155} & \canvasloss{$-$7.54}  & \canvasloss{$-$0.151} & \canvasloss{$-$7.34} \\
 & & \textsc{Sel}     & \canvasloss{$-$0.202} & \canvasloss{$-$9.78}  & \canvasloss{$-$0.197} & \canvasloss{$-$9.55} \\
\midrule
\multirow{4}{*}{\modelwithlogo{meta}{Llama3.1-70B}} & \multirow{4}{*}{0.0203}
 & \textsc{GF-Src} & \canvasloss{$-$0.104} & \canvasloss{$-$5.15}  & \canvasloss{$-$0.104} & \canvasloss{$-$5.13} \\
 & & \textsc{GF-Tgt}  & \canvasloss{$-$0.472} & \canvasloss{$-$23.29} & \canvasloss{$-$0.469} & \canvasloss{$-$23.14} \\
 & & \textsc{Rand}    & \canvasloss{$-$0.338} & \canvasloss{$-$16.66} & \canvasloss{$-$0.335} & \canvasloss{$-$16.53} \\
 & & \textsc{Sel}     & \canvasloss{$-$0.374} & \canvasloss{$-$18.44} & \canvasloss{$-$0.371} & \canvasloss{$-$18.31} \\
\midrule
\multirow{4}{*}{\modelwithlogo{meta}{Llama3.3-70B}} & \multirow{4}{*}{0.0186}
 & \textsc{GF-Src} & \canvasloss{$-$0.060} & \canvasloss{$-$3.24}  & \canvasloss{$-$0.061} & \canvasloss{$-$3.27} \\
 & & \textsc{GF-Tgt}  & \canvasloss{$-$0.446} & \canvasloss{$-$23.99} & \canvasloss{$-$0.443} & \canvasloss{$-$23.83} \\
 & & \textsc{Rand}    & \canvasloss{$-$0.295} & \canvasloss{$-$15.90} & \canvasloss{$-$0.293} & \canvasloss{$-$15.79} \\
 & & \textsc{Sel}     & \canvasloss{$-$0.337} & \canvasloss{$-$18.11} & \canvasloss{$-$0.334} & \canvasloss{$-$17.99} \\
\bottomrule
\end{tabular}
\end{adjustbox}
\label{tab:anchor_bias_variants}
\end{table*}

\subsection{CANVAS Cost Analysis}
\label{app:canvas_cost}

\paragraph{Setup.}
We measure the wall-clock cost of CANVAS on the same full CS evaluation set used for the main intervention results. For each model and example, we record direct CS answering latency and CANVAS latency, then compute the empirical cost ratio by dividing CANVAS latency by direct CS latency. This is an end-to-end latency measurement, not a per-token FLOP estimate.

\paragraph{Results.}
Table~\ref{tab:app_canvas_cost} reports the model-wise cost. CANVAS improves average F1 from 8.79 to 9.61 over the 13-model pool while using 1.13$\times$ the latency of direct CS answering on average. In absolute time, CANVAS takes 866.9 ms per example on average, i.e., less than one second per question. This is only 24.1 ms above the direct CS answering average in the same model pool. The overhead remains small because CANVAS reuses the same model and applies a hidden-state intervention during prefill, instead of calling an additional model.

\paragraph{Why some ratios are below one.}
Some models show empirical cost ratios below 1.0 even though CANVAS adds a clean prefill and a steered prefill. This happens because the reported latency includes autoregressive answer generation. If CANVAS changes the hidden state so that the model emits a shorter answer or reaches EOS earlier, the shorter decoding phase can offset the extra prefill pass. The integrated diagnostics in Table~\ref{tab:app_canvas_cost} support this interpretation: Mistral-7B and Mixtral-8x7B have cost ratios below one, shorter CANVAS outputs on average, and lower 95th-percentile latency under CANVAS. Mixtral-8x7B is especially clear: CANVAS is faster on 86.2\% of examples and reduces p95 latency from 5762 ms to 3467 ms. We therefore interpret ratios below one as an empirical decoding-length effect, not as a claim that CANVAS is cheaper per forward pass.

\paragraph{Interpretation.}
The main takeaway is that CANVAS provides a consistent F1 gain with modest wall-clock overhead. Since the method adds sub-second latency per example on average and does not require an external model call, we view it as practical for real-time or interactive CS question answering settings.

\subsection{Additional Movement Analysis}
\begin{figure}[t]
\centering
\includegraphics[width=\linewidth]{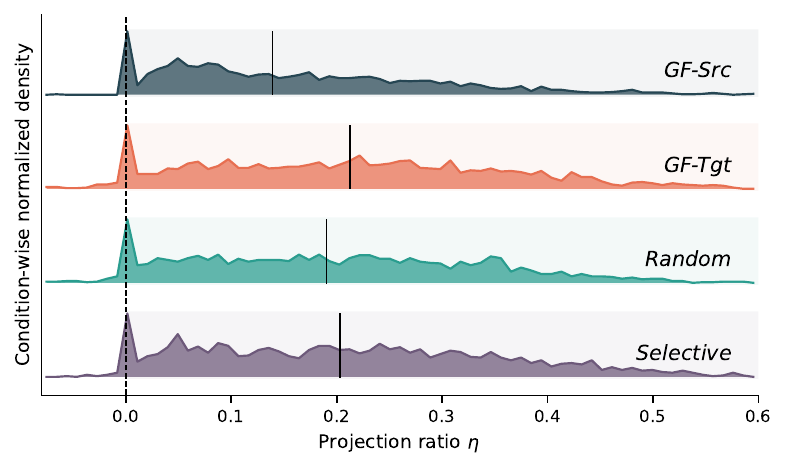}
\vspace{-7mm}
\caption{\textbf{Source-ward representation movement.} Signed projection ratio $\eta$ of the \textsc{CANVAS} displacement onto the SRC--TGT axis ($\eta>0$: source-ward; $\eta<0$: target-ward).}
\label{fig:canvas_eta}
\end{figure}

We first test whether \textsc{CANVAS} changes the internal representation in the intended direction. We aggregate examples across the eight models used for the representation analysis. For each example, we extract four question-level representations from the control layers: the source-only representation, the target-only representation, the CS representation without intervention, and after \textsc{CANVAS}. We mean-pool question-content token states within each layer and then average over the selected control layers, so the metric reflects the question representation rather than the generated answer. We measure the signed projection ratio
$\eta=\langle \mathbf{r}^{\textsc{CANVAS}}_{\mathrm{CS}}-\mathbf{r}^{\textsc{Base}}_{\mathrm{CS}},\, \mathbf{r}_{\mathrm{SRC}}-\mathbf{r}_{\mathrm{TGT}}\rangle / \|\mathbf{r}_{\mathrm{SRC}}-\mathbf{r}_{\mathrm{TGT}}\|_2^2$, where $\eta>0$ indicates source-ward movement and $\eta<0$ indicates target-ward movement. As shown in \Cref{fig:canvas_eta}, the distribution lies mostly on the positive side for all CS conditions. Across the 32 model--condition cells, \textsc{CANVAS} shifts the anchor bias \emph{toward the source side} in every cell and increases SRC similarity in every cell. Using the same upper-layer anchor bias defined in Section~\ref{sec:anchor_bias}, \textsc{CANVAS} moves $\mathrm{AB}^{\mathrm{upper}}$ source-ward by $+0.136$ on average (range $+0.071$ to $+0.224$ across the eight models; per-model values in \Cref{tab:app_canvas_modelwise_repr}), so the shift is consistently in the SRC direction. The projection metric also follows the intended direction: mean $\eta$ is positive across model--condition cells $(+0.1983)$, and $93.0\%$ of examples have $\eta>0$, indicating that most \textsc{CANVAS} updates move CS representations source-ward. These results show that \textsc{CANVAS} not only improves QA scores, but also moves CS hidden states along the intended internal anchor direction.

\subsection{Cumulative Recovery Trajectory}
\label{app:canvas_cumulative_eta}

\paragraph{Setup.}
The per-layer projection $\eta_\ell$ measures the fraction of the SRC--TGT representational gap that \textsc{CANVAS} closes at control layer $\ell$. Reading these layer-wise contributions cumulatively answers a complementary question: \emph{how far along the SRC--TGT axis does the intervention carry the CS representation by each step?} For a model with control window $\mathcal{L}_{\mathrm{ctrl}}=[\ell_0,\ell_0+N-1]$, we define the cumulative recovery at relative position $r{\in}[0,1]$ as
\[
S(r) \;=\; \sum_{\ell=\ell_0}^{\ell_0+\lfloor r(N-1)\rfloor}\eta_\ell.
\]
A value $S(r)\geq 1$ means that the intervention has accumulated a shift equal to or larger than the full SRC--TGT representational distance by relative position $r$. Using relative position rather than absolute layer count makes the trajectory comparable across models whose control windows have different lengths ($N{=}10$ for the 32-layer models, $N{=}14$ for Qwen3-30B-A3B, and $N{=}19$ for Qwen3.5-27B).

\paragraph{Aggregate trajectory.}
Table~\ref{tab:app_canvas_cum_eta_layer} reports $\bar{S}(r)$ across all $7{,}195$ examples from the eight models used in the representation analysis. The mean cumulative shift grows monotonically and reaches exactly $\bar{S}(0.5){=}1.001$ at the window midpoint, then continues to $\bar{S}(1.0){=}2.36$ by the final control layer. Per-layer contributions therefore \emph{compound} rather than cancel: by the time the intervention has used half of its allotted layers, the average representation has on aggregate already traversed the full SRC--TGT gap.

\begin{table}[t]
\centering
\small
\setlength{\tabcolsep}{6pt}
\renewcommand{\arraystretch}{1.0}
\begin{tabular}{@{}c r@{}}
\toprule
\textbf{Relative position $r$} & \textbf{Mean $S(r)$} \\
\midrule
0.0 & 0.000 \\
0.1 & 0.127 \\
0.2 & 0.306 \\
0.3 & 0.516 \\
0.4 & 0.749 \\
\rowcolor{MeanGray}
0.5 (midpoint) & 1.001 \\
0.6 & 1.261 \\
0.7 & 1.536 \\
0.8 & 1.819 \\
0.9 & 2.096 \\
1.0 (final)    & 2.359 \\
\bottomrule
\end{tabular}
\caption{\textbf{Aggregate cumulative recovery along the control window.} $r{\in}[0,1]$ is the relative position within each model's control-layer interval, and $S(r)$ is the cumulative projection up to that position averaged across $7{,}195$ examples from eight models. The mean cumulative shift reaches exactly $1.0$ at the window midpoint and continues to grow to $2.36$ by the final control layer.}
\label{tab:app_canvas_cum_eta_layer}
\end{table}

\paragraph{Crossing distribution.}
$71.7\%$ of examples cross $S{=}1$ at some point inside the control window. Among those, the median first-crossing position is $r^\ast{=}0.46$, and the mean is $0.53$ — closely aligned with the aggregate midpoint behaviour. Only $6.4\%$ cross within the first $30\%$ of the window and $37.7\%$ within the first $50\%$, so most trajectories use the second half of the control window to complete the traversal. This justifies the 30\% layer budget: a shorter span would leave the majority of examples short of full traversal, while a longer span risks perturbing lower-layer composition (see Appendix~\ref{app:canvas_layer_ablation}).

\paragraph{Model variance.}
Table~\ref{tab:app_canvas_cum_eta_model} reports the same statistics per model. Trajectory length varies widely with capacity: Qwen3.5-27B accumulates the largest final shift ($\bar{S}(1.0){=}4.72$) with $71\%$ of inputs crossing by the midpoint, while Phi-3.5-MoE accumulates the smallest ($\bar{S}(1.0){=}1.21$) with only $9\%$ crossing that early. Larger Qwen models show both higher cumulative magnitudes and earlier crossings; Llama-3.1-8B reaches the gap relatively quickly ($r^\ast{=}0.44$) but at a smaller magnitude than the 27B Qwen. Despite this spread, every model except Phi-3.5-MoE drives more than $65\%$ of its examples past full traversal.

\begin{table*}[t]
\centering
\small
\setlength{\tabcolsep}{4.5pt}
\renewcommand{\arraystretch}{1.02}
\begin{tabular}{@{}lrrrrr@{}}
\toprule
\textbf{Model} & $\bar{S}(0.5)$ & $\bar{S}(1.0)$ & \%cross by $0.5$ & \%ever cross & median $r^\ast$ \\
\midrule
\modelwithlogo{cohere}{Aya-Expanse-8B}    & 0.748 & 1.601 & 23.1\% & 65.2\% & 0.556 \\
\modelwithlogo{meta}{Llama3.1-8B}         & 1.034 & 2.140 & 40.7\% & 69.1\% & 0.444 \\
\modelwithlogo{mistral}{Mistral-7B}       & 0.770 & 1.710 & 26.1\% & 68.8\% & 0.556 \\
\modelwithlogo{microsoft}{Phi-3.5-MoE}    & 0.422 & 1.212 &  8.8\% & 53.0\% & 0.778 \\
\modelwithlogo{qwen}{Qwen3-30B-A3B}       & 1.001 & 2.687 & 40.7\% & 77.3\% & 0.462 \\
\modelwithlogo{qwen}{Qwen3.5-4B}          & 1.080 & 2.427 & 46.1\% & 76.2\% & 0.444 \\
\modelwithlogo{qwen}{Qwen3.5-9B}          & 1.067 & 2.376 & 44.9\% & 74.7\% & 0.444 \\
\modelwithlogo{qwen}{Qwen3.5-27B}         & 1.886 & 4.719 & 71.4\% & 89.1\% & 0.333 \\
\midrule
\rowcolor{MeanGray}
\textbf{\textit{Mean}} & 1.001 & 2.359 & 37.7\% & 71.7\% & 0.462 \\
\bottomrule
\end{tabular}
\caption{\textbf{Cumulative recovery per model.} $\bar{S}(r)$ is the mean cumulative projection at relative position $r$ within the control window. ``\%cross by $0.5$'' is the fraction of examples whose cumulative $S$ first reaches $1.0$ at or before the midpoint. ``\%ever cross'' is the fraction that crosses $S{=}1$ anywhere in the window. $r^\ast$ is the median relative position of the first crossing among examples that cross.}
\label{tab:app_canvas_cum_eta_model}
\end{table*}

\paragraph{Condition dependence.}
Table~\ref{tab:app_canvas_cum_eta_cond} shows the same trajectory broken down by CS condition. Cumulative shift is largest for \textsc{GF-Tgt} ($\bar{S}(1.0){=}2.58$, $76\%$ ever cross) and smallest for \textsc{GF-Src} ($\bar{S}(1.0){=}1.96$, $63\%$ ever cross), with random and selective switching in between. This ordering mirrors the adaptive $\alpha$ schedule: target-heavy inputs receive stronger interventions, and the stronger interventions translate directly into longer cumulative trajectories along the SRC--TGT axis.

\begin{table}[t]
\centering
\small
\setlength{\tabcolsep}{6pt}
\renewcommand{\arraystretch}{1.02}
\begin{tabular}{@{}lrrr@{}}
\toprule
\textbf{CS condition} & $\bar{S}(0.5)$ & $\bar{S}(1.0)$ & \%ever cross \\
\midrule
\textsc{GF-Src}    & 0.769 & 1.964 & 63.4\% \\
\textsc{GF-Tgt}    & 1.142 & 2.584 & 75.9\% \\
\textsc{Random}    & 0.996 & 2.350 & 72.6\% \\
\textsc{Selective} & 1.095 & 2.536 & 74.5\% \\
\bottomrule
\end{tabular}
\caption{\textbf{Cumulative recovery by CS condition.} Target-heavy conditions (\textsc{GF-Tgt}, \textsc{Selective}) accumulate a larger cumulative shift and cross $S{=}1$ more often than source-heavy \textsc{GF-Src}, mirroring the adaptive $\alpha$ schedule.}
\label{tab:app_canvas_cum_eta_cond}
\end{table}

\paragraph{Interpretation.}
The cumulative analysis supports three conclusions. First, per-layer interventions compound rather than cancel: by the midpoint of the control window, the mean shift already matches the full SRC--TGT gap. Second, the 30\% layer budget is not redundant: only $37.7\%$ of examples cross $S{=}1$ within the first half of the window, so the remaining layers are needed to bring the bulk of examples past full traversal. Third, the condition-level differences confirm that the adaptive controller produces \emph{representation-level} effects rather than nominal $\alpha$ adjustments alone — target-heavy inputs end the window further along the SRC--TGT axis than source-heavy inputs do, by an amount consistent with their adaptive $\alpha$.

\subsection{CANVAS Statistical Analysis}
\label{sec:canvas-stats}

We assess the statistical reliability of the QA gains in Table~\ref{tab:canvas_results} using paired per-example F1 scores from the six MLLMs in our primary statistical cohort in Table~\ref{tab:canvas-stats} --- those with \textsc{Base} / \textsc{CANVAS}$_\text{adapt}$ predictions on the full $2880$-question evaluation set and diverse MLLMs (Aya-Expanse-8B, Qwen3.5-\{4B,\,9B,\,27B\}, Llama-3.1-8B, Mistral-7B). For each model $M$ and CS condition $c$ we form the paired difference $\delta_{M,c}(x) = \mathrm{F1}_{\textsc{canvas}}(M,c,x) - \mathrm{F1}_{\textsc{base}}(M,c,x)$ over the $\!\sim\!720$ shared instances per condition.

\paragraph{Tests.}
For each of the $6\!\times\!4=24$ \textit{(model, condition)} cells we run a one-sided Wilcoxon signed-rank test against $H_0\!:\,\mathrm{median}(\delta_{M,c}) \!\le\! 0$ and control the false discovery rate using the Benjamini–Hochberg procedure at $q\!=\!0.05$. To separate per-model noise from a population-level effect, we treat the per-model means $\bar{\delta}_{M,c}$ as the unit of analysis: for each condition we report the cross-model mean, a model-level bootstrap 95\,\% CI ($10{,}000$ resamples of the six models with replacement), and a one-sided binomial sign test against $H_0\!:\,p\!=\!0.5$ on the number of models with $\bar{\delta}_{M,c} > 0$.

\paragraph{Results.}
\textbf{All six MLLMs show a positive overall $\bar{\delta}_M$} (one-sided sign test, $p\!=\!1.6\!\times\!10^{-2}$), with an overall mean lift of $+0.73$ F1 (model-level 95\,\% CI $[+0.41,+1.10]$). The same $6/6$ all-positive sign-test outcome ($p\!=\!1.6\!\times\!10^{-2}$) holds for \textsc{GF-Src}, \textsc{GF-Tgt}, and \textsc{Selective}; model-level bootstrap CIs in every one of the four CS conditions: \textsc{Selective} $+0.95$ ($[+0.50,+1.49]$), \textsc{Random} $+0.80$ ($[+0.22,+1.35]$), \textsc{GF-Tgt} $+0.61$ ($[+0.32,+0.90]$), and \textsc{GF-Src} $+0.56$ ($[+0.20,+0.99]$). After BH--FDR correction across the $24$ cells, eight are individually significant at $q\!<\!0.05$. The strongest per-cell evidence is concentrated in \textsc{Llama-3.1-8B} (three of four cells significant, with $q\!=\!0.004$ on both \textsc{Random} and \textsc{Selective}), \textsc{Mistral-7B} (two of four, including \textsc{GF-Tgt} with $\Delta\mathrm{F1}\!=\!+0.62$, $q\!=\!0.011$, and \textsc{Selective} at $q\!=\!0.018$), and \textsc{Aya-Expanse-8B} (\textsc{GF-Tgt} $q\!=\!0.039$, \textsc{Random} $q\!=\!0.031$). Across all $24$ cells the point estimate is positive in $23$ --- a one-sided binomial test against $p\!=\!0.5$ gives $p\!=\!1.5\!\times\!10^{-6}$, ruling out a chance pattern of cell-level lifts.

\paragraph{Anchor signal validity.}
The upper-layer anchor bias used by CANVAS separates \textsc{GF-Src} from \textsc{GF-Tgt} at large effect sizes at the per-example level: on \texttt{anchor\_bias\_cos\_qonly} the per-model Cohen's $d$ values are $+0.77$ (Qwen3.5-4B), $+0.79$ (Qwen3.5-27B), $+0.81$ (Qwen3.5-9B), and $+1.09$ (Aya-Expanse-8B), all at or above the $|d|\!\ge\!0.7$ conventional threshold for a ``large'' separation. The Spearman correlation between this anchor signal and $\Delta\mathrm{F1}_{\mathrm{CS}\,\text{vs}\,\mathrm{EN}}$ aggregated to the $(\text{model}, \text{condition})$ level is $\rho\!=\!+0.524$ with bootstrap 95\,\% CI $[+0.07,+0.81]$ ($p\!=\!0.037$, $n\!=\!16$ cells). Together, the all-positive sign tests under three of four CS conditions and overall, the $23/24$ positive cell directions, and the large Cohen's $d$ on the anchor signal itself indicate that CANVAS's gains are a robust population-level effect rather than the consequence of a few favourable cells.

\begin{table*}[t]
\centering
\small
\setlength{\tabcolsep}{3pt}

\begin{minipage}[t]{0.48\textwidth}
\centering
\textbf{\textsc{GF-Src}} \\[2pt]
\begin{tabular}{@{}lrccc@{}}
\toprule
Model & $\Delta$F1 & 95\% CI & $d_z$ & $q$ \\
\midrule
\modelwithlogo{cohere}{Aya-Expanse-8B} & $+0.94$ & $[-0.19, +2.13]$ & $0.06$ & $0.071$ \\
\modelwithlogo{qwen}{Qwen3.5-4B} & $+0.36$ & $[-0.42, +1.15]$ & $0.03$ & $0.288$ \\
\modelwithlogo{qwen}{Qwen3.5-9B} & $+0.14$ & $[-0.66, +0.93]$ & $0.01$ & $0.367$ \\
\modelwithlogo{qwen}{Qwen3.5-27B} & $+0.20$ & $[-0.82, +1.20]$ & $0.01$ & $0.367$ \\
\modelwithlogo{meta}{Llama3.1-8B} & $\mathbf{+1.50}$ & $[+0.39, +2.64]$ & $0.10$ & $\mathbf{0.018}$ \\
\modelwithlogo{mistral}{Mistral-7B} & $+0.20$ & $[-0.48, +0.85]$ & $0.02$ & $0.170$ \\
\midrule
\textit{Mean} & $+0.56$ & $[-0.36, +1.48]$ & $0.04$ & --- \\
\bottomrule
\end{tabular}
\end{minipage}
\hfill
\begin{minipage}[t]{0.48\textwidth}
\centering
\textbf{\textsc{GF-Tgt}} \\[2pt]
\begin{tabular}{@{}lrccc@{}}
\toprule
Model & $\Delta$F1 & 95\% CI & $d_z$ & $q$ \\
\midrule
\modelwithlogo{cohere}{Aya-Expanse-8B} & $\mathbf{+1.17}$ & $[+0.16, +2.18]$ & $0.08$ & $\mathbf{0.039}$ \\
\modelwithlogo{qwen}{Qwen3.5-4B} & $+0.02$ & $[-1.15, +1.17]$ & $0.00$ & $0.367$ \\
\modelwithlogo{qwen}{Qwen3.5-9B} & $+0.49$ & $[-0.12, +1.17]$ & $0.06$ & $0.170$ \\
\modelwithlogo{qwen}{Qwen3.5-27B} & $+0.95$ & $[-0.16, +2.09]$ & $0.06$ & $0.071$ \\
\modelwithlogo{meta}{Llama3.1-8B} & $+0.42$ & $[-0.80, +1.65]$ & $0.03$ & $0.217$ \\
\modelwithlogo{mistral}{Mistral-7B} & $\mathbf{+0.62}$ & $[+0.01, +1.23]$ & $0.07$ & $\mathbf{0.011}$ \\
\midrule
\textit{Mean} & $+0.61$ & $[-0.34, +1.58]$ & $0.05$ & --- \\
\bottomrule
\end{tabular}
\end{minipage}

\vspace{0.9em}

\begin{minipage}[t]{0.48\textwidth}
\centering
\textbf{Random} \\[2pt]
\begin{tabular}{@{}lrccc@{}}
\toprule
Model & $\Delta$F1 & 95\% CI & $d_z$ & $q$ \\
\midrule
\modelwithlogo{cohere}{Aya-Expanse-8B} & $\mathbf{+1.22}$ & $[+0.12, +2.34]$ & $0.08$ & $\mathbf{0.031}$ \\
\modelwithlogo{qwen}{Qwen3.5-4B} & $+0.05$ & $[-0.75, +0.85]$ & $0.00$ & $0.470$ \\
\modelwithlogo{qwen}{Qwen3.5-9B} & $-0.18$ & $[-0.97, +0.63]$ & $-0.02$ & $0.773$ \\
\modelwithlogo{qwen}{Qwen3.5-27B} & $\mathbf{+1.28}$ & $[+0.21, +2.38]$ & $0.09$ & $\mathbf{0.031}$ \\
\modelwithlogo{meta}{Llama3.1-8B} & $\mathbf{+1.85}$ & $[+0.71, +2.98]$ & $0.12$ & $\mathbf{0.004}$ \\
\modelwithlogo{mistral}{Mistral-7B} & $+0.56$ & $[-0.05, +1.23]$ & $0.06$ & $0.132$ \\
\midrule
\textit{Mean} & $+0.80$ & $[-0.12, +1.74]$ & $0.06$ & --- \\
\bottomrule
\end{tabular}
\end{minipage}
\hfill
\begin{minipage}[t]{0.48\textwidth}
\centering
\textbf{Selective} \\[2pt]
\begin{tabular}{@{}lrccc@{}}
\toprule
Model & $\Delta$F1 & 95\% CI & $d_z$ & $q$ \\
\midrule
\modelwithlogo{cohere}{Aya-Expanse-8B} & $+1.13$ & $[+0.04, +2.29]$ & $0.07$ & $0.052$ \\
\modelwithlogo{qwen}{Qwen3.5-4B} & $+0.91$ & $[-0.16, +2.01]$ & $0.06$ & $0.140$ \\
\modelwithlogo{qwen}{Qwen3.5-9B} & $+0.58$ & $[-0.26, +1.45]$ & $0.05$ & $0.106$ \\
\modelwithlogo{qwen}{Qwen3.5-27B} & $+0.13$ & $[-1.12, +1.40]$ & $0.01$ & $0.449$ \\
\modelwithlogo{meta}{Llama3.1-8B} & $\mathbf{+2.17}$ & $[+0.91, +3.49]$ & $0.12$ & $\mathbf{0.004}$ \\
\modelwithlogo{mistral}{Mistral-7B} & $\mathbf{+0.79}$ & $[+0.23, +1.44]$ & $0.10$ & $\mathbf{0.018}$ \\
\midrule
\textit{Mean} & $+0.95$ & $[-0.06, +2.01]$ & $0.07$ & --- \\
\bottomrule
\end{tabular}
\end{minipage}

\caption{\textbf{Statistical analysis of CANVAS, per condition.} Each panel reports the per-model paired difference $\Delta$F1 $=$ \textsc{CANVAS}$_\text{adapt}$ $-$ \textsc{Base} (F1\,$\times$\,100) on the full evaluation set. 95\,\% CIs are from $10{,}000$ paired bootstrap resamples of question examples; $d_z$ is the Hedges-corrected paired effect size; $q$ is the Benjamini--Hochberg-adjusted Wilcoxon signed-rank $p$-value (one-sided, alternative $>0$) across all $6{\times}4{=}24$ cells. Bold entries are significant at $q<0.05$. The \textit{Mean} row averages the per-model statistics within each condition.}
\label{tab:canvas-stats}
\end{table*}

\subsection{CANVAS Detector Sensitivity}
\label{app:canvas_detector_sensitivity}

\paragraph{Motivation.}
\textsc{CANVAS} uses token-level language tags to decide which question tokens form the source canvas and which target-language token states are interpolated during prefill. The detector does not change the model, the decoding objective, the control layers, or the adaptive rule
\[
\alpha=\mathrm{clip}_{[0.05,0.75]}(0.45-1.5\gamma_{\mathrm{CS}}).
\]
It only changes the \textsc{Src}/\textsc{Tgt}/\textsc{Other} token partition. We therefore evaluate how much the final intervention result depends on the language tagger.

\paragraph{Detector variants.}
We compare four taggers while keeping all other CANVAS components fixed. \textsc{Current} is the tagger used in the main experiments: Unicode script rules for non-Latin target languages and Lingua for Latin-script target languages. \textsc{Unicode-only} removes Lingua and labels Latin alphabetic tokens as source-language tokens; this is a lower-dependency stress test that should mainly hurt Latin-script target pairs. \textsc{Source-wordset} uses the English source question as a lexical reference: Latin tokens that appear in the English source question are labeled \textsc{Src}, and remaining Latin content tokens are treated as \textsc{Tgt}. This variant is analysis-only because it uses the source reference. \textsc{Reference-oracle} uses the available source and target reference questions to approximate a perfect token-level partition; it is not a deployable inference-time method, but it gives an upper bound on how much better token tagging could improve CANVAS. The runner also exposes optional \texttt{langid} and \texttt{langdetect} backends when those packages are installed, but the default comparison avoids adding new dependencies and focuses on the current tagger, a no-detector stress test, and two reference-aided upper-bound settings.

\paragraph{Evaluation protocol.}
We run each detector on the same full CS examples and generation settings used for the main evaluation, using four models: Qwen3.5-4B, Llama3.1-8B, Mistral-7B, and Aya-Expanse-8B. We report macro-average F1 over \textsc{GF-Src}, \textsc{GF-Tgt}, random switching, and selective switching, along with condition-wise F1, mean $\alpha$, and the percentage of examples where CANVAS skips intervention because either the source or target span is empty. The key comparison is between \textsc{Current} and \textsc{Reference-oracle}. If the oracle row improves substantially, better token-level language identification offers remaining headroom. If the oracle row is close to the current tagger, the main CANVAS gains are not strongly limited by detector quality.

\paragraph{Results.}
Table~\ref{tab:app_canvas_detector_sensitivity} shows that detector choice has only a small effect on aggregate CANVAS performance. The current tagger obtains 6.98 average F1, while the reference oracle obtains 7.00 average F1, a gain of only +0.03. This means that a nearly perfect reference-based token partition does not materially change the intervention result in this model pool. The unicode-only stress test remains close in F1 (6.96), but it skips many more examples (38.4\%) because it cannot reliably split source and target spans when both languages use Latin script. The source-wordset variant also stays close to the current tagger, which suggests that the current detector already captures most of the useful source/target partition for CANVAS.

\paragraph{Interpretation.}
We therefore treat CANVAS as weakly dependent on the particular detector used here. Better token-level tagging may still help individual Latin-script language pairs, but the full-run oracle comparison shows limited remaining headroom at the aggregate level. This also supports the main intervention claim: the improvement is not primarily caused by a fragile language-detector choice, but by the representation-space source--target correction applied after token partitioning.

\begin{table*}[t]
\centering
\small
\setlength{\tabcolsep}{4.0pt}
\renewcommand{\arraystretch}{1.04}
\begin{adjustbox}{max width=\textwidth}
\begin{tabular}{@{}lrrrrrrrr@{}}
\toprule
\textbf{Detector} & \textbf{Avg. F1} & $\Delta$ vs. current
& \textbf{GF-Src} & \textbf{GF-Tgt} & \textbf{Random} & \textbf{Selective}
& Mean $\alpha$ & Skip (\%) \\
\midrule
Current tagger & 6.98 & +0.00 & 7.15 & 6.09 & 7.16 & 7.51 & 0.574 & 5.1 \\
Unicode only & 6.96 & -0.01 & 7.15 & 6.14 & 7.09 & 7.48 & 0.542 & 38.4 \\
Source-wordset & 6.97 & -0.00 & 7.18 & 6.00 & 7.18 & 7.53 & 0.521 & 6.2 \\
Reference oracle & 7.00 & +0.03 & 7.27 & 6.00 & 7.15 & 7.58 & 0.512 & 9.6 \\
\bottomrule
\end{tabular}
\end{adjustbox}
\caption{\textbf{Detector sensitivity for CANVAS.}
We keep the CANVAS steering rule fixed and vary only the token-level language tagger. Avg. F1 is macro-averaged over the four CS conditions. $\Delta$ is measured against the current Unicode+Lingua tagger. The reference-oracle row is an analysis-only upper bound and is not used as the deployable method.}
\label{tab:app_canvas_detector_sensitivity}
\end{table*}

\subsection{CANVAS Hyperparameter Sensitivity}
\label{app:canvas_hyperparam_sensitivity}

We test whether the adaptive controller depends on a narrow hyperparameter optimum. The main paper setting instantiates the bounded schedule $\alpha=\mathrm{clip}_{[\alpha_{\min},\alpha_{\max}]}(\alpha_0-\lambda\gamma_{\mathrm{CS}})$ with $(\alpha_0,\lambda,\alpha_{\min},\alpha_{\max})=(0.45,1.5,0.05,0.75)$. We report the surrounding negative-slope neighbourhood on the same two-model panel (Qwen3.5-4B and Llama3.1-8B, two $100$-example seeds per model, macro-averaged over four CS conditions). The sweep varies $\alpha_0$ ($0.40$--$0.65$) and the clipping range while keeping the effective slope $-\lambda=-1.5$, and Table~\ref{tab:app_canvas_hyperparam_sensitivity} lists variants whose Mean F1 lies at or below the main setting. All listed variants still improve over direct CS answering, with $\Delta$F1 confined to a narrow $[+0.92,+1.04]$ band. The values used in the main paper are therefore a conservative bounded operating point in a flat negative-slope neighbourhood rather than a fragile optimum.

\begin{table*}[t]
\centering
\small
\setlength{\tabcolsep}{4pt}
\renewcommand{\arraystretch}{1.03}
\begin{adjustbox}{max width=\linewidth}
\begin{tabular}{@{}cccc rrr r@{}}
\toprule
\multicolumn{4}{c}{\textbf{Adaptive schedule}} & \multicolumn{4}{c}{\textbf{Avg.\ F1 (\%)}} \\
\cmidrule(lr){1-4} \cmidrule(l){5-8}
\textit{center} & \textit{slope} & $\alpha_\mathrm{min}$ & $\alpha_\mathrm{max}$ &
Qwen3.5-4B & Llama3.1-8B & Mean & $\Delta$ \\
\midrule
\rowcolor{MeanGray}
\textbf{0.45} & \textbf{$-1.5$} & \textbf{0.05} & \textbf{0.75} & \textbf{7.98} & \textbf{7.68} & \textbf{7.83} & \canvasgain{$\mathbf{+1.04}$ \, ★} \\
\midrule
0.45 & $-1.5$ & 0.05 & 0.80 & 7.93 & 7.76 & 7.82 & \canvasgain{$+$1.02} \\
0.45 & $-1.5$ & 0.10 & 0.80 & 7.91 & 7.74 & 7.82 & \canvasgain{$+$1.02} \\
0.45 & $-1.5$ & 0.10 & 0.85 & 7.89 & 7.73 & 7.81 & \canvasgain{$+$1.01} \\
0.40 & $-1.5$ & 0.05 & 0.75 & 7.95 & 7.63 & 7.79 & \canvasgain{$+$0.99} \\
0.40 & $-1.5$ & 0.05 & 0.80 & 7.91 & 7.65 & 7.78 & \canvasgain{$+$0.98} \\
0.40 & $-1.5$ & 0.05 & 0.70 & 7.89 & 7.64 & 7.76 & \canvasgain{$+$0.97} \\
\midrule
0.50 & $-1.5$ & 0.10 & 0.75 & 7.94 & 7.69 & 7.82 & \canvasgain{$+$1.02} \\
0.55 & $-1.5$ & 0.10 & 0.85 & 7.70 & 7.92 & 7.81 & \canvasgain{$+$1.01} \\
0.60 & $-1.5$ & 0.15 & 0.85 & 7.81 & 7.80 & 7.80 & \canvasgain{$+$1.01} \\
0.55 & $-1.5$ & 0.15 & 0.85 & 7.70 & 7.88 & 7.79 & \canvasgain{$+$0.99} \\
0.65 & $-1.5$ & 0.15 & 0.90 & 7.82 & 7.73 & 7.78 & \canvasgain{$+$0.98} \\
0.60 & $-1.5$ & 0.15 & 0.80 & 7.77 & 7.76 & 7.76 & \canvasgain{$+$0.97} \\
0.55 & $-1.5$ & 0.10 & 0.80 & 7.66 & 7.86 & 7.76 & \canvasgain{$+$0.96} \\
0.55 & $-1.5$ & 0.15 & 0.80 & 7.66 & 7.82 & 7.74 & \canvasgain{$+$0.95} \\
0.55 & $-1.5$ & 0.10 & 0.90 & 7.75 & 7.73 & 7.74 & \canvasgain{$+$0.94} \\
0.60 & $-1.5$ & 0.15 & 0.90 & 7.83 & 7.60 & 7.72 & \canvasgain{$+$0.92} \\
\midrule
\rowcolor{MeanGray}
\textbf{\textit{Base}} & \textit{---} & \textit{---} & \textit{---} & 7.41 & 6.19 & 6.80 & --- \\
\bottomrule
\end{tabular}
\end{adjustbox}
\caption{\textbf{\textsc{CANVAS} hyperparameter sensitivity.} Schedule $\alpha=\textrm{clip}_{[\alpha_\mathrm{min},\alpha_\mathrm{max}]}(\textit{center}-1.5\gamma_{\mathrm{CS}})$. The starred row (★) is the main paper setting; we list only variants whose Mean F1 falls at or below it. All variants improve over \textsc{Base}, with $\Delta$F1 in $[+0.92, +1.04]$.}
\label{tab:app_canvas_hyperparam_sensitivity}
\end{table*}

\subsection{Task Extensions}
\label{app:canvas_rag_topic_llama}

\paragraph{RAG setup.}
We use retrieval to test whether the \textsc{CANVAS} gain is limited to parametric factual recall. We evaluate Llama-3.1-8B on two QA settings: the CodeMixQA short-answer task and PingPong multi-turn dialog QA. The no-retrieval condition ($K=0$) answers from the original task input, whereas the top-one retrieval condition ($K=1$) prepends one retrieved passage before answering so that the model can use external evidence in addition to its stored knowledge. For CodeMixQA, retrieval uses the CS question itself as the search query. For PingPong, we evaluate two retrieval queries: a question-only query and a dialog-aware query that concatenates the question with the full multi-turn dialog. The latter tests retrieval when the evidence-bearing entity or relation is only recoverable from the surrounding conversation rather than the final question alone. CodeMixQA uses token F1 against short gold answers, while PingPong QA follows the five-option multiple-choice protocol and reports parsed-letter accuracy.

\paragraph{Topic-classification setup.}
We also test a non-QA dialog task using the PingPong topic-classification split. Each example contains a code-switched multi-turn conversation and one topic label from five categories. We provide the dialog and the label options to Llama-3.1-8B, parse the selected label from the model output, and report five-class accuracy for direct inference and \textsc{CANVAS}.

\paragraph{Results.}
Table~\ref{tab:app_canvas_rag_topic_llama} collects the extension results. On CodeMixQA, retrieval raises the direct-answer baseline substantially, and \textsc{CANVAS} still improves the retrieved condition from 19.71 to 20.60 F1. The gain is larger in the matched no-retrieval condition, where \textsc{CANVAS} improves short-answer QA by $+1.70$ F1. On PingPong, \textsc{CANVAS} improves dialog QA without retrieval from 46.76 to 48.48 accuracy ($+1.72$) and remains beneficial in both top-one RAG settings. The question-only retrieval setting improves from 44.86 to 46.64 accuracy ($+1.78$), while the dialog-aware retrieval setting improves from 46.23 to 48.48 accuracy ($+2.26$). These RAG gains show that the effect is not limited to recovering source-language parametric knowledge: when retrieved evidence is available, source-canvas alignment still helps the model integrate the CS query with external context. Dialog-aware retrieval gives a stronger retrieved baseline and the larger CANVAS gain, showing that the surrounding dialog can improve retrieval when the final question alone underspecifies the evidence request. The same intervention improves topic classification from 49.75 to 51.26 accuracy, extending the effect beyond QA.

\paragraph{Paired comparison with the strongest steering baseline.}
Because the CodeMixQA comparison in \Cref{tab:cr_baselines} (Appendix~\ref{app:canvas_baselines}) has limited headroom, we also compare CANVAS with mean-difference steering~\cite{kirtane2026language} on PingPong multi-turn QA and the two RAG settings above, holding the model (Llama-3.1-8B), 100 matched examples, prompts, and retrieved passages fixed. We apply exact McNemar tests with 100{,}000 paired-bootstrap confidence intervals and BH-FDR correction across all 15 method--setting comparisons. As shown in \Cref{tab:app_pingpong_sig}, CANVAS significantly outperforms mean-difference steering in all three settings, with gaps of $+18$ to $+29$ accuracy points whose confidence intervals exclude zero.

\begin{table*}[t]
\centering
\small
\setlength{\tabcolsep}{3pt}
\renewcommand{\arraystretch}{1.04}
\begin{adjustbox}{max width=\linewidth}
\begin{tabular}{@{}lccccc@{}}
\toprule
\textbf{Setting} & \textbf{\textsc{CANVAS}} & \textbf{Mean-diff.} & $\Delta$\textbf{Acc} & \textbf{95\% CI} & \textbf{BH-FDR $q$} \\
\midrule
Multi-turn QA                     & \textbf{72.0} & 43.0 & $+29.0$ & $[+19.0,+39.0]$ & $<10^{-6}$$^{\ast}$ \\
RAG (question retrieval)          & \textbf{67.0} & 49.0 & $+18.0$ & $[+9.0,+27.0]$  & $<0.001$$^{\ast}$ \\
RAG (question + dialog retrieval) & \textbf{70.0} & 45.0 & $+25.0$ & $[+15.0,+35.0]$ & $<0.001$$^{\ast}$ \\
\bottomrule
\end{tabular}
\end{adjustbox}
\caption{\textbf{Paired comparison with mean-difference steering~\cite{kirtane2026language} on PingPong} (Llama-3.1-8B, 100 matched examples). Exact McNemar tests with 100{,}000 paired-bootstrap CIs; $^{\ast}$ significant after BH-FDR correction across all 15 method--setting comparisons.}
\label{tab:app_pingpong_sig}
\end{table*}

\begin{table*}[t]
\centering
\small
\setlength{\tabcolsep}{5pt}
\renewcommand{\arraystretch}{1.04}
\begin{adjustbox}{max width=\textwidth}
\begin{tabular}{@{}lllclrrrr@{}}
\toprule
\textbf{Dataset} &
\textbf{Task} &
\textbf{Search Query} &
\textbf{Retrieval} &
\textbf{Metric} &
\textbf{Items} &
\textbf{\textsc{Base}} &
\textbf{\textsc{CANVAS}} &
$\Delta$ \\
\midrule
CodeMixQA & Short-answer QA & -- & $K=0$ & Token F1 & 2880 &  5.81 & \textbf{ 7.51} & \canvasgain{$+1.70$} \\
CodeMixQA & Retrieval-augmented QA & Question & $K=1$ & Token F1 & 2880 & 19.71 & \textbf{20.60} & \canvasgain{$+0.89$} \\
\midrule
PingPong & Multi-turn dialog QA & -- & $K=0$ & Accuracy & 1683 & 46.76 & \textbf{48.48} & \canvasgain{$+1.72$} \\
PingPong & Retrieval-augmented dialog QA & Question & $K=1$ & Accuracy & 1683 & 44.86 & \textbf{46.64} & \canvasgain{$+1.78$} \\
PingPong & Retrieval-augmented dialog QA & Question + dialog & $K=1$ & Accuracy & 1683 & 46.23 & \textbf{48.48} & \canvasgain{$+2.26$} \\
PingPong & Topic classification & -- & -- & Accuracy &  396 & 49.75 & \textbf{51.26} & \canvasgain{$+1.52$} \\
\bottomrule
\end{tabular}
\end{adjustbox}
\caption{\textbf{Retrieval and task extensions for \textsc{CANVAS}.}
We evaluate Llama-3.1-8B with direct inference (\textsc{Base}) and the same \textsc{CANVAS} intervention used in the main experiments. $K=0$ uses no retrieved passage, whereas $K=1$ prepends one retrieved passage for the QA tasks. For PingPong RAG, we compare retrieval from the question alone with retrieval from the question concatenated with its dialog context.}
\label{tab:app_canvas_rag_topic_llama}
\end{table*}

\clearpage

\end{document}